\pdfoutput=1

\documentclass[11pt]{article}

\usepackage[preprint]{acl}

\usepackage{times}
\usepackage{latexsym}

\usepackage[T1]{fontenc}

\usepackage[utf8]{inputenc}

\usepackage{microtype}

\usepackage{inconsolata}

\usepackage{graphicx}

\usepackage{arydshln}

\usepackage{algorithm} 
\usepackage{algpseudocode}

\usepackage[skins]{tcolorbox}
\usepackage{graphicx}
\usepackage{enumitem}
\usepackage{makecell}

\usepackage{booktabs}
\usepackage{multirow}

\usepackage{soul}

\newcolumntype{C}[1]{>{\centering\arraybackslash}m{#1}}

\tcbuselibrary{listingsutf8}
\global

\tcbset{
  aibox/.style={
    width=474.18663pt,
    top=10pt,
    colback=white,
    colframe=black,
    colbacktitle=black,
    enhanced,
    center,
    attach boxed title to top left={yshift=-0.1in,xshift=0.15in},
    boxed title style={boxrule=0pt,colframe=white,},
  }
}
\newtcolorbox{AIbox}[2][]{aibox,title=#2,#1}

\usepackage{amsmath}

\usepackage{graphicx}
\usepackage{cleveref}
\usepackage{booktabs}
\usepackage{subcaption}
\usepackage{array}

\usepackage{soul}

\usepackage{caption}
\usepackage{subcaption}

\usepackage{booktabs}
\usepackage{multirow}
\usepackage{makecell}

\usepackage{enumitem}
\usepackage{amsmath}
\usepackage{svg}

\usepackage{tabularx}             

\usepackage{annotation}
\usepackage{tikz}
\usepackage{hyperref}

\definecolor{pastelBlue}{RGB}{174,198,207}   
\definecolor{pastelRed}{RGB}{255,179,186}      
\definecolor{pastelOrange}{RGB}{255,223,186}   
\definecolor{pastelGreen}{RGB}{203,240,191}    
\definecolor{pastelViolet}{RGB}{213,181,226}   
\definecolor{pastelYellow}{RGB}{255,255,204}   
\definecolor{pastelPurple}{RGB}{230,230,250}   
\definecolor{pastelPink}{RGB}{255,192,203}     
\definecolor{highlightGreen}{HTML}{D4F4E9}

\newcommand{\framework}{FRAME}      
\newcommand{\pmetric}{P-MESA}  
\newcommand{\reasoning}{SCOPE}   
\newcommand{\fact}{fact}

\title{Re-FRAME the Meeting Summarization SCOPE: \\ Fact-Based Summarization and Personalization via Questions}

\author{Frederic Kirstein\textsuperscript{1}, Sonu Kumar, Terry Ruas, Bela Gipp \\
  University of Göttingen, Germany \\
\textsuperscript{1}\texttt{kirstein@gipplab.org} }

\begin{document}
\maketitle
\AddAnnotationRef
\begin{abstract}

Meeting summarization with large language models (LLMs) remains error-prone, often producing outputs with hallucinations, omissions, and irrelevancies.
We present \textbf{\framework{}}, a modular pipeline that reframes summarization as a semantic enrichment task.
\framework{} extracts and scores salient facts, organizes them thematically, and uses these to enrich an outline into an abstractive summary.
To personalize summaries, we introduce \textbf{\reasoning{}}, a \emph{reason-out-loud} protocol that has the model build a reasoning trace by answering nine questions before content selection.
For evaluation, we propose \textbf{\pmetric{}}, a multi-dimensional, reference-free evaluation framework to assess if a summary fits a target reader.
\pmetric{} reliably identifies error instances, achieving $\geq89\%$ balanced accuracy against human annotations and strongly aligns with human severity ratings ($\rho \geq 0.70$).
On QMSum and FAME, \framework{} reduces hallucination and omission by 2 out of 5 points (measured with MESA), while \reasoning{} improves knowledge fit and goal alignment over prompt-only baselines.
Our findings advocate for rethinking summarization to improve control, faithfulness, and personalization\footnote{Resources are available as per \Cref{app:repository} on \href{https://github.com/FKIRSTE/emnlp2025-reframe-summarization}{GitHub}.}.

\end{abstract}

\section{Introduction}
\label{sec:introduction}

\begin{figure*}[t]
    \centering
    \includegraphics[width=0.95\linewidth]{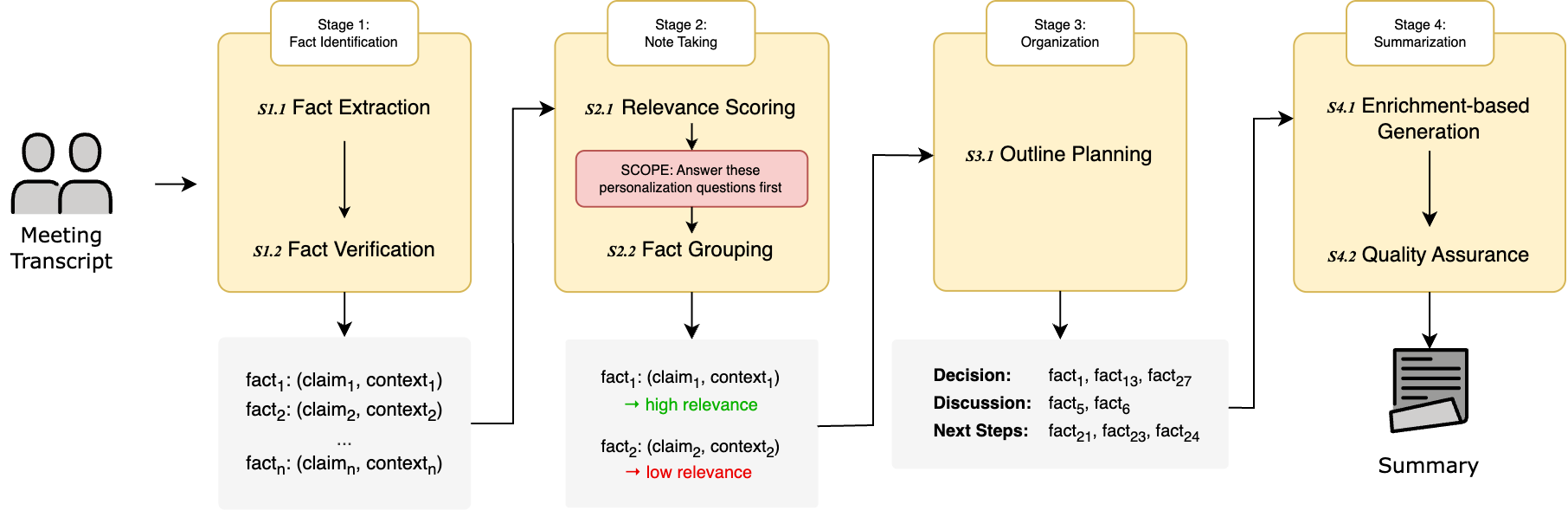}
    \caption{\framework{} pipeline with \reasoning{} integration. \framework{} structures summarization in four stages: fact identification, note taking, organization, and enrichment-based generation. \reasoning{} plugs into salience scoring by injecting a reasoning trace derived from reader-specific questions.}
    \label{fig:pipeline_diagram}
\end{figure*}

Meetings can be dense, chaotic, and high-stakes.
Summarizing them effectively is a natural language processing (NLP) challenge with value for corporate, academic, and governmental contexts \cite{ZhongYYZ21h,HuGDD23a,LaskarFCB23,KirsteinWGR25}.
Yet current large language model (LLM)-based systems \cite{LaskarFCB23,FuLKC24a} continue to produce summaries that omit key points, hallucinate content, and struggle with relevance \cite{GoliaK23, KirsteinLG25a}.
We argue that these weaknesses stem from a more profound structural mismatch that current approaches treat conversation like linear text, compressing form without reconstructing meaning.

Meetings hold three challenges compared to traditional structured texts \cite{ArabzadehAKL23a,KirsteinKWR25}:
(1) Salient content is scattered across speaker turns
(\emph{Information Distribution}),
(2) utterances depend on long-range context (\textit{Contextual Dependencies}), and
(3) salience varies per reader (\textit{Salience Ambiguity}).
Current summarization methods are not aligned with these properties.
Approaches relying on structural cues (e.g., sections, paragraphs) \cite{LiuL19b} are unsuitable for meetings' boundary-free nature.
Chunking methods \cite{ZhangNMW22b} struggle with cross-chunk dependencies, while hybrid extractive-abstractive models \cite{LiEIP21} may lose interpretability.
Each approach shares the same limitation: condensing information without reconstructing the underlying semantic structure.

Our \textbf{\framework{}} (\emph{\textbf{F}act-based \textbf{R}econstruction and \textbf{A}bstractive \textbf{ME}eting Summarization}) framework is a fact‑centric pipeline that reframes the established meeting summarization workflow as an \emph{enrichment} task (shown in \Cref{fig:pipeline_diagram}).
Drawing on research in fact extraction \cite{GunjalD24,WannerEJD24} and summary planning \cite{AmplayoAL21,GrenanderVCV25}, \framework{} mimics how humans summarize texts in four stages \cite{EndresNiggemeyer00}:
We extract self-contained verifiable \fact{}s to cut filler content (\textit{Fact Identification}), filter them by salience (\textit{Note‑Taking}), organize them into an outline by thematic relationships (\textit{Organization}), and enrich these outlines to abstractive summaries strictly using facts (\textit{Summary Writing}).
With GPT-4o \cite{OpenAIAAA24}, \framework{} improves output quality compared to recent approaches.
On QMSum \cite{ZhongYYZ21h} and FAME \cite{KirsteinKWR25}, \framework{} cuts hallucination by up to 3 out of 5 points (3→1, 4→1; lower is better), and irrelevance by 1 point (2→1, 3→1), beating GPT-4o and Gemini-1.5 pro\footnote{We will refer to them as GPT and Gemini.} \cite{GeminiTeamRST24} across six of eight MESA dimensions \cite{KirsteinLG25}.

Salience ambiguity remains widely unaddressed by current systems, despite the growing interest in personalization.
Meetings involve diverse roles and objectives, yet most systems produce a one-size-fits-all output, ignoring role-specific goals and expertise \cite{KirsteinRKG24a}.
We introduce the \textbf{\reasoning{}} protocol (\emph{\textbf{S}ummarizing \textbf{C}ontent \textbf{O}riented to \textbf{P}ersonal \textbf{E}xpectations}) that guides an LLM through an explicit \emph{reason-out-loud} approach before scoring facts. 
Drawing on cognitive science research \cite{Solomon95,Konrad17}, \reasoning{} has the model answer a questionnaire about the reader's goals, expertise, and understanding.
This creates an explicit reasoning trace to ground content selection, improving personalization.

As evaluating personalized summaries is hard with existing metrics (e.g., ROUGE \cite{Lin04}, MESA), we propose \textbf{\pmetric{}} (\textbf{P}ersonalized MESA), a reference-free LLM-based personalization metric with seven dimensions: \emph{factuality, completeness, relevance, goal alignment, prioritization, knowledge-fit}, and \emph{contextual framing}.
\pmetric{} correlates strongly with human judgment (avg. Spearman $\rho = 0.76$)  on 50 LLM-generated summaries. 
On our new \pmetric{} metric (5-point Likert score on quality impact), \reasoning{} improves knowledge-level fit (2→1) and goal alignment (3→2), reducing oversimplification and reader hallucination.
As such, \reasoning{} outperforms role-playing \cite{KirsteinRKG24a,ZhangHZ25} and reader-tailoring prompting \cite{GhodratnamaZ24} by modeling why information matters, not just to whom.

\noindent
This paper makes three key contributions:
\begin{itemize}[noitemsep, topsep=0pt, leftmargin=*] 
    \item \textbf{\framework{}}: A modular summarization pipeline that treats summarization as enrichment, improving factuality, coherence, and salience handling.  
    \item \textbf{\reasoning{}}: A personalization protocol that models reader intent via reason-out-loud, outperforming persona injection.  
    \item \textbf{\pmetric{}}: A reader-centric metric quantifying personalization quality without references that aligns with human preferences.
\end{itemize}

\section{Related Work}
\label{sec:related_work}
\paragraph{Meeting summarization} is about distilling multi-speaker dialogue with distributed information \cite{RennardSHV23,KirsteinWGR25}.
Prior work treats this as a reduction problem, aiming to condense dialogue using LLM prompting \cite{LaskarFCB23,FuLKC24a,TangSWB24}, role vectors \cite{AsiWEG22a}, hierarchical encoding \cite{NarakiSH22a}, or self-refinement \cite{KirsteinLG25a}.
These methods improve coherence but often struggle with understanding the meeting's content \cite{KirsteinWGR25} as they do not reconstruct the underlying meaning explicitly before summary generation.
In contrast, \framework{} reframes summarization as an enrichment task by first extracting and grouping \fact{}s from the transcript, planning a high-relevance summary, and enriching the outline to form a summary.

\paragraph{Fact extraction}  transforms text into self-contained, verifiable units \cite{KamoiGDD23,MinKLL23a}, increasingly used for claim extraction and fact verification \cite{ChernCCY23,ChiangL24,WangGMA24}.
Prior works on fact extraction in dialogue summarization either enrich utterances with extracted factual statements to reduce hallucinations \cite{ZhangYW24}, or target specific fact types (e.g., action items) via classification and neighborhood-based rephrasing \cite{GoliaK23}.
These approaches remain superficial, lacking a general, structured fact representation.
Typical fact representations may discard salient discourse cues or miss global dependencies \cite{GunjalD24,WannerEJD24}.
We address this with \emph{statement–context tuples}, a structured fact representation pairing claims to global context.
Unlike others, our representation retains interpretability, allowing for better content comprehension (see \Cref{sec:app_fact_representation_comparison}).

\paragraph{Personalization} is about adapting summaries to reader expectations \cite{KirsteinRKG24a,KirsteinWGR25}.
Recent approaches zero-shot LLMs to align content with a reader's profile \cite{KirsteinRKG24a,Paoli23} or model participants as graph nodes to extract personalized views \cite{JungSJC23}.
These approaches can surface relevant points but lack consistency and are prone to reader perspective hallucination \cite{ZhangRKS24}.
Human-in-the-loop systems \cite{ChenDY23a,GhodratnamaZ24} mitigate this with feedback, but remain laborious, costly, and time-consuming.
We approach robust salience detection through our \reasoning{} reason-out-loud protocol, inspired by cognitive science \cite{Solomon95} and think-out-loud protocols observed in human summarizers \cite{EndresNiggemeyer00}.
\reasoning{} guides the LLM to answer a questionnaire to build an explicit trace of the reader's intent, expertise, and goals before selecting salient facts.
\reasoning{} outperforms established approaches (see \Cref{app:personalization-approach}) and works zero-shot, making it scalable and generalizable.

\section{Methodology}
\label{sec:methodology}

\paragraph{Overview.}
Unlike sequence-to-sequence approaches that attempt one-hop transcript summarization \cite{LaskarFCB23}, \framework{} handles summarization in four stages as a structured enrichment task, inspired by the human summarization process \cite{EndresNiggemeyer00}: extracting salient information, assessing their relevance, organizing them thematically, and synthesizing a coherent narrative.
For personalization, we introduce \reasoning{}, a structured reason-out-loud protocol that enforces generating a reasoning trace to ground salience detection.
\Cref{fig:pipeline_diagram} illustrates the complete \framework{} framework and how \reasoning{} integrates.

\begin{figure}
    \centering
    \includegraphics[width=1\linewidth]{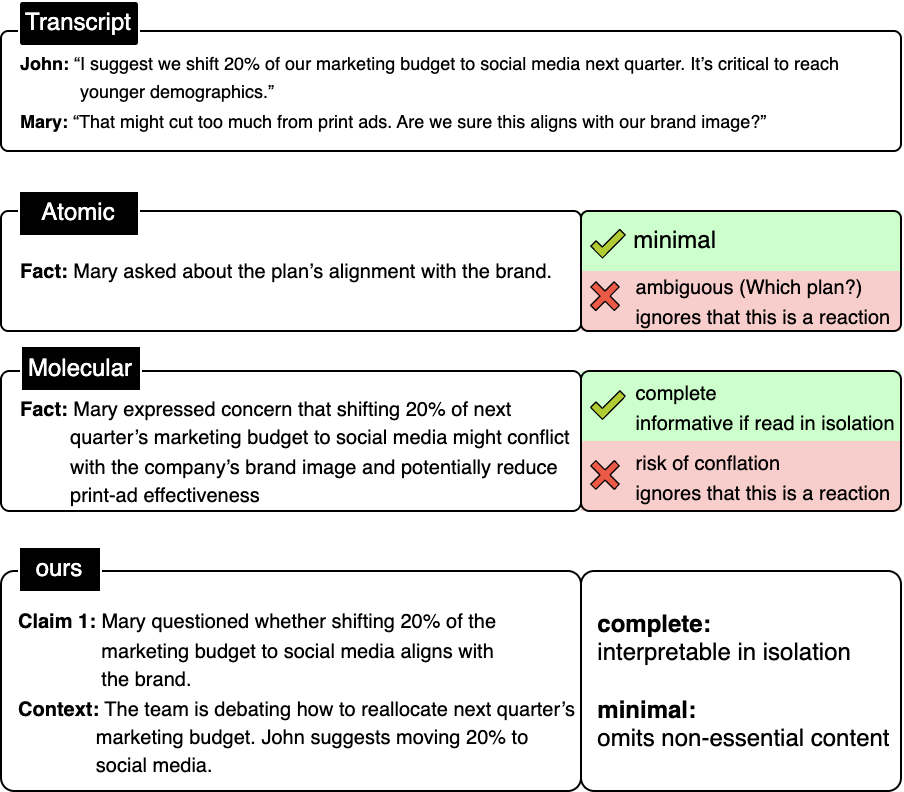}
    \caption{Comparison of our statement-context tuple (OURS) against a high-granularity fact (Atomic) and a high-context fact (Molecular).}
    \label{fig:atomic_fact_example}
\end{figure}

\subsection{Fact Definition}
\label{subsec:atomic_facts}

\paragraph{Motivation.}
Summarizing meetings requires distilling meaning from fragmented, implicit, and highly contextual speaker turns.
While explored throughout NLP \cite{NenkovaP04,ZhangB21,LiuFLZ23a,MinKLL23a}, existing fact detection setups have marked limitations.
For convoluted texts, facts can become too granular \cite{WannerEJD24}, resulting in broken context and a loss of interpretability \cite{LiOWZ16,GunjalD24}.

We therefore define a \fact{} as \emph{statement–context tuple} $\langle c, \kappa \rangle$ where $c$ is a self-contained claim, and $\kappa$ is the minimal global context required for its interpretation such that the original meaning of $c$ remains preserved \cite{ChoiPLK21a}.

\paragraph{Desiderata.}
Let $T$ be a transcript of utterances ${u_1, \dots, u_n}$.
A \fact{} must satisfy:

\noindent
\textit{Completeness:} All references in $c$ must be resolvable using $\kappa$ without requiring external knowledge.

\noindent
\textit{Minimalism:} $c$ should convey only one idea. $\kappa$ includes only the details essential for grounding $c$.

\Cref{fig:atomic_fact_example} illustrates these criteria, demonstrating how too much detail introduces noise while insufficient context creates ambiguity, contrasting our \fact{}s with existing fact setups.

\subsection{\framework{} for General Summarization}
\label{subsec:general_summary_pipeline}

\framework{} operates in four sequential stages (see \Cref{fig:pipeline_diagram}): Fact Identification, Note-Taking, Organization, and Summarization.
Complete implementation details and examples appear in \Cref{app:experimental_setup}.

\subsubsection{Stage 1: Fact Identification} \label{subsubsec:extraction}
This stage cuts filler content while preserving semantic content.

\paragraph{Fact Extraction.}
Given a transcript $T$, we prompt an LLM to extract facts $F = \{f_1, \dots, f_m\}$ as tuples $\langle c_i, \kappa_i \rangle$.
The prompt includes contrastive examples to filter fillers (\emph{"OK," "Mm-hmm"}), hedged statements (``Maybe we should...''), and compound facts \citep{ZhuLQ24}. 
Human evaluation confirms this extraction process preserves relevant content (see \Cref{app:fact_verification_impact}).

\paragraph{Fact Verification.}
To ensure reliability, we verify facts using an LLM judge inspired by FActSCORE \cite{MinKLL23a}, checking $F$ against $T$ for factuality (\textit{action}: removing unsupported claims), completeness (\textit{action}: adding missed key information), clarity (\textit{action}: re-writing $\kappa$ for self-containment), and minimalism (\textit{action}: removing extraneous details from $\kappa$).
We find refinement of facts being required in $\sim$5$\%$ cases (see \Cref{app:fact_verification_impact}).

\subsubsection{Stage 2: Note-Taking}
\label{subsubsec:relevance_detection}
This stage handles fact relevance and redundancy.

\paragraph{Relevance Scoring.}
Inspired by categorizing facts according to their content in fact-checking \cite{HassanALT17}, we task an LLM judge with assigning each fact $f_i$ a function label (i.e.,  \textsc{Decision}, \textsc{Action Item}, \textsc{Insight}, or \textsc{Context}) and a relevance score $r_i \in [1, 10]$ (higher is more relevant).
The $r_i$ ranges are chosen empirically: decisions (9–10), key insights (7–8), supporting context (4–6), and low-salience background (1–3).
On average, we retain 40\% of facts (see \Cref{app:retention_rates}), reducing information overload, allowing for content control, and mitigating positional bias \cite{LiuAGK23,XiaoTCH24}.
We show the influence of varying the $r_i$ ranges in \Cref{app:threshold_variation}, observing that too relaxed or strict thresholds impact summary quality.

\paragraph{Fact Grouping.}
To mitigate redundancy while capturing emphasis signaled by repeated stating of a fact, retained facts are grouped by their function label and relevance scores.
Within each group, we use an LLM to identify semantically overlapping facts, consolidate them by synthesizing contexts, and preserve the highest relevance score.

\subsubsection{Stage 3: Organization}
\label{subsubsec:outline_generation}
This stage creates an outline to guide the summary.

\paragraph{Outline Planning.}
We convert rated facts into a structured outline that reflects the conversation’s logic, orienting on summary planning \citep{AmplayoAL21,GrenanderVCV25}.
High-relevance ($r_i$ $\geq$ 8) and \textsc{Decision} facts become major outline points. 
Mid-relevance ($6 \leq r_i < 8$) and \textsc{Context} facts are considered during summary writing to provide background and flow.
Outlines present main topics, discussions, and next steps.
In \Cref{app:retention_rates} we find that a 250-token summary covers $\sim$9 high-relevance facts as anchors supported by $\sim$12 contextual facts.

\subsubsection{Stage 4: Summarization}
\label{subsubsec:summary_writing}
This stage enriches the outline to a summary.

\paragraph{Enrichment-Based Generation.}
An LLM converts the outline into an abstractive summary by enriching each anchor with supporting facts ($r_i \geq 6$).
Generation is constrained from introducing content beyond extracted facts.
Unsupported outline points are removed rather than speculated upon. 
In our ablation in \Cref{app:retention_rates}, we did not observe the inclusion of unsupported outline points, concluding that this is a rare occasion when using strong backbone models for \framework{}.

\paragraph{Quality Assurance.} 
An LLM acts as a reviewer, assigning error points to the summary draft on four dimensions (see \Cref{app:prompt_templates} for prompts): \emph{outline adherence} (max. 4 points), \emph{factual accuracy} (max. 3 points), \emph{information coverage} (max. 2 points), and \emph{formatting} (max. 1 point).
Each detected error equals one error point.
The maximum error points per category are empirically chosen to reflect importance.
If any dimension exceeds its maximum, or the total exceeds four points, the system initiates a revision cycle with feedback.
The feedback includes observed issues and the LLM's chain-of-thought reasoning \cite{WeiWSB24}.
In our experiments, a single revision pass sufficed.

\subsection{\reasoning{}: Personalization via Reason-Out-Loud} 
\label{subsec:personalization_reasoning}

Personalization typically happens through role-playing \cite{ZhangHZ25} and reader-tailoring \cite{GhodratnamaZ24}.
As such, the model has to implicitly interpret how a role or goal affects content selection while summarizing.
This often leads to missed needs \cite{ZhangRKS24}, as observed in \Cref{app:personalization-approach}.

As to cognitive science research, explicit verbalization enhances metacognitive awareness and decision consistency \cite{Solomon95,Konrad17}.
We propose \textbf{\reasoning{}}, a \emph{reason-out-loud} protocol for robust personalization guiding an LLM through an \emph{exploration} step before \emph{fact selection}.

In \emph{exploration}, the LLM builds an assessment trace by answering a questionnaire (details on questions in \Cref{tab:reasoning_questionnaire} in \Cref{appendix:reasoning_questionnaire}) about reader characteristics (background, expertise), specific needs (interests, knowledge gaps), information utility (actionability, responsibility alignment), and contextual relevance (need for elaboration).
This leverages a persona description, either provided or inferred from the transcript \cite{KirsteinRKG24a}, containing information about the reader's role, expertise, and standpoint.
In the subsequent \emph{fact selection}, this trace is used to ground the relevance determination process of each fact and thereby tailor the selection to the user.

\reasoning{} acts as filter during the \texttt{Note-Taking} stage.
Extracted facts are pre-selected here before being scored (\texttt{Relevance Scoring}).
For complete personalization, the reader profile is also used for \texttt{Outline Generation} and \texttt{Summary Writing}.


\section{Interlude: \pmetric{}}
\label{sec:pmesa}
\label{subsec:evaluation_gap}

Personalization in meeting summarization is gaining traction \cite{KirsteinWGR25}, but the field lacks a way to evaluate whether a summary serves a reader's preferences.
\textbf{Reference-based metrics} like ROUGE \cite{Lin04} and BERTScore \cite{ZhangKWW20} require gold summaries, which are typically unavailable in personalized forms.
\textbf{Reference-free metrics} such as MESA \cite{KirsteinLG25} assess general quality aspects (e.g., coreference, irrelevance) but do not judge alignment to reader expectations.
Recent general \textbf{personalization metrics} like EGISES \cite{VanshRDC23a} and PerSEval \cite{DasguptaCBM24} are designed to rank-order responses, but not to judge whether a single summary is helpful for a specific user.
As a result, researchers have to choose between expensive human evaluations and metrics that overlook reader-centric dimensions, stalling progress in personalized summarization.

\begin{table*}[t]
    \centering
    \renewcommand{\arraystretch}{1.0} 
    \scriptsize
    \setlength{\tabcolsep}{4.0pt}     
    
    \begin{tabular}{@{}llcccccccc c}
        \toprule
        \rowcolor{gray!20}
        & \textbf{Metric} 
            & \makecell{\textbf{Factuality} \\ \textbf{ }}
            & \makecell{\textbf{Relevance} \\ \textbf{ }}
            & \makecell{\textbf{Goal}\\\textbf{Alignment}}
            & \makecell{\textbf{Prioritization} \\ \textbf{ }} 
            & \makecell{\textbf{Personal}\\\textbf{Preferences}}
            & \makecell{\textbf{Knowledge-}\\\textbf{level Fit}}
            & \makecell{\textbf{Contextual}\\\textbf{Framing}}
            &
            & \makecell{\textbf{Overall} \\ \textbf{(Average)}} \\
        \midrule
        detection related & \textbf{B-ACC (\%)}    
            & 91.7   & 93.3 & 92.7 & 90.2 & 89.4 & 90.0 & 89.1 && 92.2 \\
        & \textbf{Cohen's $\kappa$}      
            &  0.83  & 0.74 & 0.76 & 0.81 & 0.79 & 0.62 & 0.78 && 0.81 \\
        &\textbf{FNR (\%)}      
            & 7.2  & 6.3  & 3.1   &  8.9  & 9.5  & 11.1 & 8.7  && 7.8  \\
        &\textbf{FPR (\%)}      
            & 9.4  & 10.1 & 10.3   & 10.7 & 11.8 & 11.0 & 13.1 && 10.9 \\
        \midrule
        sensitivity related &\textbf{Spearman $\rho$}  
            & 0.76 & 0.78 & 0.81   &  0.73 & 0.75 & 0.70 & 0.79 && 0.76 \\
        & \textbf{Kendall $\tau$}   
            & 0.71 & 0.74 & 0.76   &  0.66 & 0.68 & 0.62 & 0.74 && 0.70 \\
        \bottomrule
    \end{tabular}
    \caption{Analyzing P-MESA's detection accuracy and severity correlation with human annotations. For detection, we report B-ACC: Balanced Accuracy, $\kappa$: Cohen's Kappa, FNR: False Negative Rate, FPR: False Positive Rate. For severity, we show $\rho$: Spearman's rank correlation, $\tau$: Kendall's tau.}
    \label{tab:combined_p-mesa}
\end{table*}

We introduce \textbf{\pmetric{}} (\textbf{P}ersonalized - \textbf{ME}eting \textbf{S}ummary \textbf{A}ssessor), a multi-dimensional, reference-free evaluation framework designed to test whether a single summary satisfies a single target reader.
\pmetric{} scores summaries across seven personalization dimensions motivated from a 50-paper literature review (\Cref{subsec:taxonomy_construction}): \emph{factuality}, \emph{completeness}, \emph{relevance}, \emph{goal alignment}, \emph{priority structuring}, \emph{knowledge-level fit}, and \emph{contextual framing}.
Definitions are given in \Cref{app:pmetric_lit_review}.

Each dimension covers a distinct personalization characteristic.
\pmetric{} is powered by an LLM evaluator (GPT) that receives a reader profile containing role, knowledge level, goals, and interests.
This evaluator uses this context information to rate summaries on each dimension on a 5-point Likert scale, with higher scores indicating higher impact.

\subsection{Criteria Definition}
\label{subsec:taxonomy_construction}

We derive the \pmetric{} dimensions through a literature analysis and empirical refinement.

\paragraph{Step 1: Literature review.}
We review 50 papers from *CL venues (2018–2024) on terms such as ``personalization'', ``user modeling'', and ``adaptation'', and discard 14 works due to an unwanted focus on related topics (e.g., personalized agents, style transfer).
We manually screen the remaining 36 papers, and we identify nine candidate evaluation dimensions (approach detailed in \Cref{app:pmetric_lit_review}).

\paragraph{Step 2: Human study.}
To test the dimensions' clarity, coverage, and non-overlap, we conduct a human study in which annotators apply them to model-generated personalized summaries.
We construct a one-time evaluation dataset of 48 summaries using GPT and Gemini.
Each summary is generated for a distinct reader profile specifying role, prior knowledge, and goals, and is prompted with the task to either (a) \textit{summarize for} or (b) \textit{simulate} that user (24 each).
The samples are drawn evenly from QMSum and FAME, covering 8 meeting types, 14 topics, and an average of 5.9 speakers.

Three annotators (ages 22–29, C1+ English; see \Cref{app:human_evaluation} for all details) annotate each summary using the candidate dimensions.
Each summary receives two annotations.
Annotators provide structured feedback on definition clarity, dimension overlap, and any lack of criteria.
We collect feedback through forms and daily group discussions.

\paragraph{Step 3: Final dimension set.}
Following the feedback from the discussions, \emph{Personal Preference} is dropped due to poor generalizability. 
\emph{Objective Alignment} and \emph{Information Utility} are merged into \emph{Goal Alignment} for a macro-level intent alignment, contrasting from \emph{Priority Structuring} (micro-level salience and ordering).
These steps result in the seven personalization dimensions used in \pmetric{}.

\subsection{Metric Implementation}
\label{subsec:pmesa_design}

We build \pmetric{} on the structure of MESA's LLM judge \cite{KirsteinLG25} as their framework provides a reasonable compromise between thoroughness and cost.
\pmetric{} uses a three-stage evaluation pipeline:  
(1) potential error instance detection, (2) instance's severity rating, and (3) impact score aggregation per category.  
\pmetric{} also includes the reader profile, either derived or directly given, to each evaluation prompt to guide the LLM judges.
Each profile includes the reader’s role, goals, expertise, and contextual constraints.
Similar to MESA, we use GPT as the backbone model.

\subsection{\pmetric{} as Proxy for Human Judgment}
\label{subsec:metric_validation}

We assess whether \pmetric{} reliably approximates human annotation by detecting the presence of personalization errors, and assigning severity scores that reflect their perceived impact on reader utility (see \Cref{tab:combined_p-mesa}).
For this assessment, we generate a second and fully independent set of 48 personalized summaries using the same generation setup as in \Cref{subsec:taxonomy_construction}, but with no overlap in meetings to the previous set.
This dataset is used only once, strictly for evaluation purposes.

All six annotators (three male, three female; ages 22–29; C1+ English) rate each summary using the finalized seven \pmetric{} dimensions on a 0–5 Likert scale (0: no error, 5: maximal impact).
Annotators undergo one week of training, including calibration rounds and regular joint discussions.
Full human evaluation protocol details for this annotation are provided in \Cref{app:human_evaluation}.

\paragraph{Detection Accuracy of \pmetric{}.}
We bin \pmetric{} scores $\geq$1 as indicating error presence and compute balanced accuracy\footnote{Balanced accuracy averages sensitivity and specificity. Formal definition given in \Cref{sec:appendix_b-acc}.} (B-ACC) and Cohen’s $\kappa$ to evaluate agreement between \pmetric{} and human annotators.
\pmetric{} achieves high detection accuracy across all dimensions, with B-ACC exceeding 89\% and peaking for the persona-grounded criteria at 93.3\% (\texttt{Relevance}) and 92.7\% (\texttt{Goal Alignment}).
This indicates that \pmetric{} can identify weaknesses, even when subtle or varying in frequency.
Cohen’s $\kappa \geq 0.74$ indicates agreement beyond chance, showing that \pmetric{} applies consistent, human-aligned decision rules.
False negatives are rare in the persona-grounded dimensions (\texttt{Relevance}: 6.3\%, \texttt{Goal Alignment}: 3.1\%), suggesting \pmetric{} is unlikely to miss high-impact weaknesses.
False positive rates of  $\sim$11\% further reflect a conservative bias ensuring that borderline weaknesses are not overlooked.

\paragraph{Error Severity Assessment.}
We compute Spearman’s $\rho$ and Kendall’s $\tau$ between \pmetric{} and human severity ratings to test how sensitive \pmetric{} is to error severity changes.
\pmetric{} shows strong rank correlation across all dimensions, indicating a good proxy for human judgment in reflecting error impact.
Agreement is highest in \texttt{Goal Alignment} ($\rho = 0.81$ and $\tau = 0.76$) and \texttt{Relevance} ($\rho = 0.78$, $\tau = 0.74$). 
Correlations are slightly lower for more interpretive dimensions like \texttt{Knowledge}knowledge-level fit ($\rho = 0.70$), reflecting higher variability among human raters.

\section{Experiments}
\label{sec:experiments}
\begin{table}[t]
    \centering
    \renewcommand{\arraystretch}{1.1} 
    \tiny
    \setlength{\tabcolsep}{2.2pt} 

    \definecolor{highlightGreen}{HTML}{D4F4E9}

    \begin{tabular}{@{}lcccc c cccc}
        \toprule
            & \multicolumn{4}{c}{\textbf{QMSum}} 
            & \multicolumn{4}{c}{\textbf{FAME [EN]}} \\
        \cmidrule(lr){2-5} \cmidrule(lr){6-10}
            & \makecell{GPT\\4o} & \makecell{Gemini\\1.5 pro} & \makecell{\framework{}\\GPT-4o} 
            &  & \makecell{GPT\\4o} & \makecell{Gemini\\1.5 pro} & \makecell{\framework{}\\GPT-4o}  \\
        \midrule
        \rowcolor{gray!20} 
        \multicolumn{10}{c}{\textbf{MESA} (lower is better)} \\
        \midrule
        Coreference
            & \cellcolor{highlightGreen}$0_{\textit{1.22}}$ 
            & $3_{\textit{1.58}}$
            & \cellcolor{highlightGreen}$0_{\textit{1.64}}$
            & 
            & \cellcolor{highlightGreen}$0_{\textit{1.45}}$
            & $3_{\textit{1.57}}$
            & \cellcolor{highlightGreen}$0_{\textit{1.36}}$ \\
        Hallucination     
            & $3_{\textit{1.22}}$
            & $4_{\textit{2.04}}$
            & \cellcolor{highlightGreen}$1_{\textit{1.75}}$
            & 
            & $4_{\textit{0.98}}$
            & $4_{\textit{1.40}}$
            & \cellcolor{highlightGreen}$1_{\textit{0.72}}$ \\
        Incoherence     
            & $4_{\textit{1.50}}$
            & $4_{\textit{1.09}}$
            & \cellcolor{highlightGreen}$3_{\textit{1.88}}$
            & 
            & $4_{\textit{0.94}}$
            & $4_{\textit{0.72}}$
            & \cellcolor{highlightGreen}$3_{\textit{1.24}}$ \\
        Irrelevance     
            & $2_{\textit{1.70}}$
            & $3_{\textit{1.32}}$
            & \cellcolor{highlightGreen}$1_{\textit{1.45}}$
            & 
            & $3_{\textit{1.14}}$
            & $3_{\textit{1.07}}$
            & \cellcolor{highlightGreen}$1_{\textit{1.51}}$ \\
        Language     
            & \cellcolor{highlightGreen}$1_{\textit{1.30}}$
            & $2_{\textit{1.44}}$
            & \cellcolor{highlightGreen}$1_{\textit{1.40}}$
            & 
            & $1_{\textit{1.17}}$
            & $1_{\textit{1.20}}$
            & $1_{\textit{1.11}}$ \\
        Omission     
            & $3_{\textit{0.40}}$
            & $3_{\textit{0.38}}$
            & \cellcolor{highlightGreen}$1_{\textit{0.16}}$
            & 
            & $4_{\textit{0.16}}$
            & $4_{\textit{0.31}}$
            & \cellcolor{highlightGreen}$1_{\textit{0.00}}$ \\
        Repetition     
            & $4_{\textit{1.05}}$
            & $3_{\textit{0.98}}$
            & \cellcolor{highlightGreen}$1_{\textit{1.23}}$
            & 
            & $4_{\textit{0.74}}$
            & $4_{\textit{0.44}}$
            & \cellcolor{highlightGreen}$2_{\textit{0.53}}$ \\
        Structure     
            & $4_{\textit{0.90}}$
            & \cellcolor{highlightGreen}$3_{\textit{1.70}}$
            & \cellcolor{highlightGreen}$3_{\textit{1.24}}$
            & 
            & $3_{\textit{1.57}}$
            & $3_{\textit{1.53}}$
            & $3_{\textit{1.46}}$ \\
        \midrule
        \rowcolor{gray!20} 
        \multicolumn{10}{c}{\textbf{ROUGE (R-1, R-2, R-L) and BERTScore (BS)} (higher is better)} \\
        \midrule
        R-1
            & $37.73_{\textit{5.85}}$
            & $39.61_{\textit{7.21}}$
            & $22.89_{\textit{5.80}}$
            & 
            & $39.68_{\textit{5.73}}$
            & $38.82_{\textit{5.79}}$
            & $20.09_{\textit{4.13}}$ \\
        R-2
            & $7.95_{\textit{4.18}}$
            & \cellcolor{highlightGreen}$11.10_{\textit{4.86}}$
            & $4.13_{\textit{2.44}}$
            & 
            & $8.43_{\textit{3.45}}$
            & \cellcolor{highlightGreen}$8.96_{\textit{3.50}}$
            & $3.81_{\textit{2.40}}$ \\
        R-L
            & $21.39_{\textit{4.05}}$
            & \cellcolor{highlightGreen}$27.55_{\textit{6.36}}$
            & $20.78_{\textit{5.23}}$
            & 
            & \cellcolor{highlightGreen}$29.98_{\textit{4.88}}$
            & $27.81_{\textit{4.18}}$
            & $18.26_{\textit{3.81}}$ \\
        BS (F1)
            & $81.61_{\textit{2.87}}$
            & $80.64_{\textit{3.66}}$
            & \cellcolor{highlightGreen}$85.67_{\textit{1.19}}$
            & 
            & $63.80_{\textit{3.11}}$
            & $83.66_{\textit{2.49}}$
            & \cellcolor{highlightGreen}$84.63_{\textit{1.02}}$ \\
        \bottomrule
    \end{tabular}
    \caption{Results of general summarization of QMSum and FAME. GPT and Gemini results stem from \citet{KirsteinKWR25}. Values are Median$_{Std}$. MESA scores are 1--5 Likert ratings, 
    ROUGE (R-1/R-2/R-L) and BERTScore (BS) are 0--100. \colorbox{highlightGreen}{Green} is best in category.}
    \label{tab:summary_performance}
\end{table}

We evaluate the effectiveness of \framework{} and \reasoning{} in both general and personalized abstractive meeting summarization.
Our experiments are aimed at answering two questions:
(1) What impact does fact extraction and treating summarization as an enrichment task have on summary quality? 
(2) How does our guided, structured reason-out-loud protocol compare to prompt-only personalization?

\subsection{Experimental Setup}
\label{subsec:experimental_setup}

\paragraph{Backbone models.}
We implement \framework{} with GPT \citep{OpenAIAAA24} across all pipeline stages, using its 128k-token context window.
In \Cref{sec:ablations} 
we confirm improvements also hold across Gemini, Llama 3.1 8b, and Gemma 3 4b, with verification modules becoming more relevant, indicating that the gains stem from the fact-based approach.

\paragraph{Baselines.}
We compare \framework{} to GPT and Gemini using zero-shot prompting, tasking 250-token summaries \citep{KirsteinLG25a}.
We exclude refinement-based methods that rely on reusing the same backbone models \cite{KirsteinLG25a}. 
In \Cref{app:self-refinement-comparison}, we show that \framework{} can outperform a three-time revision.
For personalization, we compare \framework{} with and without \reasoning{} to GPT and Gemini, both prompted to role-play. 
Prompts are given in \Cref{app:prompt_templates}.

\paragraph{Datasets.}
We evaluate on QMSum \cite{ZhongYYZ21h}, an established benchmark combining academic (ICSI, \citet{JaninBEE03}), product (AMI, \citet{CarlettaABF06}), and parliamentary (Welsh/Canadian, WPCP) meetings.
We also use FAME \cite{KirsteinKWR25}, a synthetic dataset comprising 500 English and 300 German meetings, spanning 14 different meeting formats and 28 topics, generated by agents simulating realistic meeting dynamics.
We randomly sample 50 English samples from QMSum and FAME to test \framework{} and \reasoning{}.
Dataset details are given in \Cref{sec:dataset_characteristics}.

\begin{table}[t]
    \centering
    \renewcommand{\arraystretch}{1.1} 
    \tiny
    \setlength{\tabcolsep}{1.5pt} 

    \definecolor{highlightGreen}{HTML}{D4F4E9}

    \begin{tabular}{@{}l@{\hspace{3pt}}cccc c cccc}
        \toprule
            & \multicolumn{4}{c}{\textbf{QMSum}} 
            & \multicolumn{1}{c}{}
            & \multicolumn{4}{c}{\textbf{FAME [EN]}} \\
        \cmidrule(lr){2-5} \cmidrule(lr){7-10}
            & \makecell{GPT\\4o} & \makecell{Gemini\\1.5 pro} & \makecell{\framework{}\\ RP}  & \makecell{\framework{}\\ \reasoning{}}
            &  & \makecell{GPT\\4o} & \makecell{Gemini\\1.5 pro} & \makecell{\framework{}\\ RP}  & \makecell{\framework{}\\ \reasoning{}} \\
        \midrule
        \rowcolor{gray!20} 
        \multicolumn{10}{c}{\textbf{P-MESA} (lower is better)} \\
        \midrule
        goal alignment
            & $3_{\textit{0.80}}$ 
            & $3_{\textit{0.54}}$
            & $3_{\textit{0.79}}$
            & \cellcolor{highlightGreen}$2_{\textit{0.48}}$
            & 
            & $3_{\textit{0.53}}$
            & $3_{\textit{0.58}}$
            & $3_{\textit{0.67}}$
            & \cellcolor{highlightGreen}$2_{\textit{0.54}}$ \\
        completeness     
            & $4_{\textit{0.77}}$
            & $4_{\textit{0.88}}$
            & \cellcolor{highlightGreen}$3_{\textit{0.64}}$
            & \cellcolor{highlightGreen}$3_{\textit{0.80}}$
            & 
            & $4_{\textit{0.89}}$
            & $4_{\textit{0.59}}$
            & \cellcolor{highlightGreen}$2_{\textit{0.61}}$
            & \cellcolor{highlightGreen}$2_{\textit{0.59}}$ \\
        factuality     
            & $3_{\textit{1.44}}$
            & $4_{\textit{1.22}}$
            & \cellcolor{highlightGreen}$2_{\textit{0.37}}$
            & \cellcolor{highlightGreen}$2_{\textit{0.25}}$
            & 
            & $3_{\textit{1.36}}$
            & $4_{\textit{1.02}}$
            & \cellcolor{highlightGreen}$2_{\textit{0.34}}$
            & \cellcolor{highlightGreen}$2_{\textit{0.22}}$ \\
        knowledge level fit     
            & $2_{\textit{1.23}}$
            & $2_{\textit{0.44}}$
            & \cellcolor{highlightGreen}$1_{\textit{0.86}}$
            & \cellcolor{highlightGreen}$1_{\textit{0.86}}$
            & 
            & $2_{\textit{0.97}}$
            & $2_{\textit{0.50}}$
            & $2_{\textit{0.73}}$
            & \cellcolor{highlightGreen}$1_{\textit{0.58}}$ \\
        priority structuring     
            & $4_{\textit{0.48}}$
            & $4_{\textit{0.60}}$
            & \cellcolor{highlightGreen}$3_{\textit{0.56}}$
            & $4_{\textit{0.54}}$
            & 
            & $4_{\textit{0.28}}$
            & $4_{\textit{0.65}}$
            & $3_{\textit{0.41}}$
            & \cellcolor{highlightGreen}$2_{\textit{0.45}}$ \\
        contextual framing     
            & $4_{\textit{0.91}}$
            & $4_{\textit{1.13}}$
            & \cellcolor{highlightGreen}$3_{\textit{0.97}}$
            & \cellcolor{highlightGreen}$3_{\textit{0.88}}$
            & 
            & $4_{\textit{1.21}}$
            & $4_{\textit{0.96}}$
            & \cellcolor{highlightGreen}$3_{\textit{0.69}}$
            & \cellcolor{highlightGreen}$3_{\textit{0.81}}$ \\
        relevance     
            & $3_{\textit{0.54}}$
            & $3_{\textit{0.55}}$
            & \cellcolor{highlightGreen}$1_{\textit{0.27}}$
            & \cellcolor{highlightGreen}$1_{\textit{0.57}}$
            & 
            & $3_{\textit{0.33}}$
            & $3_{\textit{0.47}}$
            & $2_{\textit{0.89}}$
            & \cellcolor{highlightGreen}$1_{\textit{0.33}}$ \\
        \midrule
        \rowcolor{gray!20} 
        \multicolumn{10}{c}{\textbf{MESA} (lower is better)} \\
        \midrule
        Coreference
            & \cellcolor{highlightGreen}$0_{\textit{1.48}}$ 
            & $1_{\textit{1.48}}$
            & $1_{\textit{1.60}}$
            & $1_{\textit{1.55}}$
            & 
            & $1_{\textit{1.44}}$
            & $0_{\textit{1.46}}$
            & \cellcolor{highlightGreen}$0_{\textit{1.44}}$
            & \cellcolor{highlightGreen}$0_{\textit{1.55}}$ \\
        Hallucination     
            & $4_{\textit{1.79}}$
            & $3_{\textit{2.03}}$
            & $3_{\textit{1.03}}$
            & \cellcolor{highlightGreen}$2_{\textit{1.33}}$
            & 
            & $3_{\textit{1.39}}$
            & $3_{\textit{1.73}}$
            & $2_{\textit{0.89}}$
            & \cellcolor{highlightGreen}$1_{\textit{1.19}}$ \\
        Incoherence     
            & \cellcolor{highlightGreen}$3_{\textit{1.74}}$
            & $4_{\textit{1.38}}$
            & \cellcolor{highlightGreen}$3_{\textit{1.38}}$
            & \cellcolor{highlightGreen}$3_{\textit{1.09}}$
            & 
            & $3_{\textit{1.60}}$
            & $3_{\textit{1.37}}$
            & $3_{\textit{0.89}}$
            & $3_{\textit{0.75}}$ \\
        Irrelevance     
            & $2_{\textit{1.19}}$
            & $2_{\textit{1.10}}$
            & $1.5_{\textit{1.42}}$
            & \cellcolor{highlightGreen}$1_{\textit{1.48}}$
            & 
            & $3_{\textit{1.33}}$
            & $2_{\textit{1.58}}$
            & \cellcolor{highlightGreen}$1_{\textit{1.40}}$
            & \cellcolor{highlightGreen}$1_{\textit{1.23}}$ \\
        Language     
            & $2_{\textit{1.38}}$
            & $2_{\textit{1.32}}$
            & $2_{\textit{1.38}}$
            & $2_{\textit{1.24}}$
            & 
            & $1_{\textit{1.04}}$
            & $1_{\textit{1.14}}$
            & $1_{\textit{1.25}}$
            & $1_{\textit{1.10}}$ \\
        Omission     
            & $4_{\textit{0.16}}$
            & $4_{\textit{0.41}}$
            & \cellcolor{highlightGreen}$2_{\textit{0.29}}$
            & \cellcolor{highlightGreen}$2_{\textit{0.40}}$
            & 
            & $4_{\textit{0.41}}$
            & $4_{\textit{0.43}}$
            & $2_{\textit{0.28}}$
            & \cellcolor{highlightGreen}$1_{\textit{0.56}}$ \\
        Repetition     
            & $3_{\textit{0.94}}$
            & $3_{\textit{1.22}}$
            & \cellcolor{highlightGreen}$1.5_{\textit{1.17}}$
            & $2_{\textit{0.79}}$
            & 
            & $4_{\textit{0.31}}$
            & $3_{\textit{0.50}}$
            & \cellcolor{highlightGreen}$2_{\textit{0.41}}$
            & \cellcolor{highlightGreen}$2_{\textit{0.43}}$ \\
        Structure     
            & $3_{\textit{1.67}}$
            & \cellcolor{highlightGreen}$1_{\textit{1.42}}$
            & $3_{\textit{1.45}}$
            & $3_{\textit{1.38}}$
            & 
            & \cellcolor{highlightGreen}$2_{\textit{1.28}}$
            & $3_{\textit{0.47}}$
            & $3_{\textit{1.56}}$
            & \cellcolor{highlightGreen}$2_{\textit{1.21}}$ \\
        \bottomrule
    \end{tabular}
    \caption{Personalized summarization of QMSum and FAME. Values are Median$_{Std}$. MESA and \pmetric{} scores are 1--5 Likert ratings. \colorbox{highlightGreen}{Green} is best in category.}
    \label{tab:summary_performance_personal}
\end{table}

\paragraph{Evaluation.}
We evaluate summaries using ROUGE (R-1/R-2/R-L) \cite{Lin04} and BERTScore (rescaled F1) \cite{ZhangKWW20}, the reference-free MESA \cite{KirsteinLG25a} for analyzing the occurrence of eight general, and our \pmetric{} (\Cref{sec:pmesa}) to capture seven personalization dimensions.
General QMSum and FAME baselines stem from \citet{KirsteinKWR25}.

\subsection{Results: General Summarization}
\label{subsec:results_general_summ}

\paragraph{Findings.}
Reframing summarization as an enrichment task yields three benefits: (\emph{F1}) improved content understanding via fact isolation, (\emph{F2}) less hallucination due to grounded claims, and (\emph{F3}) better coherence through summary planning.

\paragraph{Quantitative analysis.}
\Cref{tab:summary_performance} shows that \framework{} consistently outperforms baselines on MESA.
However, \framework{}'s summary structure appears to diverge from the references (lower ROUGE).
The largest quality impact reduction appears in hallucination, dropping from 3→1 on QMSum and 4→1 on FAME for \framework{} (\emph{F2}).
MESA's chain-of-thought traces confirm these reductions are due to fewer unsupported claims.
We conclude that this relates to the \texttt{Fact Verification} and \texttt{Outline Planning} stages that constrain synthesis to grounded facts.

Omission and irrelevance scores drop by $\sim$2 points to 1.
We interpret that the fact-based approach helps with content understanding and enables \texttt{Fact Selection} to focus on salience without omission (\emph{F1}).
\texttt{Fact Grouping} cuts repetition (QMSum: 3→1, FAME: 4→2), yielding denser summaries.
Structure improves (4→3), reflecting the effect of \texttt{Outline Planning} (\emph{F3}).
Language scores remain stable, indicating no loss of fluency.

\paragraph{Qualitative observations.}
Baselines tend to mirror the temporal structure of a meeting and to preserve low-value information (e.g., ``Alice shared her screen…'').
\framework{} decouples summary structure from chronology, shifting to thematic progressions, and filters irrelevant content.
We link this to the \texttt{Relevance Scoring} and \texttt{Fact Grouping} stages.
A qualitative example appears in \Cref{app:qualitative_example}.

\subsection{Results: Personalized Summarization}
\label{sec:personalized_summarization}

\paragraph{Findings.}
\reasoning{} leads to (\emph{F1}) more personalized fact selection, (\emph{F2}) reduced hallucination via explicit user modeling, and (\emph{F3}) personalized adaptation without quality loss.

\paragraph{Quantitative analysis.}
\Cref{tab:summary_performance_personal} shows that adding \reasoning{} to \framework{} results in improvements on all seven \pmetric{} dimensions.
Relevance and knowledge fit drop from 2→1, meaning content better matches user expectations and language suited to their expertise (\emph{F1}).
Goal alignment and priority structuring improve from 3→2, indicating that summaries better reflect the information needs from readers (\emph{F1}).
Baselines with persona injection lag by 1–2 points across \pmetric{} criteria (\emph{F1} and \emph{F2}).
We find that \reasoning{} poses questions the model can answer reliably while role-playing, allowing the responses to function as a pre-selection mechanism.

Further, these gains do not trade off against general quality.
MESA scores for personalized \framework{} summaries are close to non-personalized scores (\Cref{subsec:results_general_summ}).
\reasoning{} reduces hallucination and omission scores (3→2, 2→1) (\emph{F3}).

\paragraph{Qualitative observations.} 

As we can observe in the example in \Cref{app:qualitative_example}, \reasoning{} addresses two weaknesses of single-prompt baselines. 
First, it reduces \emph{oversimplification} by varying granularity in line with the reader’s priorities while baseline models indiscriminately compress detail, \reasoning{} (\emph{F1}).
Second, it reduces \emph{profile hallucination} that skews content selection (\emph{F2}). 
\reasoning{}'s structured reflection on the persona and input acts as working memory for the model to ground decisions.
We conclude that this reusable working memory and reasoning outperform static persona injection.

\begin{table}[ht]
\scriptsize
\centering
\begin{tabular}{lccc}
\toprule

\textbf{Approach} & \textbf{Reader-Tailoring} & \textbf{Roleplaying} & \textbf{\reasoning{}} \\
\midrule
\rowcolor{gray!20} 
\multicolumn{4}{c}{\textbf{Huamn annotation following P-MESA criteria} (lower is better)} \\
\midrule
        goal alignment
            & $3.5_{\textit{0.70}}$
            & $2.5_{\textit{0.98}}$ 
            & \cellcolor{highlightGreen}$2_{\textit{0.81}}$ \\
        completeness     
            & $4.5_{\textit{0.88}}$
            & \cellcolor{highlightGreen}$3_{\textit{1.58}}$
            & \cellcolor{highlightGreen}$3_{\textit{0.65}}$ \\
        factuality
            & $2.5_{\textit{1.38}}$
            & $3.5_{\textit{1.58}}$
            & \cellcolor{highlightGreen}$2_{\textit{0.77}}$ \\
        knowledge level fit     
            & $2.5_{\textit{1.40}}$
            & $3_{\textit{1.53}}$
            & \cellcolor{highlightGreen}$1_{\textit{1.73}}$ \\
        priority structuring     
            & $3_{\textit{0.26}}$
            & $3_{\textit{0.30}}$
            & $3_{\textit{0.28}}$ \\  
        contextual framing     
            & $4.5_{\textit{1.35}}$
            & $4_{\textit{1.07}}$
            & \cellcolor{highlightGreen}$3_{\textit{0.70}}$ \\
        relevance     
            & $3.5_{\textit{0.48}}$
            & \cellcolor{highlightGreen}$2.5_{\textit{0.86}}$
            & \cellcolor{highlightGreen}$2.5_{\textit{0.40}}$ \\
\midrule
\rowcolor{gray!20} 
\multicolumn{4}{c}{\textbf{Ranking} (1-3, median, lower is better)} \\
\midrule
Ranking     & $3_{\textit{0.49}}$ & $2_{\textit{0.75}}$ & $1_{\textit{0.40}}$ \\
\bottomrule
\end{tabular}
\caption{Human evaluation of summaries generated through reader-tailoring, roleplaying, and SCOPE. Annotators rate summaries on the P-MESA criteria and rank the summaries. \colorbox{highlightGreen}{Green} is best in category.}
\label{tab:human_annotation_pmesa_ranking}
\end{table}


\paragraph{Human Assessment.}
To confirm that human readers perceive the performance gains of \reasoning{}, we conduct a comparative human evaluation.
We task five annotators (three male, two female; ages 22-29; C1+ English) with assessing personalized summaries generated by \reasoning{} and the two baselines, reader-tailoring and roleplaying.
We created 20 evaluation sets, each corresponding to a unique source transcript and a specific reader profile.
Each set contains three summaries, i.e., one from reader-Tailoring, roleplaying, and \reasoning{}.
All annotators evaluate all 20 sets, with the three summaries within each set presented in a randomized order to mitigate presentation bias.
The annotators perform two tasks: (1) rating each summary on the seven \pmetric{} criteria (1-5 Likert scale) and (2) ranking the three summaries from best (1) to worst (3).
The whole experimental setup follows the details in \Cref{app:human_evaluation}.
We find substantial inter-annotator agreement on the rating task (Krippendorff's $\alpha=$0.71).

The results, presented in \Cref{tab:human_annotation_pmesa_ranking}, confirm that human judges consistently prefer summaries from \reasoning{}.
It achieves the best median rank of 1 and is rated highest on five of the seven \pmetric{} criteria, with strong improvements in factuality and knowledge level fit.
This indicates that readers prefer the improved grounding and tailored language.
Interestingly, scores for priority structuring were identical across all methods, suggesting this was a less decisive factor for annotators.
In one instance, a summary from reader-tailoring was ranked highest due to a preference for its presentation, highlighting the subjective nature of structural choices in summarization.

\section{Ablations}
\label{sec:ablations}

\paragraph{Safety mechanisms.}
We assess the \texttt{Fact Verification} stage, finding only 8 in 150 facts required revision (1 false positive, 0 false negatives).
Removing this component causes no measurable quality drop with GPT, suggesting it serves as a strategic guardrail for less capable models.
Similarly, removing \texttt{summary refinement} only increases incoherence slightly (MESA: 3→3.5) without affecting other dimensions.
\Cref{app:fact_verification_impact,app:summary_refinement_analysis} provide detailed analysis.

\paragraph{Generalization.}
\framework{} shows robust performance across diverse models (Gemini 1.5 Pro, Llama 3.1 8b, Gemma 3 4b), with consistent improvements over single-LLM baselines.
Notably, \framework{} narrows the gap between commercial and open-source models on MESA dimensions.
When applied to general text summarization benchmarks, \framework{} yields an average MESA improvement of over 1 point per dimension, with the strongest and most consistent gains coming from repetition (-3) reduction and frequent large gains in omission.
This confirms that the improvements stem from fact-based reasoning, rather than specific model capabilities.
To further test generalization on domain-specific conversational data, we use the meeting simulation approach, MIMIC \cite{KirsteinKWR25}, to transform scientific articles from the PubMed dataset \cite{xiong2024benchmarking} into realistic meeting transcripts.
Applying \framework{} to summarize these simulated meetings yields gains in maintaining factual consistency, with improvements in hallucination (3 → 1), repetition (-2.5 points), and language (-2 points). 
See \Cref{sec:app_cross_model_performance,sec:app_cross_domain_performance} for complete evaluation.

\paragraph{Architecture.}
To justify our pipeline's seven-step modular design, we evaluated three compressed variants against our full framework: 
a single prompt performing all steps (\texttt{combined-1}), merging Fact Extraction and Relevance Scoring (\texttt{combined-2}, and combining Relevance Scoring and Outline Planning (\texttt{combined-3}). 
The results in \Cref{tab:stage-combination} show that collapsing stages causes a degradation in content fidelity.
All three combined variants score a 4 in omission, a 3-point drop from \framework{}, indicating they consistently miss key information.
Hallucination increases, with \texttt{combined-2} scoring a 5.
We conclude that forcing the model to extract and evaluate facts simultaneously prevents it from properly understanding the content and grounding claims.
While metric dimensions such as Structure and Incoherence show less degradation, the failure on Factuality demonstrates the necessity of our distinct pipeline stages for producing reliable summaries.

\section{Final Considerations}
\label{sec:final_considerations}
We introduced \framework{}, a fact-based pipeline treating abstractive meeting summarization as an enrichment task.
\framework{} tackled the core challenges of meetings, i.e., information distribution, contextual dependencies, and salience ambiguity, while reducing hallucinations (3→1), irrelevance (2→1), and repetition (3→1).
To handle personalization, we introduced \reasoning{}. 
\reasoning{} builds on a \textit{reason-out-loud} protocol, guiding a model through nine questions to construct an interpretation trace of the target reader and content before filtering.
This improved relevance (3→1), alignment with user goals (3→2), and knowledge level fit (2→1).
We further contributed \pmetric{}, an LLM-based personalization quality metric, which assesses the reader fit across seven criteria, offering interpretable scores and strong alignment with human judgment ($\rho \geq$70\%).

Our contributions advance multiple research directions in summarization, from fact-level content control enabling fine-grained adaptability for cross-document summarization, to robust salience detection through structured extraction and verification, to reference-free evaluation of reader-specific appropriateness.
By releasing \framework{}, \reasoning{}, and \pmetric{} as open-source, we provide a powerful toolkit that researchers can extend to multilingual applications, multi-source integration, and domains beyond meeting contexts.
This work addresses a persistent gap in summarization approaches that lack the ability to comprehend content before summarization, fostering advancements in summarization and personalization.

\section*{Limitations}
\label{sec:limitations}
The performance of \framework{} and \reasoning{} depends significantly on the capabilities of the underlying language model.
While our implementation uses GPT-4o, models with different reasoning capabilities or smaller context windows may produce less accurate fact identification or inferior reasoning traces, potentially reducing the quality of both general and personalized summaries.
Our ablation studies suggest that while performance decreases with less capable models, the core benefits of our fact-based paradigm remain intact, demonstrating the framework's architectural robustness.

The evaluation datasets, although comprehensive, cover only specific meeting types.
QMSum encompasses academic, business, and parliamentary meetings, while FAME provides synthetic meetings across 14 formats.
Other corpora exist (e.g., MeetingBank \cite{HuGDD23a}, ELITR \cite{NedoluzhkoSHG22}) but closely resemble QMSum’s formal institutional settings and would not increase the variety.
While we selected these datasets for their broadest coverage among publicly available datasets, they do not capture all real-world meeting dynamics, particularly those in specialized domains (e.g., medical consultations, legal) that may require domain-specific fact extraction patterns.

The computational requirements of \framework{} represent another limitation. 
The multi-stage pipeline, particularly the fact extraction and verification stages, incurs both increased inference time and higher computational costs compared to end-to-end summarization approaches.
While quality improvements justify this trade-off, it may limit deployment in resource-constrained environments or real-time applications.
As we demonstrate in our ablation study, downscaling the backbone model to reduce costs can still yield higher-quality summaries than those produced by end-to-end summarization.

\section*{Acknowledgements}
This work was supported by the Lower Saxony Ministry of Science and Culture and the VW Foundation.
Frederic Kirstein was supported by Mercedes-Benz AG Research and Development.

\bibliography{25_EMNLP_AtomicFacts}





\appendix

\section*{Organisation of Appendix}
This appendix provides further details and ablations for our paper.
\Cref{app:sec_open_resources,app:experimental_setup} document reproducibility information and experimental setup.
\Cref{app:human_evaluation,sec:computational_efficiency} detail our human evaluation protocols and computational efficiency.
\Cref{sec:empirical_observations,sec:ablation_studies,sec:extended_analysis} contain empirical observations and ablation studies demonstrating component contributions.
\Cref{sec:dataset_characteristics,sec:reasoning_analysis} provide dataset characteristics and SCOPE component details.
\Cref{app:pmetric_lit_review,sec:appendix_b-acc,app:self-refinement-comparison,app:qualitative_example} cover evaluation methodology, definitions, and qualitative examples.
Researchers primarily interested in reimplementing our work should focus on \Cref{app:sec_open_resources,app:experimental_setup}, and the prompt templates in Figures \ref{fig:atomic_facts_prompt}--\ref{fig:single_llm_personalized_prompt}.
Those evaluating methodology rigor should examine \Cref{app:human_evaluation,app:pmetric_lit_review,sec:appendix_b-acc}.

\section{Open Resources \& Licensing}
\label{app:sec_open_resources}

\subsection{Repository \& License}
\label{app:repository}
The \framework{} framework proposed in this paper, along with the \reasoning{} reasoning protocol and \pmetric{} metric, are available on \href{https://github.com/FKIRSTE/emnlp2025-reframe-summarization}{GitHub} under an MIT license.
The repository includes implementation code, evaluation scripts, prompt templates, and configuration files necessary for reproducing our results.

\subsection{Datasets \& Licensing}
\label{subsec:datasets_licensing}
In \Cref{tab:dataset_licensing}, we report the licensing and a high-level overview of the QMSum \cite{ZhongYYZ21h} and FAME \cite{KirsteinKWR25} datasets.
For the FAME dataset, which is not publicly available at the time of writing, we obtained research access by contacting the authors directly.
The authors have indicated plans for public release under CC BY-SA 4.0 licensing.
Researchers seeking to replicate our results can either access the dataset when published or contact the authors using the information provided in \cite{KirsteinKWR25}.
A detailed overview of dataset characteristics is given in \Cref{sec:dataset_characteristics}.

\begin{table}[ht]
    \centering
    \renewcommand{\arraystretch}{1.0}
    \scriptsize
    \setlength{\tabcolsep}{3.5pt}
\begin{tabular}{lccc>{\raggedright\arraybackslash}p{2.2cm}} 
\toprule
\rowcolor{gray!20}
\textbf{Dataset} & \textbf{License} & \textbf{Size} & \textbf{Avg. Length} & \textbf{Domain} \\
\midrule
QMSum & MIT & 232 & 7303 & academic, council, design-meeting \\
FAME & CC BY-SA 4.0 & 800 & 6250 & various, e.g., sports, technology, history, math, philosophy \\
\bottomrule
\end{tabular}
\caption{Dataset licensing and high-level overview.}
\label{tab:dataset_licensing}
\end{table}


\section{Experimental Setup Details}
\label{app:experimental_setup}
This section provides implementation details, including all prompt templates, hyperparameters, and computational infrastructure used in all experiments.
We document the exact methodology to ensure reproducibility of \framework{} and \reasoning{}, including parameters that were empirically determined.

\subsection{Implementation Details}
We address two key challenges in processing lengthy meeting transcripts: managing context limitations and maintaining information coherence across processing stages.
As we use an Azure GPT instance with a capped 4K-token output, we developed two specialized 
components to handle meetings with facts exceeding this limit:

A \textbf{Chunk Processor} divides transcripts into sequential chunks based on GPT-2 tokenizer estimation \citep{Radford19}.
Each chunk maintains access to the previous chunk to preserve cross-boundary information.
We employ dynamic chunk sizing, prioritizing complete speaker turns over fixed token counts while respecting maximum context limitations.

A \textbf{Memory Bank} provides a centralized fact repository that decouples extraction from downstream processing.
To address cross-chunk fact redundancy, we apply a lightweight text similarity function that combines character-level sequence alignment and token-level word overlap to merge contexts with $\geq 70\%$ similarity while preserving higher relevance scores.

\framework{}'s modular design further allows adaptation to varying LLM context capacities, enabling integration with both smaller open-source models and commercial API-based services.

\subsection{Model Specs}
\label{app:model_specs}

We employ four different language models in this work.
\Cref{app:model_specs} provides an overview of publicly disclosed information about these models.

\begin{table}[ht]
\small
\centering
\begin{tabular}{llll}
\toprule
\rowcolor{gray!20}
\textbf{Model} & \textbf{Version} & \textbf{Parameters} & \textbf{Provider} \\
\midrule
GPT & 4o, 2024-08-06 & $\sim$200B & OpenAI \\
Gemini & 1.5-pro-002 & not disclosed & Google \\
Llama & 3.1 & $\sim$8B & Meta AI \\
Gemma & 3 & $\sim$4B & Google \\
\bottomrule
\end{tabular}
\caption{Model specifications as reported in original papers and disclosing works. GPT and Gemini are our main models for experiments (\Cref{sec:experiments}), Llama and Gemma for ablations \Cref{sec:app_cross_model_performance}.}
\label{tab:model_specs}
\end{table}

\subsection{Prompt Templates}
\label{app:prompt_templates}
The \framework{} pipeline and \reasoning{} protocol rely on carefully crafted prompt templates for each processing stage.
Figures \ref{fig:atomic_facts_prompt}--\ref{fig:single_llm_personalized_prompt} present the complete set of prompts used throughout our experiments.

\textbf{Core \framework{} Prompts} (Figures \ref{fig:atomic_facts_prompt}--\ref{fig:summary_refinement_prompt}):
These prompts guide the four main stages of our general summarization pipeline. Stage 1 prompts (\Cref{fig:atomic_facts_prompt,fig:fact_verification_prompt}) handle fact extraction and verification.
Stage 2 prompts (\Cref{fig:feature_extraction_prompt}) focus on relevance scoring and fact grouping. 
Stage 3 prompts (\Cref{fig:outline_generation_prompt}) guide outline planning, while Stage 4 prompts (Figures \ref{fig:summary_generation_prompt}--\ref{fig:summary_refinement_prompt}) manage summary generation and quality assurance.

\textbf{\reasoning{} Protocol Prompts} (Figures \ref{fig:persona_filter_prompt}--\ref{fig:persona_summary_validation_prompt}): These prompts implement our personalization approach. They guide the model through explicit reasoning about reader preferences (Figure \ref{fig:persona_filter_prompt}), fact selection based on persona relevance ( \Cref{fig:persona_salient_feature_prompt_part1,fig:persona_salient_feature_prompt_part2}), and persona-focused outline and summary generation (Figures \ref{fig:persona_outline_prompt}--\ref{fig:persona_summary_validation_prompt}).

\textbf{Baseline Comparison Prompt} (Figure \ref{fig:single_llm_personalized_prompt}): This prompt enables direct comparison with single-LLM personalization approaches by implementing reader-tailoring in a one-shot generation task.

All prompts follow a consistent structure with clearly defined input and output formats, explicit instructions, and constraints that guide the model toward the desired behavior. We designed these prompts iteratively through systematic experimentation, refining each prompt to maximize effectiveness while maintaining reproducibility.

\subsection{Hyperparameters}
\label{app:hyperparameters}

For our experiments, we overly use default values for key hyperparameters, i.e., \texttt{top-p} = 1.0, \texttt{frequency penalty} = 0.0, \texttt{presence penalty} = 0.0.
We empirically chose \texttt{temperature} = 0.1 to have the model behave more focused and deterministic.
All values are fixed across the different model backbones used (\Cref{app:model_specs}).


\section{Human Evaluation Protocols}
\label{app:human_evaluation}

This section details our human evaluation methodology, which undergirds both the development of our \pmetric{} metric and the assessment of the suitability of \pmetric{} as a proxy for human annotations (see \Cref{sec:pmesa}).

\subsection{Annotator Recruitment \& Demographics}

We have an annotation team of six participants (three male, three female, ages 22-29) through a structured recruitment process.
All annotators were employed as research assistants or doctoral candidates with standardized contracts.
We selected annotators based on their availability to complete tasks without time pressure, demonstrated English proficiency (native speakers or C1-C2 certified), and academic background relevant to text analysis.
This selection process yielded a team with diverse disciplinary perspectives: two computer science students, three psychology students, and one communication science student.
All annotators provided explicit consent for their anonymized annotations to be used in this research, and the entire annotation protocol received approval from our institution's ethics committee before implementation.

\subsection{Training \& Quality-Control}
\paragraph{Preparation:}
We have prepared a comprehensive handbook for our annotators, detailing the project context and defining the criteria (a short version is presented in \Cref{tab:p-mesa_criteria} and an extended version with more details).
Each definition includes two examples: one with minimal impact on quality and one with high impact.
The handbook explains the 1 - 5 Likert rating for the individual questionnaires.
The handbook does not specify an order for processing the items.
We provide the handbook in English and the annotators' native languages, using professional translations.

We structured our timeline as a four-week process: one week dedicated to onboarding, followed by three weeks for primary annotation.
The first annotation week featured twice-weekly check-ins, which transitioned to weekly meetings for the subsequent periods.
In parallel, the research team conducted quality assessments without the annotator's presence weekly to identify emerging issues
(Note: week refers to a regular working week.)

\paragraph{Onboarding:}
The onboarding week is dedicated to getting to know the project and familiarizing oneself with the definitions and data.
We begin with a kick-off meeting to introduce the project and explain the handbook, particularly focusing on each definition.
We generate ten additional samples for the individual tasks following the respective approaches to familiarize.
After processing the first five samples, we hold individual meetings to clarify any confusion.
The remaining five samples are then annotated to confirm clarity. 
A second group meeting this week addresses any misaligned understanding among the reviewers.
After the group meeting, we meet individually with the annotators to review their work and ensure their quality and understanding of the task and samples. 
Judging from the reasoning they provide for each decision and annotation, all annotators demonstrate reliable performance and good comprehension of the task and definitions.

\paragraph{Annotation Process:}
For the primary annotation workflow, we distributed the workload equally among annotators with distinct approaches for the two evaluation phases.
During the \pmetric{} development phase, each annotator evaluated $\sim$17 samples, with every sample receiving annotations from two independent evaluators.
For the \pmetric{} validation, each annotator assessed $\sim$8 samples, with every sample being evaluated by four distinct annotators to ensure robust reliability assessment.

To maintain unbiased evaluation, annotators remained blind to the summarization architecture, which generated each summary, the source dataset, and other annotators' ratings.
We randomized the sample presentation order for each annotator to mitigate positional bias.
They are given a week to complete their set at their own pace and with their break times. 
Quiet working rooms were provided if needed for concentration.
Annotators can choose their annotation order for each sample and are allowed to revisit previous samples.

Regular meetings are held to address any emerging issues or questions on definitions. 
During the quality checks the authors perform, we look for incomplete annotations, missing explanations, and signs of misunderstanding based on the provided reasoning. 
If the authors find such a lack of quality, the respective annotators will be notified to redo the annotation.
At halftime of the annotation cycle, we compute inter-annotator agreement scores.
If we observed a significant difference among annotators, we planned a dedicated meeting with all annotators and a senior annotator to discuss such cases. 
On average, annotators spend 25 minutes per sample.

\paragraph{Handling of unexpected cases:}
Given that our annotators have other commitments, we anticipate potential scheduling conflicts. 
We allow flexibility for annotators to complete their samples beyond the week limit if needed, reserving an additional week as a buffer. 
Despite these provisions, all annotators complete their assigned samples within the original weekly timeframes. 
We further allow faster annotators to continue with an additional sample set.
This additional work was voluntary.

\subsection{Inter-Annotator Agreement Formulas}
We assessed annotation reliability using Krippendorff's alpha ($\alpha$), which we selected for its ability to accommodate multiple annotators, ordinal data, and handle missing values—characteristics that make it well-suited for analyzing Likert-scale ratings across multiple dimensions. 
This metric ranges from 0 (agreement attributable to chance) to 1 (perfect agreement), with established thresholds in computational linguistics literature for interpreting reliability strength.

As shown in Table \ref{tab:krippendorffs_alpha_human_annotation}, all dimensions achieved substantial to strong agreement.
The Factuality dimension demonstrated the highest consistency ($\alpha = 0.839$), likely due to its more concrete definition and readily observable manifestations in summary text.
The relatively lower agreement on Knowledge Level Fit ($\alpha = 0.681$) reflects the inherent subjectivity in assessing information prioritization, though it still comfortably exceeds the threshold for substantial reliability.
These strong reliability indicators validate our annotation protocol and suggest that \pmetric{} criteria are consistently interpretable across different human evaluators.

\begin{table}
  \centering
  \small
  \begin{tabular}{lc}
    \toprule
    \rowcolor{gray!20}
    \textbf{Dimension} & \textbf{Krippendorff's $\alpha$} \\
    \midrule
    Factuality  & 0.839  \\
    Completeness & 0.811  \\
    Relevance &  0.762 \\
    Goal Alignment     & 0.758   \\
    Priority Structuring     & 0.758   \\
    Knowledge Level Fit     & 0.681  \\
    Contextual Framing   & 0.792  \\
    \bottomrule
  \end{tabular}
  \caption{Inter-rater reliability for the human annotations, measured by Krippendorff's alpha. Scores $\geq$0.667 mean moderate agreement, and scores $\geq$0.8 mean strong agreement.}
  \label{tab:krippendorffs_alpha_human_annotation}
\end{table}


\section{Computational Efficiency Analysis}
\label{sec:computational_efficiency}

This section compares the computational requirements of our proposed approaches with those of the baselines.
We measure efficiency through token utilization, execution time, and associated API costs, providing practical considerations for deployment.

\subsection{Token \& API Cost Breakdown}

We calculated token usage by instrumenting all API calls to track input and output tokens across each pipeline component.
API costs were calculated using OpenAI's published GPT-4o pricing ($\$0.01$ per 1K input tokens and $\$0.03$ per 1K output tokens as of August 2024).
Each reported value represents the mean across 50 meeting summarizations from our evaluation dataset, with meetings averaging 7,303 words.

\Cref{tab:token_costs} provides a detailed breakdown of computational requirements across pipeline components and alternative approaches.
The most resource-intensive components of \framework{} are \texttt{Fact Extraction} (24.1K input tokens, 3.8K output tokens) and \texttt{Fact Verification} (26.9K input tokens, 3.8K output tokens), together accounting for $\sim$62\% of the pipeline's total computation. 

\texttt{Fact Verification} can be seen as optional based on our ablation studies (\Cref{sec:ablations}, detailed in \Cref{sec:ablation_studies}), which show minimal quality degradation when it is removed.
This offers a potential 33\% reduction in compute time (59 seconds) and cost reduction of $\$0.07$ per meeting for deployment scenarios with stricter efficiency requirements.

The \reasoning{} personalization protocol adds minimal overhead when integrated with \framework{}, requiring only 3076 additional input tokens and 3179 output tokens while delivering notable personalization improvements as demonstrated in \Cref{subsec:personalization_reasoning}.

\begin{table}[ht]
\tiny
\centering
\begin{tabular}{lllll}
\toprule
\rowcolor{gray!20}
\textbf{Approach} & \makecell{\textbf{Input}\\ \#Tokens} & \makecell{\textbf{Output}\\ \#Tokens} & \makecell{\textbf{Est. Cost}\\ \$/meeting} &  \makecell{\textbf{Time}\\ seconds} \\
\midrule
\framework{}                            & 72,059    & 11,674    & 0.21          & 225 \\
\hspace*{2mm}Fact Extraction            & 24,143.8  & 3,753.5   & 0.06          & 58\\
\hspace*{2mm}Fact Verification            & 26,925.2  & 3,813.0   & 0.07          & 59\\
\hspace*{2mm}Fact Scoring \& Fact Grouping    & 7,287.0   & 3,880.0   & 0.02          & 76\\
\hspace*{2mm}Outline Planning         & 859       & 523.2     & $\leq$0.01    & 7\\
\hspace*{2mm}Enrichment-based Generation            & 12,704.8  & 462.5     & 0.03          & 14\\
\hspace*{2mm}Quality Assurance       & 12,692.0  & 160.5     & 0.03          & 11\\
\midrule
\reasoning{} (additional) & 3076.1 & 3179.2 & 0.03 & 13\\
\midrule
Single LLM & 23,725.5 & 233.0 & 0.06 & 5 \\
\bottomrule
\end{tabular}
\caption{Token usage and estimated API costs for the individual components and approaches considered.}
\label{tab:token_costs}
\end{table}

\subsection{Quality-Cost Trade-off Analysis}
\begin{figure*}[ht]
\centering
\includegraphics[width=\linewidth]{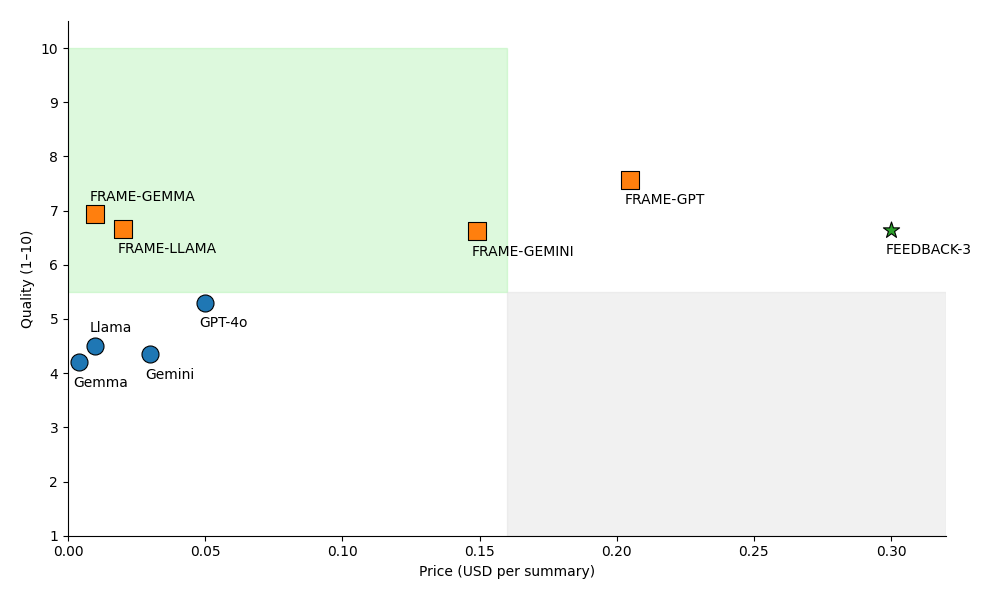}
\caption{4-quadrant plot of total architecture cost (avg.) vs quality measured by MESA. The top left indicates ideal high quality and low cost. Blue dots are single LLM instances for GPT-4o, Gemini 1.5 pro, Llama 3.1 8b, and Gemma 3 4b. Organce squares are \framework{} summaries with the different backbones. FEEDBACK-3 relates to the self-refinement baseline by \citet{KirsteinLG25a} with a GPT-4o backbone and three refinement loops.}
\label{fig:pareto_curve}
\end{figure*}

\Cref{fig:pareto_curve} illustrates the relationship between computational cost and summary quality across different approaches.
The quality metric represents a composite score derived from the eight MESA dimensions (\Cref{sec:experiments}), with higher values (1-10 scale) indicating better quality. 
The horizontal axis represents the average cost per summary in US dollars.

The analysis reveals that single-model baselines (blue circles) appear in the bottom-left quadrant, offering low cost ($\$0.01$-$\$0.05$ per summary) but limited quality (scores 4.0-5.3).
\framework{} implementations (orange squares) consistently achieve higher quality scores (6.5-7.5) across various backbone models, with costs ranging from $\$0.02$ to $\$0.18$ per summary.

\framework{}-LLAMA (\framework{} with Llama 3.1) offers an excellent balance of quality and cost-efficiency, achieving a quality score of approximately 6.7 at just $\$0.03$ per summary.
This represents a substantial quality improvement over the single LLM baselines while maintaining competitive costs.
\framework{}-GPT delivers the highest quality among \framework{} implementations (score $\sim$7.5) but at higher cost ($\$0.18$), placing it top-right.

The feedback-based approach with three iterations (FEEDBACK-3, green star) achieves competitive quality (score $\sim$6.7) but at the highest cost ($\sim\$0.30$), demonstrating that \framework{}'s structured approach provides better efficiency than iterative refinement. 
The light green region highlights the ideal zone of high quality and low cost, where \framework{} implementations with open-source models (\framework{}-GEMMA, \framework{}-LLAMA) deliver particularly favorable trade-offs for practical deployment scenarios.

\section{Empirical Observations in the Pipeline}
\label{sec:empirical_observations}

To understand how \framework{} processes and filters information, we analyze the \textit{Fact Selection} across 100 summaries generated using a GPT backbone model (setup from \Cref{sec:experiments}).
We examine how many facts are typically extracted from a meeting, how many are deemed relevant, and how many are utilized for outlining and summary writing.

\Cref{tab:retention_rates} presents the number of facts throughout the processing stages.
From an average of 103.3 initially extracted facts per meeting, approximately 40\% (mean=41.0, SD=3.1) receive relevance scores $r_i \geq 6$ and are retained for potential inclusion in the summary outline.
The structured outline formation follows a hierarchical relevance approach.
High-priority facts (those receiving scores $r_i \geq 8$) used to define outline points average 8.67 facts per summary, with the majority (1.97) belonging to the \texttt{Decision} category.
Supporting context facts (those with $6 \leq r_i < 8$) average 11.98 per summary, providing background and elaboration for the core points while maintaining a manageable information density.
The distribution of facts across categories relates to domain-specific patterns.
Parliamentary meetings (from the QMSum dataset) show a pronounced emphasis on \texttt{Decision} facts, with summaries containing up to 13 decision-related points and minimal \texttt{Discussion} or \texttt{Next Steps} content.
This is expected, as parliament meetings are centered on decisions regarding petitions.

Notably, we observe no instances where outline points require removal due to hallucination or factual errors when using the GPT backbone.
This suggests that when operating with advanced LLMs, the \framework{} pipeline maintains high fidelity between extracted facts and generated outline points without introducing non-factual content.
Based on our ablation studies with smaller models (\Cref{sec:ablations}), we hypothesize that less capable models would likely exhibit higher rates of outline point removal, potentially necessitating more robust quality control measures.
For deployment scenarios using smaller models, we recommend enhancing the \texttt{Summary Verification} step to actively identify and replace potentially hallucinated outline points rather than simply removing them.

In sum, \framework{}'s multi-stage approach effectively condenses extensive meeting transcripts (averaging 7,303 words) into focused summaries built around approximately 20 high-relevance facts, achieving an overall compression ratio of $\sim$80\% from extraction to final outline formation.

\label{app:retention_rates}

\begin{table}[ht]
\scriptsize
\centering
\begin{tabular}{lll}
\toprule
\rowcolor{gray!20}
\textbf{Metric} & \textbf{Mean} & \textbf{Std. Dev.} \\
\midrule
Facts Extracted (total) & 103.3 & 4.6 \\
Facts Retained ($r_i \geq 6$) & 41.0 & 3.1 \\
Retention Rate (\%) & 39.8 & 2.3 \\

\midrule
Outline Facts ($r_i \geq 8$, main) & 8.67 & 5.18 \\
\hspace*{2mm} Decision & 1.97 & 3.15 \\
\hspace*{2mm} Discussion & 0.78 & 1.54 \\
\hspace*{2mm} Next Steps & 0.74 & 1.47 \\

\midrule
Outline Facts ($r_i \geq 6$, context) & 11.98 & 3.35 \\
\hspace*{2mm} Decision & 4.10 & 3.15 \\
\hspace*{2mm} Discussion & 1.85 & 1.51 \\
\hspace*{2mm} Next Steps & 2.45 & 1.53 \\

\midrule
Facts Removed (main or context) & 0.0 & 0.0 \\
\hspace*{2mm} Decision & 0.0 & 0.0 \\
\hspace*{2mm} Discussion & 0.0 & 0.0 \\
\hspace*{2mm} Next Steps & 0.0 & 0.0 \\

\bottomrule
\end{tabular}
\caption{Fact retention statistics across the \framework{} pipeline showing information filtering patterns and distribution of selected facts by category and relevance tier.}
\label{tab:retention_rates}
\end{table}


\section{Ablation \& Sensitivity Studies}
\label{sec:ablation_studies}

We evaluate the contribution of \framework{}'s verification mechanisms (i.e., \texttt{Fact Verification}, \texttt{Quality Assurance}) and the sensitivity of the pipeline to parameter settings during \texttt{Outline Planning}.
Through controlled ablation experiments, we quantify the impact of each verification component and analyze how varying relevance thresholds affect the summary quality.
In sum, while these mechanisms show minimal impact, they provide safeguards against potential failures.

\subsection{Fact Verification Impact}
\label{app:fact_verification_impact}

To evaluate the effectiveness of the fact verification component, we analyze 150 statement-context tuples randomly sampled from our experimental summaries (see \Cref{sec:experiments}).
Three human annotators from our annotator pool (described in \Cref{app:human_evaluation}, we follow a similar onboarding process) assess whether each extracted fact adheres to our Completeness and Minimalism criteria (\Cref{subsec:atomic_facts}) when compared against the original meeting transcript.
\Cref{fig:extracted_facts_example} presents representative examples of extracted facts from academic meetings, illustrating how our approach captures specific statements with appropriate contextual information.
The examples demonstrate both technical facts (e.g., ``The ANN performs nonlinear discriminant analysis'') and comparative observations (e.g., ``Without the neural network, the performance is better'').

Our analysis reveals that the fact verification module identifies and regenerates facts in  $\sim$5\% of cases (8 out of 150 instances).
Among these, human annotators identify one false positive, resulting in a 12.5\% false positive rate.
Two facts from the total set are found to contain excess contextual information that extends beyond what is explicitly stated in the transcript.
The final example in \Cref{fig:extracted_facts_example} illustrates this tendency, where the context includes interpretive elements (``suggesting an alternative method for feature processing'').
The verification stage successfully detects both of these cases. 
Human annotators identify no violations of the Minimalism criterion.

To assess the impact of fact verification on overall summary quality, we compare MESA scores for summaries generated with and without this verification component.
\Cref{tab:verification_impact} presents this comparison using median scores and standard deviations across all evaluation dimensions.
The results demonstrate minimal differences between the two configurations, with slightly higher variance in hallucination scores when verification is disabled.
This suggests that while fact verification provides limited benefit for \framework{} with a GPT backbone used in our main experiments, it may serve as an important guardrail for deployment scenarios with less capable models or more challenging meeting content.

\begin{figure*}[t]
    \begin{AIbox}{Target Summary Generation}
    \parbox[t]{\textwidth}{

     \{\\
        "fact": "Highly mismatched means clean training with close microphone training and distant microphone testing.",\\
        "context": "PhD C explains that highly mismatched conditions involve clean training with close microphone training and distant microphone testing, which are the most noisy cases." \\
      \},\\
      \{\\
        "fact": "Without the neural network, the performance is better.",\\
        "context": "PhD C mentions that performance is better without the neural network, indicating that adding neural networks might be causing issues."\\
      \},\\
      \{\\
        "fact": "The TIMIT noises include 'Car', 'Babble', 'Subway', and 'Train station'.",\\
        "context": "PhD C lists the types of noises in the TIMIT dataset, which include 'Car', 'Babble', 'Subway', and 'Train station'."\\
      \},\\
      \{\\
        "fact": "The neural net was not helping until the second path with pure features was added.",\\
        "context": "Professor B and PhD C discuss that the neural net was not helping until the second path with pure features was added, indicating the importance of combining neural net features with pure features."\\
      \},\\
      \{\\
        "fact": "The ANN performs nonlinear discriminant analysis.",\\
        "context": "PhD C mentions that the ANN performs nonlinear discriminant analysis, which is similar to LDA but not linear."\\
      \},\\
      \{\\
        "fact": "Non-tandem features were transformed using PCA in the proposal.",\\
        "context": "PhD C states that non-tandem features were transformed using PCA in the proposal, suggesting an alternative method for feature processing."\\
      \},\\

    }
    \end{AIbox}
    \caption{Snippet of facts extracted from a QMSum meeting.}
    \label{fig:extracted_facts_example}
\end{figure*}

\begin{table}[ht]
\scriptsize
\centering
\begin{tabular}{lcc}
\toprule
\rowcolor{gray!20}
\textbf{Approach} & \makecell{\textbf{\framework{}} \\ w/ \texttt{Fact Verification}} & \makecell{\textbf{\framework{}} \\ w/o \texttt{Fact Verification}} \\
\midrule
Coreference     & $0_{\textit{1.64}}$ & $0_{\textit{1.65}}$ \\
Hallucination   & $1_{\textit{1.75}}$ & $1_{\textit{1.81}}$ \\
Incoherence     & $3_{\textit{1.88}}$ & $3_{\textit{1.88}}$ \\
Irrelevance     & $1_{\textit{1.45}}$ & $1_{\textit{1.44}}$ \\
Language        & $1_{\textit{1.40}}$ & $1_{\textit{1.40}}$ \\
Omission        & $1_{\textit{0.16}}$ & $1_{\textit{0.19}}$ \\
Repetition      & $1_{\textit{1.23}}$ & $1_{\textit{1.23}}$ \\
Structure       & $3_{\textit{1.24}}$ & $3_{\textit{1.23}}$ \\
\bottomrule
\end{tabular}
\caption{Changes in MESA scores for running \framework{} on a GPT backbone with and without \texttt{Fact Verification} step. Values are Median$_{Std}$. MESA scores are 1--5 Likert ratings.}
\label{tab:verification_impact}
\end{table}

\subsection{Summary Refinement Analysis}
\label{app:summary_refinement_analysis}

We examine the contribution of the summary verification step by comparing 10 \framework{} summaries generated with and without this component, evaluating performance using both reference-based metrics (ROUGE, BERTScore) and reference-free MESA dimensions.
\Cref{tab:summary_verification_influence} presents this comparison, with highlighting indicating superior performance for each metric.

With \texttt{Summary Verification} enabled, we observe improvements in incoherence (3.5→3), ROUGE-1 (+1.48), ROUGE-L (+1.41), and BERTScore (+1.24).
Analysis of individual summaries reveals that the verification component primarily enforces the target length constraint (250 tokens), editing overly verbose summaries while preserving their core content.
Content modifications were minimal in our test set, suggesting that the primary function of this component is enforcing structural and length constraints rather than correcting factual content when using a GPT-4o backbone.
Based on these findings, we conclude that \texttt{Summary Verification} serves as a quality assurance mechanism that is particularly valuable for maintaining consistent output format and length constraints across diverse meeting types.
For deployment scenarios with strict length requirements or when using less capable models, this component provides an important safeguard.

\begin{table}[ht]
\scriptsize
\centering
\begin{tabular}{lcc}
\toprule

\textbf{Approach} & \makecell{\textbf{\framework{}} \\ w/ \texttt{S. Verification}}  & \makecell{\textbf{\framework{}} \\ w/o \texttt{S. Verification}} \\
\midrule
\rowcolor{gray!20} 
\multicolumn{3}{c}{\textbf{MESA} (lower is better)} \\
\midrule
Coreference     & $0_{\textit{1.64}}$ & $0_{\textit{1.59}}$\\
Hallucination   & $1_{\textit{1.75}}$ & $1_{\textit{1.68}}$ \\
Incoherence     & \cellcolor{highlightGreen}$3_{\textit{1.88}}$ & $3.5_{\textit{1.76}}$ \\
Irrelevance     & $1_{\textit{1.45}}$ & $1_{\textit{1.35}}$ \\
Language        & $1_{\textit{1.40}}$ & $1_{\textit{1.37}}$ \\
Omission        & $1_{\textit{0.16}}$ & $1_{\textit{0.41}}$ \\
Repetition      & $1_{\textit{1.23}}$ & $1_{\textit{0.30}}$ \\
Structure       & $3_{\textit{1.24}}$ & $3_{\textit{1.64}}$ \\
\midrule
\rowcolor{gray!20} 
\multicolumn{3}{c}{\textbf{General Evaluation Metrics} (higher is better)} \\
\midrule
R-1             & \cellcolor{highlightGreen}$22.89_{\textit{5.80}}$ & $21.41_{\textit{4.72}}$\\
R-2             & $4.13_{\textit{2.44}}$ & \cellcolor{highlightGreen}$4.79_{\textit{2.10}}$\\
R-L             & \cellcolor{highlightGreen}$20.78_{\textit{5.23}}$ & $19.37_{\textit{4.39}}$\\
BS (F1)         & \cellcolor{highlightGreen}$85.67_{\textit{1.19}}$ & $84.43_{\textit{1.09}}$\\
\bottomrule
\end{tabular}
\caption{Changes in MESA scores for running \framework{} on a GPT-4o backbone with (\textit{w/}) and without (\textit{w/o}) \texttt{Summary Verification} step. Values are Median$_{Std}$. MESA scores are 1--5 Likert ratings, 
    ROUGE (R-1/R-2/R-L) and BERTScore (BS) are 0--100. \colorbox{highlightGreen}{Green} is best in category.}
\label{tab:summary_verification_influence}
\end{table}

\subsection{Threshold Variation Analysis}
\label{app:threshold_variation}

To understand how fact retention thresholds affect summary quality, we conducted a sensitivity analysis comparing three threshold configurations:
\begin{itemize}
    \item \textbf{Default}: High-relevance ($r_i \geq 8$) for outline points, supporting facts ($r_i \geq 6$) for context
    \item \textbf{Low}: Relaxed thresholds with $r_i \geq 6$ for outline points, $r_i \geq 3$ for context facts
    \item \textbf{High}: Strict thresholds with $r_i \geq 10$ for outline points, $r_i \geq 8$ for context facts
\end{itemize}

\Cref{tab:threshold_analysis} presents comprehensive evaluation results across these configurations.

The high-threshold configuration demonstrates worse scores across most dimensions, particularly in hallucination (+2 points), language quality (+2 points), and omission (+3 points).
We hypothesize that this degradation occurs because the stringent thresholds eliminate moderately important facts that create essential context and connectivity between main points.
The resulting summaries become fragmented, focusing on the highest-scored facts without adequate supporting information.

The low-threshold configuration shows mixed results, with improvements in language quality (-0.5 points) compared to the default.
However, it underperforms on coreference (+3 points), omission (+3 points), and repetition (+3 points).
This pattern suggests that incorporating too many lower-relevance facts introduces redundancy and dilutes the summary's focus, creating challenges in maintaining coherent entity references.

The default configuration achieves the best overall performance, showing optimal balance across nearly all evaluation dimensions.
This suggests that our empirically determined thresholds ($r_i \geq 8$ for outline points, $r_i \geq 6$ for context) effectively balance informativeness and coherence, sufficiently capturing high-relevance content while excluding noise that could degrade summary quality.
These findings validate our parameter selection for the main experiments in \Cref{sec:experiments} and demonstrate the importance of appropriate fact filtration thresholds in the \framework{} pipeline.
They also highlight the framework's sensitivity to these parameters, suggesting that threshold tuning could further optimize performance for specific meeting types or domains.

\begin{table}[ht]
\scriptsize
\centering
\begin{tabular}{lccc}
\toprule

\textbf{Approach} & \textbf{default} & \textbf{low} & \textbf{high} \\
\midrule
\rowcolor{gray!20} 
\multicolumn{4}{c}{\textbf{MESA} (lower is better)} \\
\midrule
Coreference     & \cellcolor{highlightGreen}$0_{\textit{1.64}}$ & $3_{\textit{0.41}}$ & $3_{\textit{0.00}}$ \\
Hallucination   & \cellcolor{highlightGreen}$1_{\textit{1.75}}$ & $1.5_{\textit{1.94}}$ & $3_{\textit{1.82}}$ \\
Incoherence     & \cellcolor{highlightGreen}$3_{\textit{1.88}}$ & $4_{\textit{1.60}}$ & $4_{\textit{0.55}}$ \\
Irrelevance     & \cellcolor{highlightGreen}$1_{\textit{1.45}}$ & $1.5_{\textit{1.64}}$ & $2_{\textit{1.48}}$ \\
Language        & $1_{\textit{1.40}}$ & \cellcolor{highlightGreen}$0.5_{\textit{1.47}}$ & $3_{\textit{1.52}}$ \\
Omission        & \cellcolor{highlightGreen}$1_{\textit{0.16}}$ & $4_{\textit{0.00}}$ & $4_{\textit{0.00}}$ \\
Repetition      & \cellcolor{highlightGreen}$1_{\textit{1.23}}$ & $4_{\textit{0.41}}$ & $2_{\textit{1.52}}$ \\
Structure       & \cellcolor{highlightGreen}$3_{\textit{1.24}}$ & \cellcolor{highlightGreen}$3_{\textit{1.03}}$ & $4_{\textit{0.89}}$ \\
\midrule
\rowcolor{gray!20} 
\multicolumn{4}{c}{\textbf{General Evaluation Metrics} (higher is better)} \\
\midrule
R-1             & \cellcolor{highlightGreen}$22.89_{\textit{5.80}}$ & $21.71_{\textit{2.50}}$ & $21.15_{\textit{3.79}}$\\
R-2             & $4.13_{\textit{2.44}}$ & \cellcolor{highlightGreen}$4.19_{\textit{1.37}}$ & $3.65_{\textit{1.16}}$\\
R-L             & \cellcolor{highlightGreen}$20.78_{\textit{5.23}}$ & $19.25_{\textit{2.71}}$ & $19.10_{\textit{3.53}}$\\
BS (F1)         & \cellcolor{highlightGreen}$85.67_{\textit{1.19}}$ & $85.52_{\textit{0.48}}$ & $85.46_{\textit{0.64}}$\\
\bottomrule
\end{tabular}
\caption{Impact of different retention thresholds on summary quality. Values are Median$_{Std}$. MESA scores are 1--5 Likert ratings, 
    ROUGE (R-1/R-2/R-L) and BERTScore (BS) are 0--100. \colorbox{highlightGreen}{Green} is best in category.}
\label{tab:threshold_analysis}
\end{table}


\section{Extended Analysis}
\label{sec:extended_analysis}
This section extends our evaluation of \framework{} through three complementary analyses: (1) a comparison of fact representation approaches, (2) a cross-model performance evaluation, and (3) a cross-domain application assessment. 
In sum, these experiments evaluate the generalizability of our approach beyond meetings.

\subsection{Fact Representation Comparison}
\label{sec:app_fact_representation_comparison}
Our fact-based summarization approach depends critically on how facts are represented.
To isolate the contribution of our statement-context tuple representation (\Cref{subsec:atomic_facts}), we conduct a controlled comparison against an alternative approach using molecular facts \cite{GunjalD24}, defined as atomic statements with local contextual information derived from the immediate utterance, without the explicit additional global context that our approach incorporates.
We implemented both fact representation strategies within the \framework{} pipeline using identical GPT-4o backbone models and evaluation protocols, testing on the same set of meetings as used for the main experiments \Cref{sec:experiments}. 
\Cref{tab:fact_representation} presents the comparative results.

Our statement-context tuple approach outperforms molecular facts across nearly all evaluation dimensions (MESA, ROUGE, BERTScore).
The most pronounced improvements appear in hallucination (3.5-point improvement), omission (3-point improvement), repetition (3-point improvement), and irrelevance (2-point improvement).
These findings demonstrate that global context is essential for fact interpretation, enabling the model to resolve references, understand broader implications, and connect information across speaker turns. 

The similar scores in structure suggest that both approaches can establish similar organizational frameworks.
However, the differences in content quality metrics highlight that global context enables more accurate and coherent content selection and generation.
This validates our design choice of incorporating broader meeting context into fact representation, suggesting that meeting summarization benefits from rich contextual grounding.

\begin{table}[ht]
\scriptsize
\centering
\begin{tabular}{lcc}
\toprule

\textbf{Approach} & \textbf{our \fact{}} & \textbf{molecular fact} \\
\midrule
\rowcolor{gray!20} 
\multicolumn{3}{c}{\textbf{MESA} (lower is better)} \\
\midrule
Coreference     & \cellcolor{highlightGreen}$0_{\textit{1.64}}$ & $3_{\textit{1.64}}$\\
Hallucination   & \cellcolor{highlightGreen}$1_{\textit{1.75}}$ & $4.5_{\textit{0.45}}$ \\
Incoherence     & \cellcolor{highlightGreen}$3_{\textit{1.88}}$ & $3.5_{\textit{1.22}}$ \\
Irrelevance     & \cellcolor{highlightGreen}$1_{\textit{1.45}}$ & $3_{\textit{0.79}}$ \\
Language        & \cellcolor{highlightGreen}$1_{\textit{1.40}}$ & $1_{\textit{1.14}}$ \\
Omission        & \cellcolor{highlightGreen}$1_{\textit{0.16}}$ & $4_{\textit{0.21}}$ \\
Repetition      & \cellcolor{highlightGreen}$1_{\textit{1.23}}$ & $4_{\textit{0.45}}$ \\
Structure       & $3_{\textit{1.24}}$ & $3_{\textit{1.52}}$ \\
\midrule
\rowcolor{gray!20} 
\multicolumn{3}{c}{\textbf{General Evaluation Metrics} (higher is better)} \\
\midrule
R-1             & \cellcolor{highlightGreen}$22.89_{\textit{5.80}}$ & $20.73_{\textit{5.17}}$\\
R-2             & $4.13_{\textit{2.44}}$ & \cellcolor{highlightGreen}$4.31_{\textit{2.11}}$\\
R-L             & \cellcolor{highlightGreen}$20.78_{\textit{5.23}}$ & $18.48_{\textit{5.08}}$\\
\midrule
BS (F1)         & \cellcolor{highlightGreen}$85.67_{\textit{1.19}}$ & $84.79_{\textit{0.08}}$\\
\bottomrule
\end{tabular}
\caption{Comparison of \framework{} using our fact definition (see \Cref{sec:methodology}) and molecular facts, i.e., facts with local context. Values are Median$_{Std}$. MESA scores are 1--5 Likert ratings, 
    ROUGE (R-1/R-2/R-L) and BERTScore (BS) are 0--100. \colorbox{highlightGreen}{Green} is best in category.}
\label{tab:fact_representation}
\end{table}

\subsection{Cross-Model Performance Comparison}
\label{sec:app_cross_model_performance}

To determine whether \framework{}'s effectiveness depends on specific model capabilities or generalizes across different backbone models, we evaluate the pipeline with four different language models as backbones: GPT-4o and Gemini 1.5 pro (commercial-scale models), and Llama 3.1 8b and Gemma 3 4b (smaller open-source models).
For each model, we apply identical experimental protocols, comparing \framework{} against single-pass summarization with the same backbone model.

\Cref{tab:cross_model} presents comprehensive results across all model variants, highlighting differences in both absolute performance and relative improvement from the base model to the \framework{} pipeline.

Across all model variants, \framework{} consistently improves reference-free MESA metrics while showing mixed results on reference-based metrics.
The most substantial improvements appear in hallucination (a 2-2.5 points improvement across all models), omission (2 points), and repetition (up to 3 points improvement).
These improvements persist regardless of model scale, suggesting that \framework{}'s structured fact-based approach provides benefits independent of model size or architecture.

The performance gap between commercial and open-source models narrows when using \framework{}, particularly for hallucination (commercial: 1-1.5 vs. open-source: 2.5) and irrelevance (both model types: 0.5-1).
We conclude that \framework{}'s explicit fact extraction and verification stages effectively compensate for some limitations of smaller models.
Notably, Llama 3.1 8b with \framework{} achieves comparable performance to commercial models on dimensions like irrelevance and structure, making it a viable option for production deployment.

The consistent pattern of improved factuality and reduced omission across all model variants confirms that \framework{}'s benefits stem from its architectural approach rather than from specific model capabilities.
This cross-model generalizability demonstrates the robustness of our fact-based approach and its potential applicability across diverse deployment scenarios with varying computational constraints.

\subsection{Cross-Domain Evaluation}
\label{sec:app_cross_domain_performance}

To assess whether \framework{}'s benefits extend beyond meeting summarization, we apply our pipeline to three standard document summarization datasets, i.e., arXiv \citep{CohanDKB18a}, XSum \citep{NarayanCL18}, BigPatent \citep{SharmaLW19}, and a specialized domain summarization dataset, PubMed \cite{xiong2024benchmarking}.
These datasets represent varied domains with different linguistic structures: scientific papers (arXiv), news articles (XSum), patent applications (BigPatent), and medical papers (PubMed).
This evaluation assesses whether \framework{}'s fact-centric approach applies to structurally distinct source documents.
For each dataset, we compared \framework{} with GPT-4o backbone against standard single-pass summarization using MESA, ROUGE, and BERTScore for evaluation.

\Cref{tab:cross_domain} presents the results. 
\framework{} consistently improves scores in specific quality dimensions, most notably repetition, where it achieves 2.5 to 3.5-point improvements across all four domains.
A similar strong trend is observed for omission, with 2.5 to 3-point improvements in three of the four datasets. 
This suggests that \framework{}'s structured approach to information extraction and organization helps prevent redundancy and, in most cases, content loss, regardless of the source document type.

Domain-specific patterns also emerge. 
For scientific papers on arXiv, \framework{} shows substantial improvements in irrelevance (-1.5) and language quality (-2.5).
On the PubMed dataset, the most significant gains are seen in hallucination (-2), language (-2), and repetition (-2.5), reflecting its ability to handle complex, structured content. 
For news articles (XSum), \framework{} demonstrates its most substantial improvements in incoherence (-2), omission (-3), and repetition (-3), suggesting benefits for sources with high factual density. 
For patents (BigPatent), improvements are most notable in omission (-3) and repetition (-3) but are more modest in other dimensions, as baseline performance was already strong.

Consistent with our main findings in \Cref{sec:experiments}, reference-based metrics show mixed results.
The structural reorganization performed by \framework{} leads to deviations from the reference summaries, resulting in slightly lower ROUGE scores on arXiv, BigPatent, and PubMed.
This reinforces our conclusion that while ROUGE is a valuable metric, it may not fully capture quality improvements related to better organization and reduced hallucination.

These cross-domain results confirm that \framework{}'s benefits extend beyond conversational content to diverse document types. 
The consistent improvements in repetition and frequent, significant gains in reducing omission suggest that fact-based summarization offers robust advantages for information preservation and non-redundancy across domains, while other benefits may be domain-specific.
These findings motivate \framework{}'s applicability as a general-purpose summarization approach beyond its original meeting summarization context.

\begin{table*}[ht]
\centering
\renewcommand{\arraystretch}{1.1}
\scriptsize
\setlength{\tabcolsep}{2.2pt}
\begin{tabular}{@{}lcc c cc c cc c cc}
\toprule

& \multicolumn{2}{c}{\textbf{arXiv}} 
&& \multicolumn{2}{c}{\textbf{XSum}} 
&& \multicolumn{2}{c}{\textbf{BigPatent}}
&& \multicolumn{2}{c}{\textbf{PUBMED}}\\

\cmidrule(lr){2-3} \cmidrule(lr){5-6} \cmidrule(lr){8-9} \cmidrule(lr){11-12}

& \makecell{GPT\\4o} & \makecell{\framework{}\\GPT-4o} &
& \makecell{GPT\\4o} & \makecell{\framework{}\\GPT-4o} &
& \makecell{GPT\\4o} & \makecell{\framework{}\\GPT-4o} &
& \makecell{GPT\\4o} & \makecell{\framework{}\\GPT-4o} \\

\midrule
\rowcolor{gray!20} 
\multicolumn{12}{c}{\textbf{MESA} (lower is better)} \\
\midrule

Coreference     
    & $0_{\textit{0.32}}$ 
    & $0_{\textit{1.00}}$     
    && $0_{\textit{1.45}}$  
    & $0_{\textit{1.49}}$ 
    && $0_{\textit{0.32}}$  
    & $0_{\textit{0.67}}$  
    && $1_{\textit{1.30}}$  
    & \cellcolor{highlightGreen}$0_{\textit{1.10}}$  \\
Hallucination   
    & $0_{\textit{1.49}}$ 
    & $0_{\textit{1.76}}$     
    && $0.5_{\textit{1.55}}$
    & \cellcolor{highlightGreen}$0_{\textit{0.84}}$ 
    && $0_{\textit{0.95}}$ 
    & $0_{\textit{0.95}}$ 
    && $3_{\textit{1.30}}$  
    & \cellcolor{highlightGreen}$1_{\textit{1.00}}$  \\
Incoherence     
    & \cellcolor{highlightGreen}$0_{\textit{1.96}}$ 
    & $0.5_{\textit{1.69}}$     
    && $3_{\textit{1.89}}$  
    & \cellcolor{highlightGreen}$1_{\textit{1.48}}$ 
    && $0_{\textit{1.79}}$ 
    & $0_{\textit{1.52}}$  
    && $3_{\textit{0.55}}$  
    & \cellcolor{highlightGreen}$2_{\textit{1.58}}$  \\
Irrelevance     
    & $2_{\textit{1.26}}$ 
    & \cellcolor{highlightGreen}$0.5_{\textit{1.51}}$     
    && $2_{\textit{1.05}}$  
    & \cellcolor{highlightGreen}$0.5_{\textit{1.32}}$ 
    && $0_{\textit{1.32}}$ 
    & $0_{\textit{1.25}}$  
    && $2_{\textit{0.89}}$  
    & \cellcolor{highlightGreen}$1_{\textit{1.34}}$  \\
Language        
    & $2.5_{\textit{1.37}}$ 
    & \cellcolor{highlightGreen}$0_{\textit{1.48}}$   
    && $0_{\textit{0.71}}$  
    & \cellcolor{highlightGreen}$0.5_{\textit{1.42}}$ 
    && $0.5_{\textit{0.92}}$ 
    & \cellcolor{highlightGreen}$0_{\textit{1.34}}$  
    && $2_{\textit{1.30}}$  
    & \cellcolor{highlightGreen}$0_{\textit{1.30}}$  \\
Omission        
    & $4_{\textit{0.42}}$ 
    & \cellcolor{highlightGreen}$1.5_{\textit{0.33}}$     
    && $4_{\textit{0.42}}$  
    & \cellcolor{highlightGreen}$1_{\textit{0.48}}$ 
    && $4_{\textit{0.11}}$ 
    & \cellcolor{highlightGreen}$1_{\textit{0.00}}$  
    && $4_{\textit{0.45}}$  
    & $4_{\textit{0.00}}$  \\
Repetition      
    & $4_{\textit{0.84}}$ 
    & \cellcolor{highlightGreen}$0.5_{\textit{1.01}}$     
    && $4_{\textit{1.35}}$  
    & \cellcolor{highlightGreen}$1_{\textit{1.49}}$ 
    && $4_{\textit{0.85}}$ 
    & \cellcolor{highlightGreen}$1_{\textit{0.82}}$  
    && $3_{\textit{1.22}}$  
    & \cellcolor{highlightGreen}$0.5_{\textit{0.30}}$  \\
Structure       
    & \cellcolor{highlightGreen}$0_{\textit{0.95}}$ 
    & $0.5_{\textit{1.51}}$     
    && \cellcolor{highlightGreen}$0_{\textit{0.97}}$  
    & $0.5_{\textit{1.73}}$ 
    && $0_{\textit{0.43}}$ 
    & $0_{\textit{1.26}}$  
    && $2_{\textit{1.22}}$  
    & \cellcolor{highlightGreen}$1_{\textit{0.89}}$  \\

\midrule
\rowcolor{gray!20} 
\multicolumn{12}{c}{\textbf{ROUGE (R-1, R-2, R-L) and BERTScore (BS)} (higher is better)} \\
\midrule

R-1             
    & \cellcolor{highlightGreen}$28.92_{\textit{9.59}}$ 
    & $23.97_{\textit{6.45}}$ 
    && $14.80_{\textit{4.28}}$ 
    & \cellcolor{highlightGreen}$15.08_{\textit{3.51}}$ 
    && \cellcolor{highlightGreen}$31.32_{\textit{5.21}}$ 
    & $23.22_{\textit{3.45}}$ 
    && \cellcolor{highlightGreen}$27.66_{\textit{6.25}}$  
    & $25.58_{\textit{5.65}}$  \\
R-2            
    & \cellcolor{highlightGreen}$8.71_{\textit{5.06}}$ 
    & $6.19_{\textit{2.47}}$ 
    && $2.32_{\textit{1.86}}$ 
    & \cellcolor{highlightGreen}$2.95_{\textit{2.44}}$ 
    && \cellcolor{highlightGreen}$10.19_{\textit{6.14}}$ 
    & $4.35_{\textit{2.86}}$
    && $6.10_{\textit{3.12}}$  
    & \cellcolor{highlightGreen}$6.34_{\textit{2.63}}$  \\
R-L             
    & \cellcolor{highlightGreen}$25.77_{\textit{8.50}}$ 
    & $20.71_{\textit{5.62}}$ 
    && $13.24_{\textit{3.84}}$ 
    & \cellcolor{highlightGreen}$13.64_{\textit{3.99}}$ 
    && \cellcolor{highlightGreen}$29.09_{\textit{5.44}}$ 
    & $20.18_{\textit{3.38}}$ 
    && \cellcolor{highlightGreen}$23.43_{\textit{5.73}}$  
    & $23.33_{\textit{5.17}}$  \\
BS (F1)         
    & \cellcolor{highlightGreen}$83.62_{\textit{2.19}}$ 
    & $83.55_{\textit{1.19}}$ 
    && $84.64_{\textit{1.24}}$ 
    & \cellcolor{highlightGreen}$85.15_{\textit{0.90}}$ 
    && \cellcolor{highlightGreen}$86.13_{\textit{1.32}}$ 
    & $84.66_{\textit{1.50}}$ 
    && \cellcolor{highlightGreen}$85.45_{\textit{1.32}}$  
    & $85.40_{\textit{1.19}}$  \\
\bottomrule
\end{tabular}
\caption{Comparison of information preservation and hallucination rates across fact representation approaches. Values are Median$_{Std}$. MESA scores are 1--5 Likert ratings, 
    ROUGE (R-1/R-2/R-L) and BERTScore (BS) are 0--100. \colorbox{highlightGreen}{Green} is best in category.}
\label{tab:cross_domain}
\end{table*}

\begin{table*}[ht]
\scriptsize
\centering
\renewcommand{\arraystretch}{1.1} 
\scriptsize
\setlength{\tabcolsep}{2.2pt} 
\begin{tabular}{@{}lcc c cc c cc c cc}

\toprule
& \multicolumn{2}{c}{\textbf{GPT-4o}} 
&& \multicolumn{2}{c}{\textbf{Gemini 1.5 pro}} 
&& \multicolumn{2}{c}{\textbf{Llama 3.1 8b}}
&& \multicolumn{2}{c}{\textbf{Geamm 3 4b}}\\

\cmidrule(lr){2-3} \cmidrule(lr){5-6} \cmidrule(lr){8-9} \cmidrule(lr){11-12}

& \makecell{\framework{}} & \makecell{Single LLM} &
& \makecell{\framework{}} & \makecell{Single LLM} &
& \makecell{\framework{}} & \makecell{Single LLM} &
& \makecell{\framework{}} & \makecell{Single LLM} \\

\midrule
\rowcolor{gray!20} 
\multicolumn{12}{c}{\textbf{MESA} (lower is better)} \\
\midrule

Coreference     & \cellcolor{highlightGreen}$0_{\textit{1.64}}$ & $0_{\textit{1.22}}$ && \cellcolor{highlightGreen}$0_{\textit{1.59}}$  & $3_{\textit{1.58}}$ && \cellcolor{highlightGreen}$0_{\textit{1.43}}$ & $1_{\textit{1.21}}$  && \cellcolor{highlightGreen} $2_{\textit{1.43}}$ & $4_{\textit{1.13}}$\\
Hallucination   & \cellcolor{highlightGreen}$1_{\textit{1.75}}$ & $3_{\textit{1.22}}$ && \cellcolor{highlightGreen}$1_{\textit{1.68}}$ & $4_{\textit{2.04}}$ && \cellcolor{highlightGreen}$2.5_{\textit{1.64}}$  & $5_{\textit{1.47}}$ && \cellcolor{highlightGreen}$2.5_{\textit{2.09}}$ & $4_{\textit{1.82}}$\\
Incoherence     & \cellcolor{highlightGreen}$3_{\textit{1.88}}$ & $4_{\textit{1.50}}$ && \cellcolor{highlightGreen}$3.5_{\textit{1.76}}$ & $4_{\textit{1.09}}$ &&$3_{\textit{1.64}}$  & $3_{\textit{1.74}}$ &&$1.5_{\textit{1.60}}$ & \cellcolor{highlightGreen} $1_{\textit{1.40}}$\\
Irrelevance     & \cellcolor{highlightGreen}$1_{\textit{1.45}}$ & $2_{\textit{1.70}}$ && \cellcolor{highlightGreen}$1_{\textit{1.35}}$ & $3_{\textit{1.32}}$ && \cellcolor{highlightGreen}$0.5_{\textit{1.64}}$  & $3_{\textit{1.63}}$ && \cellcolor{highlightGreen}$0.5_{\textit{1.69}}$ & $4_{\textit{1.63}}$\\
Language        & $1_{\textit{1.40}}$ & $1_{\textit{1.30}}$ && \cellcolor{highlightGreen}$1_{\textit{1.37}}$ & $2_{\textit{1.44}}$ &&$2_{\textit{1.64}}$  & $2_{\textit{1.83}}$ &&$1_{\textit{1.59}}$ & $1_{\textit{1.33}}$\\
Omission        & \cellcolor{highlightGreen}$1_{\textit{0.16}}$ & $3_{\textit{0.40}}$ && \cellcolor{highlightGreen}$1_{\textit{0.41}}$ & $3_{\textit{0.38}}$ &&\cellcolor{highlightGreen} $2_{\textit{1.64}}$  & $4_{\textit{1.79}}$ && \cellcolor{highlightGreen}$2_{\textit{1.50}}$ & $4_{\textit{1.45}}$ \\
Repetition      & \cellcolor{highlightGreen}$1_{\textit{1.23}}$ & $4_{\textit{1.05}}$ && \cellcolor{highlightGreen}$1_{\textit{0.30}}$ & $3_{\textit{0.98}}$ && \cellcolor{highlightGreen}$2_{\textit{1.64}}$  & $3_{\textit{1.74}}$&& \cellcolor{highlightGreen}$1_{\textit{1.11}}$ & $2_{\textit{1.46}}$\\
Structure       & \cellcolor{highlightGreen}$3_{\textit{1.24}}$ & $4_{\textit{0.90}}$ && $3_{\textit{1.64}}$ & $3_{\textit{1.70}}$ &&$3_{\textit{1.64}}$  & $3_{\textit{1.43}}$ && \cellcolor{highlightGreen}$3_{\textit{1.29}}$ & $5_{\textit{1.33}}$\\

\midrule
\rowcolor{gray!20} 
\multicolumn{12}{c}{\textbf{ROUGE (R-1, R-2, R-L) and BERTScore (BS)} (higher is better)} \\
\midrule

R-1             & $22.89_{\textit{5.80}}$ & \cellcolor{highlightGreen}$37.73_{\textit{5.85}}$ &&$21.41_{\textit{4.72}}$ & \cellcolor{highlightGreen}$39.61_{\textit{7.21}}$ &&\cellcolor{highlightGreen}$27.27_{\textit{7.32}}$ & $20.56_{\textit{4.85}}$ && $15.23_{\textit{5.93}}$ & \cellcolor{highlightGreen}$20.10_{\textit{5.72}}$\\
R-2             & $4.13_{\textit{2.44}}$ & \cellcolor{highlightGreen}$7.95_{\textit{4.18}}$ && $4.79_{\textit{2.10}}$ & \cellcolor{highlightGreen}$11.10_{\textit{4.86}}$ &&\cellcolor{highlightGreen}$4.37_{\textit{2.13}}$ & $1.36_{\textit{1.02}}$ && \cellcolor{highlightGreen}$1.82_{\textit{1.52}}$ & $1.53_{\textit{0.44}}$ \\
R-L             & $20.78_{\textit{5.23}}$ & \cellcolor{highlightGreen}$21.39_{\textit{4.05}}$ && $19.37_{\textit{4.39}}$ & \cellcolor{highlightGreen}$27.55_{\textit{6.36}}$ &&\cellcolor{highlightGreen}$25.00_{\textit{4.36}}$& $18.69_{\textit{4.32}}$ &&$15.23_{\textit{5.84}}$ & \cellcolor{highlightGreen}$19.14_{\textit{6.02}}$ \\
BS (F1)         & $85.67_{\textit{1.19}}$ & \cellcolor{highlightGreen}$81.61_{\textit{2.87}}$ && \cellcolor{highlightGreen}$84.43_{\textit{1.09}}$ & $80.64_{\textit{3.66}}$ && \cellcolor{highlightGreen}$86.23_{\textit{1.07}}$ & $85.31_{\textit{2.12}}$ &&\cellcolor{highlightGreen}$85.20_{\textit{2.27}}$ & $84.66_{\textit{1.27}}$\\
\bottomrule
\end{tabular}
\caption{\framework{} powered with different backbone models. We report MESA, ROUGE, and BERTScore. Values are Median$_{Std}$. MESA scores are 1--5 Likert ratings, 
    ROUGE (R-1/R-2/R-L) and BERTScore (BS) are 0--100. \colorbox{highlightGreen}{Green} is best in category.}
\label{tab:cross_model}
\end{table*}

\subsection{Architecture Minimality Analysis}
To validate that each stage of \framework{} is essential, we conduct an architectural ablation study.
We evaluate three combinations:
\begin{itemize}
    \item \texttt{combined-1}: A fully collapsed, single-prompt approach.
    \item \texttt{combined-2}: Merges the Fact Extraction and Relevance Scoring stages.
    \item \texttt{combined-3}: Merges the Relevance Scoring and Outline Planning stages.
\end{itemize}

The results in \Cref{tab:stage-combination} demonstrate that every stage is critical for maintaining quality, especially regarding content fidelity.
While the collapsed pipelines produce summaries with reasonable structure and coherence, they fail on factuality.

Specifically, all three variants score a 4.0 on Omission, a 3-point degradation compared to the complete \framework{}, showing they consistently fail to include critical information.
The most degradation is in Hallucination, where \texttt{combined-2} scores a 5.0, where \framework{} is reported with 1.0.
This strongly suggests that forcing a model to extract and evaluate facts in a single step prevents it from establishing a stable set of grounded claims, leading to rampant hallucination.
These findings confirm that our modular, seven-step architecture is essential for producing reliable and comprehensive summaries.

\begin{table}[ht]
\centering
\renewcommand{\arraystretch}{1.1}
\scriptsize
\setlength{\tabcolsep}{2.2pt}
\begin{tabular}{@{}lc c c c}
\toprule

& \multicolumn{1}{c}{\textbf{combined-1}} 
& \multicolumn{1}{c}{\textbf{combined-2}} 
& \multicolumn{1}{c}{\textbf{combined-3}}
& \multicolumn{1}{c}{\textbf{\framework{}}}\\

\cmidrule(lr){2-2} \cmidrule(lr){3-3} \cmidrule(lr){4-4} \cmidrule(lr){5-5}

& \makecell{GPT\\4o}
& \makecell{GPT\\4o}
& \makecell{GPT\\4o}
& \makecell{GPT\\4o} \\

\midrule
\rowcolor{gray!20} 
\multicolumn{5}{c}{\textbf{MESA} (lower is better)} \\
\midrule

Coreference     
    & \cellcolor{highlightGreen}$0_{\textit{1.25}}$  
    & $1.5_{\textit{1.62}}$  
    & \cellcolor{highlightGreen}$0_{\textit{1.70}}$  
    & \cellcolor{highlightGreen}$0_{\textit{1.64}}$  \\
Hallucination   
    & $2_{\textit{2.08}}$  
    & $5_{\textit{0.70}}$  
    & $4.5_{\textit{1.15}}$  
    & \cellcolor{highlightGreen}$1_{\textit{1.75}}$  \\
Incoherence     
    & \cellcolor{highlightGreen}$3_{\textit{1.96}}$  
    & $4_{\textit{2.07}}$  
    & \cellcolor{highlightGreen}$3_{\textit{1.43}}$  
    & \cellcolor{highlightGreen}$3_{\textit{1.88}}$  \\
Irrelevance     
    & $2.5_{\textit{1.40}}$  
    & $3_{\textit{1.66}}$  
    & $2.5_{\textit{1.40}}$  
    & \cellcolor{highlightGreen}$1_{\textit{1.45}}$  \\
Language        
    & $2_{\textit{1.65}}$  
    & $2.5_{\textit{1.32}}$  
    & $1.5_{\textit{1.41}}$  
    & \cellcolor{highlightGreen}$1_{\textit{1.40}}$  \\
Omission        
    & $4_{\textit{0.32}}$  
    & $4_{\textit{0.32}}$  
    & $4_{\textit{0.32}}$  
    & \cellcolor{highlightGreen}$1_{\textit{0.16}}$  \\
Repetition      
    & $3_{\textit{1.52}}$  
    & $2.5_{\textit{1.43}}$  
    & $2.5_{\textit{0.53}}$  
    & \cellcolor{highlightGreen}$1_{\textit{1.23}}$  \\
Structure       
    & \cellcolor{highlightGreen}$3_{\textit{1.65}}$  
    & $3.5_{\textit{1.27}}$  
    & \cellcolor{highlightGreen}$3_{\textit{1.55}}$  
    & \cellcolor{highlightGreen}$3_{\textit{1.24}}$  \\

\midrule
\rowcolor{gray!20} 
\multicolumn{5}{c}{\textbf{ROUGE (R-1, R-2, R-L) and BERTScore (BS)} (higher is better)} \\
\midrule

R-1             
    & $22.33_{\textit{4.39}}$  
    & $18.00_{\textit{3.50}}$  
    & $19.02_{\textit{3.64}}$  
    & \cellcolor{highlightGreen}$22.89_{\textit{5.89}}$  \\
R-2            
    & $3.85_{\textit{2.46}}$  
    & $2.82_{\textit{1.34}}$  
    & $3.36_{\textit{2.30}}$  
    & \cellcolor{highlightGreen}$4.13_{\textit{2.44}}$  \\
R-L             
    & $20.24_{\textit{3.82}}$  
    & $16.46_{\textit{3.17}}$  
    & $17.41_{\textit{2.80}}$  
    & \cellcolor{highlightGreen}$20.78_{\textit{5.23}}$  \\
BS (F1)         
    & $85.36_{\textit{1.14}}$  
    & $84.79_{\textit{0.96}}$  
    & $84.86_{\textit{0.85}}$  
    & \cellcolor{highlightGreen}$85.67_{\textit{1.19}}$  \\
\bottomrule
\end{tabular}
\caption{Ablation study on pipeline architecture. We compare our full pipeline (\framework{}) against three variants with collapsed stages: \textbf{combined-1} (all-in-one), \textbf{combined-2} (extraction+scoring), and \textbf{combined-3} (scoring+planning). Collapsing stages severely degrades factuality (Hallucination, Omission). Values are Median$_{Std}$. \colorbox{highlightGreen}{Green} is best in category.}
\label{tab:stage-combination}
\end{table}


\section{Dataset Characteristics and Coverage}
\label{sec:dataset_characteristics}

Our evaluation employs two complementary datasets representing different aspects of meeting summarization: QMSum \citep{ZhongYYZ21h}, an established benchmark containing diverse meeting types, and FAME \citep{KirsteinKWR25}, a recently developed synthetic dataset with controlled properties.
Together, the datasets provide a comprehensive evaluation environment that spans various meeting domains, structures, and characteristics.

QMSum encompasses 232 meetings across three distinct domains: academic meetings (ICSI), product design discussions (AMI), and government proceedings (WCPC).
This diversity enables evaluation of \framework{}'s effectiveness across varying discourse styles, from structured parliamentary debates to informal product brainstorming sessions.
The meetings contain on average 7.2 speakers and 521 turns per meeting, with variation in meeting length (mean: 7,303 words, SD: 4,232).

FAME complements QMSum with 800 synthetic meetings (500 English, 300 German) generated by LLM agents simulating realistic conversational dynamics.
It spans 14 meeting formats and 28 distinct topics, including academic discussions, corporate planning sessions, and technical workshops.
A key feature of FAME is its controlled inclusion of conversational interruptions (approximately 50\% of meetings contain interruptions), enabling assessment of how effectively summarization approaches handle overlapping speech.

\Cref{tab:corpora_metrics} presents comprehensive statistics across these datasets.
Two characteristics are particularly relevant to summarization:

\begin{itemize}
    \item \textbf{Vocabulary diversity:} QMSum exhibits the most extensive vocabulary (20,505 unique tokens), suggesting greater topical diversity and potentially more challenging summarization.
    
    \item \textbf{Meeting length variation:} Standard deviations in word counts exceed 50\% of means across all datasets, indicating that summarization approaches must handle substantial variation in source length.
    
\end{itemize}

For our evaluation experiments, we randomly select English meetings from each dataset (QMSum and FAME), stratified to maintain the original distribution of meeting types, ensuring comprehensive domain coverage.
This sampling strategy enables assessment across diverse meeting scenarios while maintaining reasonable computational requirements.

\begin{table*}[t]
    \renewcommand{\arraystretch}{1.2} 
    \setlength{\tabcolsep}{5pt} 
    \tiny
    \centering
    \begin{tabular}{lllllllllll}
        \toprule
        \rowcolor{gray!20} 
        Dataset & \textbf{\# Meetings} & \textbf{\# Speaker} & \textbf{\# Unique Spea.} & \textbf{\# Turns} & \textbf{\# Words} & \textbf{Vocab.} & \textbf{Token Overlap} & \textbf{Sum. Len.} & \textbf{Interruptions} & \textbf{Language} \\
        \midrule
        AMI   & 137 & $4.0_{\textit{0.00}}$ & 4    & $513.5_{\textit{266.2}}$ & $4937.5_{\textit{1999.3}}$ & 9388 & - & $109.9_{\textit{27.1}}$ & no &  informal \\
        ICSI  & 44  & $6.2_{\textit{1.3}}$   & 35   & $757.5_{\textit{374.8}}$ & $9889.4_{\textit{3794.9}}$ & 9164 & - & $93.3_{\textit{22.2}}$  & no &  formal \\
        WCPC  & 51  & $16.8_{\textit{18.7}}$ & 316  & $337.3_{\textit{277.3}}$ & $11427.8_{\textit{4574.0}}$ & 13780 & - & $122.3_{\textit{39.2}}$  & no &  informal \\
        QMSum & 232 & $7.2_{\textit{10.1}}$  & 330  & $521.0_{\textit{320.4}}$ & $7303.4_{\textit{4232.2}}$  & 20505 & - & $109.5_{\textit{30.7}}$  & no &  both \\
        \midrule
        EN    & 500 & $5.1_{\textit{2.8}}$  & 3200 & $405.0_{\textit{330.3}}$ & $6223.4_{\textit{4084.4}}$ & 10347 & 0.081 & $207.7_{\textit{22.7}}$ & yes ($\sim$ 0.5)&  both  \\
        GER   & 300 & $5.0_{\textit{2.8}}$  & 1000 & $393.3_{\textit{323.2}}$ & $6272.4_{\textit{3793.2}}$ & 9589  & 0.096 & $170.3_{\textit{29.0}}$ & yes ($\sim$ 0.5) & both  \\
        \bottomrule
    \end{tabular}
    \caption{Statistics on FAME for English and QMSum corpora. Values are Mean$_{Std}$. Table stems from \cite{KirsteinKWR25}.}
    \label{tab:corpora_metrics}
\end{table*}

\section{\reasoning{} Component Analysis}
\label{sec:reasoning_analysis}

This section examines the cognitive foundations and empirical effectiveness of \reasoning{}'s structured reasoning approach to personalization.
We analyze how explicit the \emph{reason-out-loud} questioning enhances personalized summarization compared to alternative approaches, and investigate which reasoning components contribute most significantly to improved personalization.

\subsection{\reasoning{} Reasoning Questionnaire}
\label{appendix:reasoning_questionnaire}

\reasoning{} employs a structured questionnaire derived from cognitive psychology research on metacognitive reasoning and think-aloud protocols \citep{Solomon95,Konrad17}.
These protocols reveal that explicitly verbalizing reasoning processes enhances decision consistency and reduces biases by activating higher-order analytical thinking. 
\Cref{tab:reasoning_questionnaire} presents our nine-question protocol to guide the model through a systematic reasoning process before selecting salient facts for personalization.

The questionnaire follows a cognitive progression through four phases:

\begin{enumerate}
    \item \textbf{Planning} (Q1-Q3): Establishes the target reader's knowledge context, current projects, and primary goals. This activation primes subsequent relevance judgments with specific criteria rather than generic role stereotypes.
    
    \item \textbf{Initial Assessment} (Q4-Q7): Guides explicit reasoning about relevance for each potential fact, considering urgency, comprehension needs, and concrete applications. This phase emphasizes justification over intuition.
    
    \item \textbf{Controlling} (Q8): Provides an explicit filtering mechanism, requiring reconsideration of initial assessments through a critical lens. This meta-review reduces confirmation bias by encouraging active elimination rather than just selection.
    
    \item \textbf{Evaluation} (Q9): Prompts uncertainty awareness, encouraging identification of ambiguous or difficult-to-classify information. This final metacognitive step acknowledges limitations in judgment confidence.
\end{enumerate}

\Cref{fig:questions_answered_example} illustrates how a language model engages with this reasoning process when taking the perspective of a graduate student from the ICSI dataset.
The model explicitly articulates the student's knowledge background, current projects, and priorities before evaluating specific facts.
This explicit articulation creates a more stable and consistent representation of the target reader than implicit role-playing, reducing the likelihood of perspective hallucination or inconsistent fact selection.

\begin{table*}[t]
    \centering
    \renewcommand{\arraystretch}{1.2} 
    \scriptsize
    \setlength{\tabcolsep}{4pt} 
    \begin{tabularx}{\linewidth}{l X p{0.14\linewidth}}
        \toprule
        \rowcolor{gray!20} 
         & \textbf{Definition} & Category \\
        Q1    & What prior knowledge do you have? & Planning \\
        Q2    & Which project are you currently working on? & Planning\\
        Q3    & What are your primary interests and goals? & Planning\\
        Q4    & Read each fact carefully and think about which information is most relevant to you in your role. Explain why. & Initial Assessment\\
        Q5    & Is there an urgency or priority that aligns particularly closely with your current responsibilities or known concerns? & Initial Assessment\\
        Q6    & Which information might require simplification or additional context to ensure clear comprehension? & Initial Assessment\\
        Q7    & You've selected information that you consider important. Review this selection once more and provide concrete examples explaining why these details are relevant for you. & Initial Assessment\\
        Q8    & Now, go through the list a second time and identify which information you consider irrelevant or unimportant for your role/persona, providing reasons for your decisions. & Controlling\\
        Q9    & Are there any topics you found difficult to classify or about which you felt unsure? If so, what are they? & Evaluation\\
       
        \bottomrule
    \end{tabularx}
    \caption{\reasoning{} questionnaire used to familiarize the LLM with the input before selecting salient facts for a target reader.}
    \label{tab:reasoning_questionnaire}
\end{table*}

\begin{figure*}[t]
    \begin{AIbox}{\reasoning{} questionnaire answering}
    \parbox[t]{\textwidth}{

     \textbf{(1) What prior knowledge do you have?}
     
     I have a solid understanding of probabilistic models, feature extraction, and the importance of cues in determining modes of interaction.
     I am familiar with the concepts of binary variables and the categorization of time-related variables. \\
     
     \textbf{(2) Which project are you currently working on?}
     
     I am currently working on a project that involves analyzing and modeling human behavior based on various cues and variables.
     This includes understanding how different factors influence decisions and actions in specific contexts. \\
     
     \textbf{(3) What are your primary interests and goals?}
     
     My primary interests include improving the accuracy of behavior prediction models, enhancing feature extraction techniques, and understanding the impact of different cues on decision-making processes.
     My goals are to refine the models to better predict outcomes and ensure that the extracted features are relevant and useful. \\
     
     \textbf{(4) Read each fact carefully and think about which information is most relevant to you in your role. Explain why.}
     
     - "\textit{The final destination, business, hurry, and tourist variables depend probabilistically on other factors.}" This is relevant because understanding the probabilistic dependencies helps in refining the model.
     
     - "\textit{The mode depends on all the inferred variables.}" This is crucial as it directly impacts the accuracy of the mode prediction.
     
     - "\textit{Features are extracted from discourse and parsed, but no nodes are created for them.}" This is important for feature extraction techniques.
     
     - "\textit{Admission fee is a clue that people are trying to enter a place.}" This helps in identifying specific modes of interaction.
     
     - "\textit{Certain cues are either lexical or topic-based concept cues.}" This is relevant for understanding how different types of cues influence behavior.
     
     - "\textit{Some cues come from utterances, while others are based on world knowledge or situational factors.}" This broadens the scope of cue sources, which is important for comprehensive modeling.
     
     - "\textit{The 'admission fee' cue points directly to the mode 'Enter'.}" This provides a clear link between a specific cue and a mode.
     
     - "\textit{Landmarks are considered touristic landmarks, not navigational landmarks.}" This helps in categorizing landmarks correctly for analysis.
     
     - "\textit{The 'admission fee' is a binary variable.}" This is relevant for setting up the model correctly.
     
     - "\textit{The 'time of day' variable includes morning, afternoon, and night.}" This helps in categorizing time-related variables accurately. \\

    \textbf{(5) Is there an urgency or priority that aligns particularly closely with your current responsibilities or known concerns?}
     
    Yes, understanding the dependencies of variables and the sources of cues is urgent, as it directly impacts the accuracy and reliability of the behavior prediction models. \\

    [...]

    }
    \end{AIbox}
    \caption{Example of a role-playing GPT answering the questionnaire \Cref{tab:reasoning_questionnaire} before fact selection.}
    \label{fig:questions_answered_example}
\end{figure*}

\begin{figure*}[t]
    \begin{AIbox}{Stage 1: Facts Extraction Prompt}
    \parbox[t]{\textwidth}{
        You are an expert at breaking down meeting transcripts into facts. Your task is to extract clear, factual statements with proper context. 
        \newline
        
        \textbf{IMPORTANT RULES}: \newline
        1. Output must be a valid JSON list of objects \newline
        2. NEVER add information not in the transcript \newline
        3. Skip unclear or ambiguous content \newline
        4. Each fact must be atomic (single piece of information) \newline
        5. NO hallucination or inference \newline
        
        \textbf{CONTENT GUIDELINES – STRICTLY FOLLOW}: \newline
        \textbf{1. INCLUDE}: \newline
        \hspace*{1em}– Clear, explicit statements \newline
        \hspace*{1em}– Complete, meaningful information \newline
        \hspace*{1em}– Actionable items or decisions \newline
        \hspace*{1em}– Important discussion points \newline
        \hspace*{1em}– Concrete facts or outcomes \newline
        
        \textbf{2. EXCLUDE}: \newline
        \hspace*{1em}– Filler statements (e.g., ``OK'', ``Right'', ``Mm-hmm'') \newline
        \hspace*{1em}– General acknowledgments \newline
        \hspace*{1em}– Incomplete or unclear statements \newline
        \hspace*{1em}– Transcription artifacts like \{\texttt{disfmarker}\} or \{\texttt{vocalsound}\} \newline
        \hspace*{1em}– Redundant and Ambiguous information \newline
        
        \textbf{3. For each included fact, provide}: \newline
        \hspace*{1em}– ``\texttt{fact}'': Single, atomic piece of information \newline
        \hspace*{1em}– ``\texttt{context}'': Comprehensive context with history \newline
        
        \textbf{Output Format}: Return a JSON list of objects (no additional keys), e.g., \newline
        \{\newline
        \hspace*{1em}\texttt{"fact"}: ``Team agreed to launch product in Q3'',\newline
        \hspace*{1em}\texttt{"context"}: ``Following previous delays and market analysis, Q3 was chosen for optimal impact''\newline
        \}\newline
        
        \textbf{Your Task}: \newline
        Break down this transcript chunk into atomic facts with context. \newline
        Remember: Must return a valid JSON list. Only include clear, explicit information. Skip all filler words, acknowledgments, and artifacts. Break compound statements into facts. Exclude unclear or ambiguous content. \newline
        
        \textbf{Input Variables}: \newline
        Previous context: \textbf{\{previous\_chunk\_context\}} \newline
        Current chunk: \textbf{\{chunk\}} \newline
    }
    \end{AIbox}
    \caption{Prompt template to extract facts from meeting transcripts as JSON output.}
    \label{fig:atomic_facts_prompt}
\end{figure*}

\begin{figure*}[t]
    \begin{AIbox}{Stage 1: Fact Verification Prompt}
    \parbox[t]{\textwidth}{
        You are an expert at detecting hallucinations in extracted information. Your task is to validate facts against the source text. \newline

        \textbf{PROCESS FOR EACH FACT}: \newline
        1. Compare the fact directly with \textbf{SOURCE TEXT} \newline
        2. Verify that \textbf{context} contains only information supported by the source \newline
        3. Flag any unsupported assumptions, inferences, or exaggerations \newline

        \textbf{VALIDATION CHECKLIST}: \newline
        – Is the \textit{fact} explicitly supported by the source? \newline
        – Does \textit{context} stay strictly within source content? \newline
        – Are any details hallucinated or embellished? \newline

        \textbf{IMPORTANT GUIDELINES}: \newline
        – Process each fact individually \newline
        – Be specific about any hallucinated information \newline
        – Flag \textit{all} content not found in the source text \newline
        – Evaluate both \textit{fact} and \textit{context} fields \newline

        \textbf{OUTPUT FORMAT}: Return \textbf{one} JSON object with exactly these keys: \newline
        \{ \newline
        \hspace*{1em}\texttt{"overall\_score"}: score 0–100, where 0 = no hallucination, 100 = completely irrelevant,\newline
        \hspace*{1em}\texttt{"feedback"}: list of specific hallucination or mismatch notes,\newline
        \hspace*{1em}\texttt{"summary"}: brief summary of validation findings\newline
        \} \newline

        \textbf{Your Task}: \newline
        Validate these facts against the SOURCE TEXT. \newline
        – Flag \textit{any} unsupported information \newline
        – Identify hallucinated details, exaggerations, or assumptions \newline
        – Focus on hallucination detection in your overall evaluation \newline
 
        \textbf{Input Variables}: \newline
        Context: \textbf{\{previous\_chunk\_context\}} \newline
        SOURCE TEXT: \textbf{\{chunk\}} \newline
        Facts to validate: \textbf{\{atomic\_facts\}} \newline
    }
    \end{AIbox}
    \caption{Prompt template for validating extracted facts against source text and resolving hallucinations.}
    \label{fig:fact_verification_prompt}
\end{figure*}

\begin{figure*}[t]
    \begin{AIbox}{Stage 2: Relevance Scoring}
    \parbox[t]{\textwidth}{
        You are an AI tasked with identifying and ranking the most salient features from meeting transcript facts. Your task is to extract and prioritize key information based on its importance for the final summary. \newline
        
        \textbf{INSTRUCTIONS}: \newline
        1.\;Analyze the provided facts carefully. \newline
        2.\;For each potential feature, first reason about its importance by considering: \newline
        \hspace*{1em}– Is this a critical decision point or major outcome? \newline
        \hspace*{1em}– Does it represent an action item or task assignment? \newline
        \hspace*{1em}– Is it a key insight or discussion point? \newline
        \hspace*{1em}– How does it contribute to the overall context? \newline
        \hspace*{1em}– What impact does this have on the meeting’s objectives? \newline
        3.\;Based on your reasoning, then: \newline
        \hspace*{1em}\textbf{a.\;Assign an importance score (1–10):} \newline
        \hspace*{2em}10: Critical decisions, major outcomes, key action items \newline
        \hspace*{2em}7–9: Important discussions, significant insights \newline
        \hspace*{2em}4–6: Supporting information, context \newline
        \hspace*{2em}1–3: Background details \newline
        \hspace*{1em}\textbf{b.\;Identify the feature type:} \newline
        \hspace*{2em}\texttt{DECISION}\,– Final choices or agreements reached \newline
        \hspace*{2em}\texttt{ACTION}\,– Tasks, assignments, or next steps \newline
        \hspace*{2em}\texttt{INSIGHT}\,– Important realizations, findings, or discussion points \newline
        \hspace*{2em}\texttt{CONTEXT}\,– Background or supporting information \newline
        4.\;Provide a certainty score (0–100 \%) indicating confidence in your assessment. \newline
        \textbf{Do Not Hallucinate}: Use only information present in the atomic facts. \newline
        \textbf{Order}: List highly important features first, followed by medium and then least important ones. \newline
        
        \textbf{STRICT OUTPUT FORMAT} – Return a \emph{valid JSON list} of objects. \newline
        Each object must contain exactly these keys (no extras, no omissions): \newline
        \{\newline
        \hspace*{1em}\texttt{"feature"}: The text of the identified salient feature. \newline
        \hspace*{1em}\texttt{"reasoning"}: Concise explanation of \emph{why} this feature matters, grounded in the atomic facts. \newline
        \hspace*{1em}\texttt{"importance\_score"}: Importance level per guidance above. \newline
        \hspace*{1em}\texttt{"feature\_type"}: One of \texttt{DECISION}, \texttt{ACTION}, \texttt{INSIGHT}, \texttt{CONTEXT}. \newline
        \hspace*{1em}\texttt{"certainty\_score"}: Confidence percentage for this assessment. \newline
        \} \newline

        \textbf{Your Task}: \newline
        Analyze the following \textbf{facts} and output a ranked JSON list of salient features that adheres \emph{strictly} to the format and rules above. \newline

        \textbf{Input Variable}: \newline
        facts: \textbf{\{facts\}} 

    }
    \end{AIbox}
    \caption{Prompt template for ranking salient features from facts.}
    \label{fig:feature_extraction_prompt}
\end{figure*}

\begin{figure*}[t]
    \begin{AIbox}{Stage 3: Outline Planning}
    \parbox[t]{\textwidth}{
        Your task is to generate an outline for a summary based on the salient features of the meeting transcript. This outline will guide the summarization process.
 \newline
        \newline
        \textbf{MEETING FEATURE CATEGORIES}: \newline
        1.\;\texttt{DECISION}\;(\,score 8–10\,)\,: Key decisions and agreements made. \newline
        2.\;\texttt{HIGH\_PRIORITY}\;(\,score 8–10\,)\,: Critical discussion points. \newline
        3.\;\texttt{MEDIUM\_PRIORITY}\;(\,score 6–7\,)\,: Important supporting points. \newline
        4.\;\texttt{CONTEXT}\,: Background information. \newline

        \textbf{RECOMMENDED STRUCTURE FOR THE OUTLINE}: \newline
        \textbf{1.\;Meeting Overview}\;(2–3 lines) \newline
        \hspace*{1em}– Main topic and key outcomes \newline
        \hspace*{1em}– Critical decisions \newline
        \textbf{2.\;Key Decisions} \newline
        \hspace*{1em}– Each decision with decision-maker(s) \newline
        \hspace*{1em}– Rationale for the decision \newline
        \hspace*{1em}– Anticipated impact \newline
        \textbf{3.\;Main Discussion Points} \newline
        \hspace*{1em}– Major topics covered \newline
        \hspace*{1em}– Important insights \newline
        \hspace*{1em}– Agreements reached \newline
        \textbf{4.\;Next Steps} \newline
        \hspace*{1em}– Action items \newline
        \hspace*{1em}– Follow-up tasks \newline
        \hspace*{1em}– Assigned responsibilities \newline

        This outline will be used to generate a coherent summary that captures: \newline
        – The flow of the meeting \newline
        – Key outcomes and decisions \newline
        – Important context and reasoning \newline
        – Next actions and responsibilities \newline

        \textbf{OUTPUT FORMAT}: Return a \emph{list of outline points}, each representing one line in the outline, following the section order above. \newline

        \textbf{Your Task}: \newline
        Create a clear, section-based outline for summary generation using the following facts: \textbf{\{important\_facts\}}. \newline
        Focus on organizing information to tell a clear story of what happened in the meeting. \newline
    }
    \end{AIbox}
    \caption{Prompt template for generating an outline from prioritized meeting facts.}
    \label{fig:outline_generation_prompt}
\end{figure*}

\begin{figure*}[t]
    \begin{AIbox}{Stage 4: Enrichment-Based Generation}
    \parbox[t]{\textwidth}{
        You are an expert summarization agent tasked with creating a \emph{structured meeting summary}. Your primary goal is to \textbf{follow the outline exactly} while using the matched facts with their contexts to provide detailed information for each outline point. \newline

        \textbf{CRITICAL CONSTRAINTS}: \newline
        – \textit{DO NOT} add any information that is not present in the provided contexts. \newline
        – \textit{DO NOT} hallucinate or infer information. \newline
        – Use \textit{ONLY} facts and contexts explicitly provided. \newline
        – Summary \textit{MUST} be \textit{at most 250 tokens}. \newline
        – Follow the outline structure \textit{exactly}. \newline

        \textbf{KEY REQUIREMENTS}: \newline
        \textbf{1.\;Outline Adherence}: \newline
        \hspace*{1em}– \textit{STRICTLY} follow the provided outline structure. \newline
        \hspace*{1em}– Address each outline point in order, maintaining its hierarchy. \newline
        \hspace*{1em}– Ensure all major sections are covered using \textit{ONLY} provided facts \& contexts. \newline
        \textbf{2.\;Using Enhanced Context}: For each outline point: \newline
        \hspace*{1em}a. Find relevant matched facts whose \textit{category/type} and \textit{importance score} suit the point. \newline
        \hspace*{1em}b. Employ the matched fact’s context verbatim; no extra interpretation or assumptions. \newline
        \textbf{3.\;Content Organization}: \newline
        \hspace*{1em}a. Begin with high-importance facts (scores 8–10). \newline
        \hspace*{1em}b. Support with medium-importance facts (scores 6–7). \newline
        \hspace*{1em}c. Add lower-importance facts only if space permits. \newline
        \textbf{4.\;Integration Guidelines}: \newline
        \hspace*{1em}– Connect ideas \emph{only} when explicitly supported. \newline
        \hspace*{1em}– Skip outline points with no matching facts; \underline{never speculate}. \newline
        \hspace*{1em}– Stay within 150–200 words. \newline
        \textbf{5.\;Special Cases}: \newline
        \hspace*{1em}a. If an outline point has \textbf{no} direct fact matches, skip it, \emph{do not invent}. \newline
        \hspace*{1em}b. If multiple facts match → prioritize by \textit{importance\_score}, remain concise. \newline
        \textbf{Presentation Rules}: \newline
        – \textit{NEVER} add information not in context. \newline
        – Format as cohesive paragraphs; \textit{no} bullet points, headers, or numbered lists. \newline
        – Produce a smooth, narrative flow covering key points from the outline. \newline

        \textbf{Your Task}: \newline
        Generate a \textit{150–200 word} meeting summary that follows \textit{this outline exactly}:
        \textbf{\{outline\}} \newline
        Use \textit{ONLY} these matched facts and their contexts for each outline point: \textbf{\{matched\_facts\}} \newline
        Previous Feedback (if any): \textbf{\{feedback\_prompt\}} \newline
    }
    \end{AIbox}
    \caption{Prompt template for generating an abstractive summary by enriching an outline with grounded facts.}
    \label{fig:summary_generation_prompt}
\end{figure*}

\begin{figure*}[t]
    \begin{AIbox}{Stage 4: Quality assurance (part 1)}
    \parbox[t]{\textwidth}{
        You are a \textbf{checker agent} evaluating a meeting summary. \newline
        \newline
        \textbf{EVALUATION CRITERIA \& POINTS}: \newline
        \textbf{1.\;Outline Adherence (maximum 4 error points)} \newline
        \hspace*{1em}– Each outline point is addressed in order. \newline
        \hspace*{1em}– No points are skipped unless \emph{no} matching facts exist. \newline
        \hspace*{1em}– Information appears under the correct outline sections. \newline
        \hspace*{1em}– Outline’s hierarchical structure is maintained. \newline
        \textbf{2.\;Content Accuracy (maximum 3 error points)} \newline
        \hspace*{1em}– Uses \underline{ONLY} provided facts and contexts. \newline
        \hspace*{1em}– No hallucinated or inferred information. \newline
        \hspace*{1em}– Facts are placed under relevant outline points. \newline
        \hspace*{1em}– Context is preserved exactly. \newline
        \textbf{3.\;Information Coverage (maximum 2 error points)} \newline
        \hspace*{1em}– High-importance facts (scores 8–10) are included. \newline
        \hspace*{1em}– Critical decisions and actions are covered. \newline
        \hspace*{1em}– Essential context is present; nothing vital is missing. \newline
        \textbf{4.\;Format Requirements (maximum 1 error points)} \newline
        \hspace*{1em}– Length is \textbf{150–200 words}. \newline
        \hspace*{1em}– Professional tone, clear and concise writing. \newline
        \hspace*{1em}– Logical flow between points. \newline
        
        \textbf{SCORING DEDUCTIONS}: \newline
        Errors such as missing outline point \emph{with} available facts, wrong information placement, hallucinated content, missing high-importance fact, incorrect context usage, or outside word limit add one error point in the respective category.

        \textbf{OUTPUT FORMAT}: Return a single JSON object with exactly: \newline
        \{\newline
        \hspace*{1em}\texttt{"confidence\_score"}: (0–100),\newline
        \hspace*{1em}\texttt{"feedback"}: "<\emph{specific issues and suggestions}>"\newline
        \} \newline
        \textit{Example default output (for reference only)}: \newline
        \{\texttt{"confidence\_score"}: 90,\; \texttt{"feedback"}: "Summary follows outline and uses provided facts correctly."\} \newline
        
        \textbf{Your Task}: \newline
        Evaluate the following summary against outline and requirements. Provide detailed feedback on any issues found and assign a \texttt{confidence\_score}. \newline
        
        Outline to Follow: \textbf{\{outline\}} \newline
        Available Facts and Contexts: \textbf{\{summary\_input['matched\_information']\}} \newline
        Unmatched Features: \textbf{\{summary\_input['unmatched\_features']\}} \newline
        Generated Summary: \textbf{\{generated\_summary\}} \newline
    }
    \end{AIbox}
    \caption{Prompt template to validate a generated meeting summary.}
    \label{fig:summary_validation_prompt}
\end{figure*}

\begin{figure*}[t]
    \begin{AIbox}{Stage 4: Quality assurance (part 2)}
    \parbox[t]{\textwidth}{
        You are an expert editor tasked with refining a document summary. Your goal is to create a polished final summary \textbf{maximum 250 tokens} that flows naturally and maintains a coherent narrative throughout. \newline

        \textbf{CRITICAL REFINEMENT GUIDELINES}: \newline
        1.\;{\textbf{Focus on Narrative Flow}} \newline
        \hspace*{1em}– Craft smooth, logical transitions between points. \newline
        \hspace*{1em}– Establish clear relationships among facts; use appropriate discourse markers. \newline
        \hspace*{1em}– Avoid abrupt topic jumps. \newline
        2.\;{\textbf{Maintain Topical Coherence}} \newline
        \hspace*{1em}– Group related information; follow a logical progression. \newline
        \hspace*{1em}– Use topic sentences to introduce new conceptual areas. \newline
        3.\;{\textbf{Concision and Length}} \newline
        \hspace*{1em}– Produce \textbf{at most 250 tokens}. \newline
        \hspace*{1em}– Eliminate redundancies while preserving key details. \newline
        4.\;{\textbf{Style and Clarity}} \newline
        \hspace*{1em}– Keep consistent tense and professional tone. \newline
        \hspace*{1em}– Remove empty phrases (e.g., ``The summary is…''). \newline

        \textbf{OUTPUT GUIDELINES}: \newline
        – Present the final summary as \textbf{1–2 well-structured paragraphs}. \newline
        – Include \underline{only} information from the original content; no new facts. \newline
        – Ensure a cohesive reading experience. \newline

        \textbf{WHAT TO AVOID}: \newline
        – Abrupt transitions or disconnected statements. \newline
        – Introducing new or inferred information. \newline
        – Excessive focus on one topic at the expense of others. \newline
        – Uneven or inconsistent coverage. \newline

        \textbf{Your Task}: \newline
        Refine the following document summary \textbf{\{combined\_summary\}} into a polished, single narrative that: \newline
        1.\;Maintains all key information. \newline
        2.\;Flows naturally with smooth transitions. \newline
        3.\;Presents a coherent storyline. \newline
        4.\;Contains \textbf{at most 250 tokens}. \newline
        5.\;Reads as a single, unified piece. \newline
    }
    \end{AIbox}
    \caption{Prompt template for refining the meeting summary.}
    \label{fig:summary_refinement_prompt}
\end{figure*}

\begin{figure*}[t]
  \begin{AIbox}{\reasoning{}: Exploration and Fact Selection}
    \parbox[t]{\textwidth}{
      You are tasked to \textbf{embody the following persona} and select the facts from a meeting that are most relevant \emph{to that persona}. Answer solely from the persona’s perspective—never refer to yourself or mention that you are role-playing. \newline

      \textbf{Persona Profile}: \textbf{\{character\_profile\}} \newline

      There was just a meeting, and someone has provided you with a list of facts extracted from the transcript. Your job is to identify the key takeaways \emph{you} would care about. \newline

      \textbf{IMPORTANT – Your response \textit{must include TWO parts}:} \newline
      \textbf{Part 1 – Detailed Reasoning Process (think aloud)} \newline
      Address every point below in order, narrating your thoughts: \newline
      \hspace*{1em}(1) What prior knowledge do you have? \newline
      \hspace*{1em}(2) Which project are you currently working on? \newline
      \hspace*{1em}(3) What are your primary interests and goals? \newline
      \hspace*{1em}(4)  Read each fact carefully and think about which information is most relevant to you in your role.\\
      \hspace*{2.2em} Explain why. \newline
      \hspace*{1em}(5) Is there an urgency or priority that aligns particularly closely with your current responsibilities or\\
      \hspace*{2.2em} known concerns? \newline
      \hspace*{1em}(6) Which information might require simplification or additional context to ensure clear \\
      \hspace*{2.2em} comprehension? \newline
      \hspace*{1em}(7) You've selected information that you consider important. Review this selection once more and \\
      \hspace*{2.2em} provide concrete examples explaining why these details are relevant for you. \newline
      \hspace*{1em}(8) Now, go through the list a second time and identify which information you consider irrelevant or\\
      \hspace*{2.2em} unimportant for your role/persona, providing reasons for your decisions. \newline
      \hspace*{1em}(9) Are there any topics you found difficult to classify or about which you felt unsure? \\
      \hspace*{2.2em} If so, what are they? \newline
      Share \emph{all} reflections, even minor ones. \newline

      \textbf{Part 2 – Structured JSON Output} \newline
      After reasoning, output your final selection in exactly this format: \newline
      \texttt{[} \newline
      \hspace*{1em}\texttt{\{} \newline
      \hspace*{2em}\texttt{"fact"}:\ \{\,\texttt{"fact"},\ \texttt{"context"},\ \texttt{"verbose\_context"}\,\},\newline
      \hspace*{2em}\texttt{"certainty\_score"}:\ (0–100)\newline
      \hspace*{1em}\texttt{\}}, \ldots \newline
      \texttt{]} \newline
      \textbf{Rules for the JSON list:} \newline
      – Copy the \texttt{"fact"}, \texttt{"context"}, and \texttt{"verbose\_context"} \emph{verbatim}. \newline
      – Include only items where \texttt{certainty\_score $\geq$ 40}. \newline
      – Sort by \texttt{certainty\_score} descending. \newline
      \newline
      \textbf{CRITICAL ALERTS}: \newline
      • Do \emph{not} add, alter, or summarise any field inside the fact objects. \newline
      • Never hallucinate or assume information beyond the transcript.

      \textbf{Atomic Facts Provided}: \textbf{\{atomic\_facts\}} \newline
    }
  \end{AIbox}
  \caption{Prompt template for the reasoning-out-loud task proceeding the actual fact selection.}
  \label{fig:persona_filter_prompt}
\end{figure*}

\begin{figure*}[t]
  \begin{AIbox}{Persona-Focused Salient Feature Extraction \& Ranking - Part 1}
    \parbox[t]{\textwidth}{
      You are an expert at identifying and ranking important features from meeting transcripts \emph{while considering specific persona preferences}. Your goal is to extract and prioritize information that would be most relevant and valuable to the given persona, \textbf{especially focusing on what \emph{others} said that the persona needs to know or act upon}. \newline
      \newline
      \textbf{CRITICAL PERSPECTIVE SHIFT}: \newline
      – \textbf{Prioritize} information spoken by \emph{others} that is relevant to the persona. \newline
      – \textbf{De-prioritize} information spoken by the persona themselves (they already know this). \newline
      – Ask: “What would this persona want to know from the meeting?” \newline
      – Focus on insights, actions, requests, and decisions from others that affect the persona’s role. \newline
      
      \textbf{INSTRUCTIONS}: \newline
      1.\;Analyze \textbf{atomic facts} \emph{and} \textbf{persona preferences} carefully. \newline
      2.\;For each potential feature, evaluate: \newline
      \hspace*{1em}\textbf{a.\;General Importance (1–10)} \newline
      \hspace*{2em}8–10: Critical decisions/outcomes affecting the persona \newline
      \hspace*{2em}6–7: Important discussions relevant to persona’s role \newline
      \hspace*{2em}3–5: Supporting details needed by the persona \newline
      \hspace*{2em}1–2: Background info useful to the persona \newline
      \hspace*{1em}\textbf{b.\;Persona Alignment (1–10)} — match with persona’s project interests, decision factors, information needs, background, priorities, and knowledge gaps. \newline
      \hspace*{1em}\textbf{c.\;Feature Type}: \texttt{DECISION}, \texttt{ACTION}, \texttt{INSIGHT}, \texttt{CONTEXT}. \newline
      \hspace*{1em}\textbf{d.\;Certainty Score (0–100 \%)} — your confidence in the assessment. \newline

      [continue in part 2]
    }
  \end{AIbox}
  \caption{Prompt template for extracting and ranking meeting features that are most relevant to a specific persona, emphasizing information provided by others.}
  \label{fig:persona_salient_feature_prompt_part1}
\end{figure*}

\begin{figure*}[t]
  \begin{AIbox}{Persona-Focused Salient Feature Extraction \& Ranking - Part 2}
    \parbox[t]{\textwidth}{

    [continue from part 1]
      
      3.\;Provide \emph{detailed reasoning}: \newline
      – Why is this feature important to the persona? \newline
      – How does information from \emph{others} align with persona’s needs? \newline
      – What context makes it actionable for the persona? \newline
      \textbf{OUTPUT FORMAT – Return a JSON \underline{list}}: \newline
      \texttt{\{} \newline
      \hspace*{1em}\texttt{"feature"}:\ "Extracted feature text",\newline
      \hspace*{1em}\texttt{"reasoning"}:\ "Why this is important to the persona",\newline
      \hspace*{1em}\texttt{"importance\_score"}:\ 1–10,\newline
      \hspace*{1em}\texttt{"persona\_alignment\_score"}:\ 1–10,\newline
      \hspace*{1em}\texttt{"feature\_type"}:\ "DECISION/ACTION/INSIGHT/CONTEXT",\newline
      \hspace*{1em}\texttt{"certainty\_score"}:\ 0–100,\newline
      \hspace*{1em}\texttt{"alignment\_explanation"}:\ "How/why this aligns with persona’s information needs"\newline
      \texttt{\}} \newline

      \textbf{CRITICAL RULES}: \newline
      – Use \underline{only} facts from the provided atomic facts. \newline
      – Base alignment \underline{strictly} on the provided persona preferences. \newline
      – \underline{Prioritize} information from others; \underline{de-prioritize} what the persona already said. \newline
      – \underline{No} hallucination or inference beyond provided data. \newline
      – Explain \emph{all} scoring decisions. \newline
      – Order features by combined importance and alignment scores. \newline

      \textbf{Your Task}: \newline
      Analyze the following persona and atomic facts to generate a ranked list of features. \newline
      \textbf{Character Sheet (Persona Preferences)}: \textbf{\{character\_sheet\}} \newline
      \textbf{Atomic Facts}: \textbf{\{atomic\_facts\}} \newline
    }
  \end{AIbox}
  \caption{Prompt template for extracting and ranking meeting features that are most relevant to a specific persona, emphasizing information provided by others.}
  \label{fig:persona_salient_feature_prompt_part2}
\end{figure*}

\begin{figure*}[t]
  \begin{AIbox}{Persona-Focused Outline Generation}
    \parbox[t]{\textwidth}{
      Create a personalized outline from ranked features that balances overall importance with persona-alignment scores, \textbf{focusing on what \emph{others} said that the persona needs to know}. Think of the result as \emph{“notes the persona would take for themselves.”} \newline
      \newline
      \textbf{CRITICAL PERSPECTIVE SHIFT}: \newline
      – Highlight insights, actions, and decisions voiced by \emph{others} that affect the persona. \newline
      – De-emphasize information provided by the persona; they already know it. \newline
      – Prioritize features with high persona-alignment. \newline
      – Group items by the persona’s interests and preferred detail level. \newline
      \newline
      \textbf{OUTLINE STRUCTURE}: \newline
      \textbf{1.\;Critical Information for the Persona} (combined score 8–10) \newline
      \hspace*{1em}– Key decisions by others that affect the persona \newline
      \hspace*{1em}– Actions required of the persona \newline
      \hspace*{1em}– Important insights from others relevant to the persona’s role \newline
      \textbf{2.\;Important Considerations} (combined score 6–7) \newline
      \hspace*{1em}– Relevant discussions initiated by others \newline
      \hspace*{1em}– Context for understanding key decisions \newline
      \hspace*{1em}– Information supporting the persona’s responsibilities \newline
      \textbf{3.\;Supporting Information} (combined score 4–5) \newline
      \hspace*{1em}– Background details that enhance understanding \newline
      \hspace*{1em}– Additional context matching persona information needs \newline
      \textbf{4.\;Additional Context} (combined score 1–3, only if needed) \newline
      \hspace*{1em}– Low-priority items that clarify higher sections \newline
      \newline
      \textbf{GUIDELINES}: \newline
      – Sort features by combined importance + alignment scores. \newline
      – Use feature type, certainty, and alignment explanations to decide placement. \newline
      – Skip items that do not serve the persona’s needs. \newline
      – Keep the outline hierarchical and concise. \newline
      \newline
      \textbf{Your Task}: \newline
      Using the following persona-aligned features \textbf{\{important\_features\}}, generate a clear hierarchical outline focused on what the persona needs to learn from others. \newline
    }
  \end{AIbox}
  \caption{Prompt template for building a personalized outline that emphasizes information voiced by others and most relevant to the target persona.}
  \label{fig:persona_outline_prompt}
\end{figure*}

\begin{figure*}[t]
  \begin{AIbox}{Persona-Focused Summary Generation}
    \parbox[t]{\textwidth}{
      You are an expert summarization agent tasked with creating a \textbf{highly personalized} meeting summary \emph{from the persona’s perspective}, focusing on what \textbf{others} said that is relevant to the persona. \newline
      \newline
      \textbf{CRITICAL PERSPECTIVE SHIFT}: \newline
      – Do \underline{not} recap what the persona said; concentrate on insights, actions, decisions, and requests voiced by \emph{others}. \newline
      – Treat the output as “meeting notes” the persona would write for themselves. \newline
      – Skip anything the persona already knows or presented. \newline
      \newline
      \textbf{CRITICAL CONSTRAINTS}: \newline
      – Length \textbf{150–200 words}. \newline
      – Use \underline{only} the provided facts and contexts—no hallucination. \newline
      – Follow the outline conceptually, yet present as cohesive paragraphs (no bullets, numbers, or headers). \newline
      \newline
      \textbf{PERSONA CONSIDERATIONS}: \newline
      1.\;Match their information preferences (detail level, format, key interests). \newline
      2.\;Emphasize their decision factors (values, risk tolerance, time sensitivity). \newline
      3.\;Supply context they require (technical depth, background, rationale). \newline
      \newline
      \textbf{SUMMARIZATION GUIDELINES}: \newline
      1.\;For each outline point, pick facts with the highest combined importance + alignment scores that impact the persona. \newline
      2.\;Highlight: \newline
      \hspace*{1em}– Decisions by others affecting the persona. \newline
      \hspace*{1em}– Actions/requests directed at the persona. \newline
      \hspace*{1em}– Insights requiring the persona’s input or expertise. \newline
      3.\;Provide only context the persona needs; omit superfluous detail. \newline
      4.\;Skip outline points lacking matched facts; never invent content. \newline
      5.\;Ensure smooth narrative flow and logical transitions. \newline
      \newline
      \textbf{OUTPUT FORMAT}: One or two cohesive paragraphs (no bullets, headers, or lists). \newline
      \newline
      \textbf{Your Task}: \newline
      Produce a \textbf{150–200-word} summary for this persona using: \newline
      – \textbf{Character Sheet}: \textbf{\{character\_sheet\}} \newline
      – \textbf{Outline (conceptual guide)}: \textbf{\{outline\}} \newline
      – \textbf{Matched Facts \& Contexts}: \textbf{\{summary\_input['matched\_information']\}} \newline
      – \textbf{Unmatched Features (optional)}: \textbf{\{summary\_input['unmatched\_features']\}} \newline
      – \textbf{Previous Feedback (if any)}: \textbf{\{feedback\_prompt\}} \newline
      \newline
      Remember to: focus on what \emph{others} said, match the persona’s style and priorities, stay within the word limit, and write a smooth, paragraph-style narrative only. \newline
    }
  \end{AIbox}
  \caption{Prompt template for generating a 150–200 word meeting summary tailored to a specific persona, emphasizing information provided by others and formatted as cohesive notes.}
  \label{fig:persona_summary_prompt}
\end{figure*}

\begin{figure*}[t]
  \begin{AIbox}{Persona-Focused Summary Validation (Checker Agent)}
    \parbox[t]{\textwidth}{
      You are a \textbf{checker agent} evaluating a personalized meeting summary. \newline
      \newline
      \textbf{EVALUATION CRITERIA \& POINT VALUES}: \newline
      \textbf{1.\;Persona Alignment (40 pts)} \newline
      \hspace*{1em}– Focuses on what \emph{others} said that is relevant to the persona (not what the persona said). \newline
      \hspace*{1em}– Addresses the persona’s key interests and information needs. \newline
      \hspace*{1em}– Highlights actions, insights, and decisions from others that affect the persona’s role. \newline
      \hspace*{1em}– Uses the persona’s preferred style and detail level. \newline
      \textbf{2.\;Content Organization (30 pts)} \newline
      \hspace*{1em}– Follows the outline structure. \newline
      \hspace*{1em}– Prioritizes meeting information that matters to the persona. \newline
      \hspace*{1em}– Provides appropriate context and maintains logical flow. \newline
      \textbf{3.\;Information Accuracy (20 pts)} \newline
      \hspace*{1em}– Uses \underline{only} the provided facts and contexts. \newline
      \hspace*{1em}– No hallucinated or misplaced content; context is accurate. \newline
      \textbf{4.\;Format Requirements (10 pts)} \newline
      \hspace*{1em}– Length is 150–200 words. \newline
      \hspace*{1em}– Professional tone at the persona’s technical level. \newline
      \hspace*{1em}– Clear, well-organized narrative (no bullets or headers). \newline
      \newline
      \textbf{SCORING DEDUCTIONS}: \newline
      – Focus on persona’s own speech, not what they need to know (–15) \newline
      – Missing key insights from others that affect the persona (–15) \newline
      – Poor persona alignment (–10) \newline
      – Missing key persona interests (–10) \newline
      – Wrong information placement (–8) \newline
      – Hallucinated content (–15) \newline
      – Inappropriate context (–8) \newline
      – Outside word limit (–10) \newline
      \newline
      \textbf{OUTPUT FORMAT – Return exactly:} \newline
      \texttt{\{} \newline
      \hspace*{1em}\texttt{"confidence\_score"}:\ (0–100),\newline
      \hspace*{1em}\texttt{"feedback"}:\ "<\emph{specific issues and suggestions}>"\newline
      \texttt{\}} \newline
      \newline
      \textbf{Your Task}: \newline
      Evaluate the personalized summary below against the criteria above and provide detailed feedback on perspective alignment and persona relevance. \newline
      \newline
      \textbf{Character Sheet (Persona)}:\ \textbf{\{character\_sheet\}} \newline
      \textbf{Outline Structure}:\ \textbf{\{outline\}} \newline
      \textbf{Available Facts \& Contexts}:\ \textbf{\{summary\_input['matched\_information']\}} \newline
      \textbf{Unmatched Features}:\ \textbf{\{summary\_input['unmatched\_features']\}} \newline
      \textbf{Generated Summary}:\ \textbf{\{generated\_summary\}} \newline
      \newline
    }
  \end{AIbox}
  \caption{Prompt template for a checker agent to validate a persona-oriented meeting summary, scoring alignment, organization, accuracy, and format, and returning a JSON evaluation.}
  \label{fig:persona_summary_validation_prompt}
\end{figure*}


\begin{figure*}[t]
  \begin{AIbox}{Single-Call Meeting Summarizer for Specific Reader (Reader Tailoring)}
    \parbox[t]{\textwidth}{
      You are an expert summarization agent. Your task is to summarize a meeting transcript for a specific reader, focusing on what would be most relevant and important to them. \newline
      \newline
      \textbf{Your Task:} Summarize the following meeting for a specific reader. \newline
      \newline
      \textbf{Reader Details:} \{character\_sheet\} \newline
      \textbf{Meeting Transcript:} \{transcript\} \newline
      \newline
      \textbf{Instructions:} \newline
      – Summarize the meeting in 150–200 words. \newline
      – Focus on what would be most relevant for this specific reader. \newline
      – Emphasize what \emph{other} people said that the reader would find important. \newline
      – Use paragraph format (no bullet points or headers). \newline
      – Only include information explicitly mentioned in the transcript. \newline
    }
  \end{AIbox}
  \caption{Prompt template for a single-call meeting summarizer tailored to a specific reader, emphasizing relevance to their interests.}
  \label{fig:single_llm_reader_summary_prompt}
\end{figure*}

\begin{figure*}[t]
  \begin{AIbox}{Single LLM Personalized Meeting Summarizer (Role Playing)}
    \parbox[t]{\textwidth}{
      Take the role of the given persona and summarize the meeting \textit{from their perspective}, focusing on what \textbf{others} said that is relevant and important to the persona. \newline
      \newline
      \textbf{CRITICAL PERSPECTIVE SHIFT}: \newline
      – The summary is \underline{not} about what the persona said or did. \newline
      – Emphasize insights, actions, decisions, and requests voiced by \emph{others} that affect the persona’s role. \newline
      – Treat the output as “meeting notes” that the persona would write for themselves—exclude anything they already know or have presented. \newline
      \newline
      \textbf{CRITICAL CONSTRAINTS}: \newline
      – Length \textbf{strictly 150–200 words}. \newline
      – Use \underline{only} facts found in the transcript; no hallucinations. \newline
      – Present as cohesive paragraphs—\emph{no} bullet points, numbers, or headers. \newline
      \newline
      \textbf{PERSONA CONSIDERATIONS}: \newline
      1.\;Information Preferences — match detail level, format, and key interests. \newline
      2.\;Decision Factors — emphasize values, risk tolerance, and time sensitivity. \newline
      3.\;Context Requirements — supply background at the persona’s technical level to support decision-making. \newline
      \newline
      \textbf{STYLE AND FORMAT}: \newline
      – Smooth narrative with logical transitions. \newline
      – Appropriate technical register for the persona. \newline
      – Highlight only what is new and actionable for the persona. \newline
      – Remain within the 150–200-word limit. \newline
      \newline
      \textbf{Your Task}: \newline
      \textbf{Generate a 150–200-word personalized meeting summary} for the following persona, based solely on the transcript.  Remember to focus on what \emph{others} said that matters to this persona and write it as their own meeting notes. \newline
      \newline
      \textbf{Character Sheet (Persona Preferences)}: \textbf{\{character\_sheet\}} \newline
      \textbf{Meeting Transcript}: \textbf{\{transcript\}} \newline
    }
  \end{AIbox}
  \caption{Prompt template for a single-call personalized meeting summarizer that produces 150–200-word “notes” from the persona’s perspective, emphasizing information provided by others.}
  \label{fig:single_llm_personalized_prompt}
\end{figure*}

\begin{figure*}[t]
  \begin{AIbox}{Single LLM Personalized Meeting Summarizer with \reasoning{}}
    \parbox[t]{\textwidth}{
      You are an expert summarization agent tasked with creating a highly personalized meeting summary. Your goal is to produce a summary from the \emph{perspective of the persona}, focusing on what \textbf{others} said that is relevant and important for this specific persona. \newline
      \textbf{PROCESS}: \newline
      1.\;First, conduct a detailed reasoning process to determine which information will be most relevant. \newline
      2.\;Then, based on that reasoning, generate the final personalized summary. \newline
      \textbf{CRITICAL PERSPECTIVE SHIFT}: \newline
      – The summary is \emph{not} about what the persona said or did. \newline
      – Emphasize insights, actions, decisions, and requests voiced by \emph{others} that affect the persona’s role. \newline
      – Treat the output as “meeting notes” the persona would write for themselves—omit anything they already know or presented. \newline
      \textbf{CRITICAL CONSTRAINTS}: \newline
      – Final summary \textbf{must} be 150–200 words. \newline
      – Use \emph{only} facts provided in the transcript—no hallucinations or inferences. \newline
      – Present as cohesive paragraphs; \emph{no} bullet points, headers, or lists. \newline
      \textbf{REASONING LAYER}: \newline
      \emph{Step 1: Detailed Reasoning Process (reason out loud)} \newline
      \hspace*{1em}(1) What prior knowledge do you have? \newline
      \hspace*{1em}(2) Which project are you currently working on? \newline
      \hspace*{1em}(3) What are your primary interests and goals? \newline
      \hspace*{1em}(4)  Read each fact carefully and think about which information is most relevant to you in your role.\\
      \hspace*{2.2em} Explain why. \newline
      \hspace*{1em}(5) Is there an urgency or priority that aligns particularly closely with your current responsibilities or\\
      \hspace*{2.2em} known concerns? \newline
      \hspace*{1em}(6) Which information might require simplification or additional context to ensure clear \\
      \hspace*{2.2em} comprehension? \newline
      \hspace*{1em}(7) You've selected information that you consider important. Review this selection once more and \\
      \hspace*{2.2em} provide concrete examples explaining why these details are relevant for you. \newline
      \hspace*{1em}(8) Now, go through the list a second time and identify which information you consider irrelevant or\\
      \hspace*{2.2em} unimportant for your role/persona, providing reasons for your decisions. \newline
      \hspace*{1em}(9) Are there any topics you found difficult to classify or about which you felt unsure? \\
      \hspace*{2.2em} If so, what are they? \newline
      Share your full thought process, even minor reflections. \newline
      \newline
      \emph{Step 2: Final Summary} \newline
      Generate a \textbf{150–200 word} personalized meeting summary that: Focuses on what \emph{others} said that is relevant to the persona (not what the persona said); Highlights insights, actions, decisions, and requests from others that affect the persona’s role; Matches the persona’s preferred style and detail level; Is formatted as cohesive paragraphs that flow naturally.
        \textbf{OUTPUT FORMAT – Return exactly:} \newline
      \texttt{\{} \newline
      \hspace*{1em}\texttt{"Full\_reasoning\_response"}:\ <YOUR DETAILED CHAIN-OF-THOUGHT HERE>,\newline
      \hspace*{1em}\texttt{"Summary"}:\ "<\emph{<YOUR FINAL 150–200 WORD PARAGRAPH SUMMARY HERE>}>"\newline
      \texttt{\}} \newline
      \textbf{INPUT FORMAT}: \newline
      \textbf{Character Sheet (Persona Preferences)}: \textbf{\{character\_sheet\}} \newline
      \textbf{Meeting Transcript}: \textbf{\{transcript\}} 
    }
  \end{AIbox}
  \caption{Prompt template for a single-call personalized meeting summarizer with an explicit reasoning layer and strict JSON output of reasoning and summary.}
  \label{fig:personalized_summary_with_reasoning_prompt}
\end{figure*}

\subsection{Personalization Approach Comparison}
\label{app:personalization-approach}

To evaluate the effectiveness of \reasoning{}'s reasoning protocol, we compare it against two alternative personalization approaches commonly used in current research:

\begin{itemize}
    \item \textbf{Reader-Tailoring}: The model receives a description of the target reader and is instructed to customize content for their needs, without explicit reasoning about fact selection (see prompt \Cref{fig:single_llm_reader_summary_prompt}).
    
    \item \textbf{Role-Playing}: The model is instructed to embody a specific persona and directly generate a summary from that perspective, without explicit reasoning steps (see prompt \Cref{fig:single_llm_personalized_prompt}).
\end{itemize}

We implemented each approach using a single LLM, isolating the effect of fact extraction.
\Cref{tab:personalization_comparison} presents the evaluation results across both personalization-specific (P-MESA) and general quality (MESA) dimensions.

\reasoning{} demonstrates consistent advantages across most \pmetric dimensions.
The largest improvements appear in completeness (2 points better than Reader-Tailoring, 1 point better than Role-Playing) and knowledge-level fit (0.5 points better than both alternatives).
While Reader-Tailoring achieves comparable factuality scores (both at 2), \reasoning{} outperforms Role-Playing by 2 points in this dimension.
These results indicate that \reasoning{}'s explicit reasoning enhances the model's ability to include relevant information and adapt its presentation to the reader's knowledge.

For general quality dimensions measured by MESA, \reasoning{} shows notable improvements in hallucination reduction (1-2 points better) and omission (1.5 points better) compared to alternative approaches.
This suggests that explicit reasoning not only improves personalization but also enhances factual accuracy and information coverage.
Role-playing demonstrates stronger performance in structural coherence but shows weaknesses in factuality and hallucination.
Reader-Tailoring achieves reasonable factuality but struggles with completeness and contextual framing.
These patterns suggest that while alternative approaches may excel in specific dimensions, \reasoning{}'s structured reasoning provides the most balanced and comprehensive approach to personalization, addressing both content selection and presentation aspects.

\begin{table}[ht]
\scriptsize
\centering
\begin{tabular}{lccc}
\toprule

\textbf{Approach} & \textbf{Reader-Tailoring} & \textbf{Roleplaying} & \textbf{\reasoning{}} \\
\midrule
\rowcolor{gray!20} 
\multicolumn{4}{c}{\textbf{P-MESA} (lower is better)} \\
\midrule
        goal alignment
            & $3_{\textit{0.80}}$ 
            & $3_{\textit{0.00}}$
            & \cellcolor{highlightGreen}$2.5_{\textit{0.41}}$ \\
        completeness     
            & $5_{\textit{0.88}}$
            & $4_{\textit{0.58}}$
            & \cellcolor{highlightGreen}$3_{\textit{0.10}}$ \\
        factuality     
            & \cellcolor{highlightGreen}$2_{\textit{1.53}}$
            & $4_{\textit{0.58}}$
            & \cellcolor{highlightGreen}$2_{\textit{0.37}}$ \\
        knowledge level fit     
            & $2_{\textit{0.40}}$
            & $2_{\textit{0.53}}$
            & \cellcolor{highlightGreen}$1.5_{\textit{1.73}}$ \\
        priority structuring     
            & $4_{\textit{0.76}}$
            & $4_{\textit{0.60}}$
            & $4_{\textit{0.58}}$ \\
        contextual framing     
            & $5_{\textit{1.23}}$
            & $4_{\textit{1.08}}$
            & \cellcolor{highlightGreen}$3.5_{\textit{0.71}}$ \\
        relevance     
            & $3_{\textit{0.68}}$
            & $3_{\textit{1.16}}$
            & \cellcolor{highlightGreen}$2.5_{\textit{0.70}}$ \\
\midrule
\rowcolor{gray!20} 
\multicolumn{4}{c}{\textbf{MESA} (lower is better)} \\
\midrule
Coreference     & $0_{\textit{1.74}}$ & $0_{\textit{1.73}}$     & $0_{\textit{0.00}}$ \\
Hallucination   & $3_{\textit{2.52}}$ & $4_{\textit{1.15}}$   & \cellcolor{highlightGreen}$2_{\textit{2.52}}$ \\
Incoherence     & $4_{\textit{1.74}}$ & $4_{\textit{0.58}}$     & $4_{\textit{2.31}}$ \\
Irrelevance     & $2_{\textit{0.00}}$ & $2_{\textit{0.14}}$   & $2_{\textit{1.52}}$ \\
Language        & $4_{\textit{1.51}}$ & $4_{\textit{0.56}}$   & $4_{\textit{2.08}}$ \\
Omission        & $4_{\textit{1.25}}$ & $4_{\textit{0.27}}$     & \cellcolor{highlightGreen}$2.5_{\textit{0.00}}$ \\
Repetition      & \cellcolor{highlightGreen}$3_{\textit{1.00}}$ & $4_{\textit{0.58}}$     & \cellcolor{highlightGreen}$3_{\textit{0.58}}$ \\
Structure       & $4_{\textit{0.78}}$ & \cellcolor{highlightGreen}$3_{\textit{1.53}}$     & $4_{\textit{1.72}}$ \\
\bottomrule
\end{tabular}
\caption{Impact of different personalization approaches on summary quality of a single GPT-4o instance. Values are Median$_{Std}$. MESA and \pmetric{} scores are 1--5 Likert ratings. \colorbox{highlightGreen}{Green} is best in category.}
\label{tab:personalization_comparison}
\end{table}

\section{Literature Review and Criteria Selection}
\label{app:pmetric_lit_review}

This section details our systematic development of the seven \pmetric{} dimensions, expanding on the three-step process outlined in \Cref{subsec:taxonomy_construction}.
We provide comprehensive documentation of our literature review methodology, human evaluation protocol, and dimension refinement process.

\subsection{Step 1: Literature Review}
\label{app:literature_review}

\paragraph{Corpus Construction}
We collected 50 papers published between 2018 and 2024 from major computational linguistics venues, including ACL, EMNLP, NAACL, COLING, EACL, and TACL.
Our primary source was Semantic Scholar, with results ranked via the Allen AI Paper Finder \citep{AllenAI25}.
We use the terms ``personalization,'' ``adaptation,'' and ``user modeling'' as anchor terms, combined with targeted keywords such as ``generation,'' ``summarization,'' ``data-to-text,'' ``evaluation,'' and ``language model.''

\paragraph{Relevance Assessment}
Each paper undergoes an initial screening based on title and abstract.
We classify 25 papers as \textit{highly relevant} (core focus on personalization in generation), 16 as \textit{relevant} (discuss personalization but not methodologically central), and discard 9 papers that focused on adjacent topics like personalized agents or style transfer without clear evaluative criteria.
After subsequent full-text screening, we removed an additional 5 papers that focused on end-to-end personalization or task-specific fine-tuning without analyzing personalization mechanisms, resulting in 36 papers for final analysis.

\paragraph{Candidate Dimension Extraction}
From the 36 selected works, we identify recurring characteristics of effective content personalization. 
We clustered these into nine candidate dimensions covering different aspects of personalization quality:

\begin{enumerate}
    \item \textit{Factual Accuracy}: Correspondence between summary content and source text
    \item \textit{Content Completeness}: Inclusion of information required by the target reader
    \item \textit{Information Relevance}: Focus on content pertinent to user's role and needs
    \item \textit{Objective Alignment}: Addressing high-level user goals and intentions
    \item \textit{Content Prioritization}: Ordering information by importance to the user
    \item \textit{Knowledge Appropriateness}: Matching the user's expertise level
    \item \textit{Contextual Framing}: Providing necessary background information
    \item \textit{Personal Preferences}: Matching stylistic and format preferences
    \item \textit{Information Utility}: Providing actionable content for the user's context
\end{enumerate}

These dimensions represented the standard evaluation criteria used across personalization research, though with varying terminology across different domains.

\subsection{Step 2: Human Study}
\label{subsec:human_study}

To test whether these nine criteria are distinguishable, applicable across meeting domains, and comprehensive in capturing personalization failures, we conduct a refinement study with human annotators evaluating model-generated personalized summaries.

\paragraph{Evaluation Dataset Construction}
We create a one-time evaluation dataset of 48 personalized summaries using GPT-4o and Gemini 1.5 Pro, divided equally between two personalization approaches, i.e., 24 summaries where the model was instructed to \textit{summarize for} a specific reader and 24 summaries where the model was instructed to \textit{simulate} being the target reader.

Each summary was generated for a distinct reader profile specifying role, prior knowledge level, and goals.
The samples were drawn equally from QMSum and FAME datasets, covering eight different meeting types, 14 diverse topics, and transcripts with an average of 5.9 speakers per meeting.
\Cref{tab:evaluation_dataset} shows the distribution of the dataset.

\paragraph{Annotation Protocol}
Three annotators (ages 22–29, C1+ English proficiency; see \Cref{app:human_evaluation} for details) evaluate each summary using the nine candidate dimensions on a 1-5 Likert scale.
Each summary receives annotations from two independent evaluators to enable assessment of inter-annotator reliability. 
Annotators provided structured feedback on:
\begin{itemize}
    \item Definition clarity for each dimension
    \item Potential overlap between dimensions
    \item Missing aspects of personalization not covered by the dimensions
    \item Examples of successful and unsuccessful personalization for each dimension
\end{itemize}

We collect this feedback through standardized evaluation forms and conduct daily group discussions throughout the annotation period to align understanding and address emerging questions.

\begin{table}[ht]
    \centering
    \small
    \begin{tabular}{lcc}
        \toprule
        \rowcolor{gray!20}
        \textbf{Dataset Characteristic} & \textbf{Count} \\
        \midrule
        Total Summaries & 48\\
        \midrule
        \textit{Generation Approach} & \\
        \hspace{2mm} Summarize For + GPT & 12 (6 QMSum, 6 FAME) \\
        \hspace{2mm} Summarize For + Gemini & 12 (6 QMSum, 6 FAME)\\
        \hspace{2mm} Simulate + GPT & 12 (6 QMSum, 6 FAME)\\
        \hspace{2mm} Simulate + Gemini & 12 (6 QMSum, 6 FAME) \\
        \bottomrule
    \end{tabular}
    \caption{Characteristics of the evaluation dataset used for \pmetric{} dimension refinement.}
    \label{tab:evaluation_dataset}
\end{table}

\subsection{Step 3: Final Dimension Set}
\label{subsec:final_dimensions}

Based on annotator feedback and empirical assessment, we make two refinements to the candidate dimensions:

\begin{enumerate}
    \item \textbf{Removal of Personal Preferences}: This dimension shows poor cross-context generalizability and inconsistent applicability across different meeting domains due to missing information in the datasets.
    
    \item \textbf{Consolidation of Intent-Related Dimensions}: The original \textit{Objective Alignment} and \textit{Information Utility} dimensions are merged into a single \textit{Goal Alignment} dimension.
    This consolidated dimension focuses on macro-level intent alignment, specifically whether the summary effectively addresses the reader's overarching goals.
    In contrast, the separate \textit{Priority Structuring} dimension addresses micro-level content organization, including the salience and ordering of specific information.
\end{enumerate}

No additional dimensions are identified as missing during the feedback process.
This refinement yields the final set of seven personalization dimensions used in P-MESA: Factuality, Completeness, Relevance, Goal Alignment, Priority Structuring, Knowledge-Level Fit, and Contextual Framing.
Complete definitions and indicators for these dimensions are provided in \Cref{tab:p-mesa_criteria}.

\Cref{tab:combined_p-mesa} in \Cref{sec:pmesa}  shows the correlation between \pmetric's automated scores and human judgments across these seven dimensions, demonstrating strong alignment (Spearman's $\rho$ ranging from 0.69 to 0.81, with average $\rho$ = 0.77).
A complete archive of our literature review coding scheme, annotation guidelines, and the final P-MESA implementation will be made available via GitHub (see \Cref{app:repository}).

\begin{table*}[t]
    \centering
    \renewcommand{\arraystretch}{1.2} 
    \scriptsize
    \setlength{\tabcolsep}{4pt} 

    \begin{tabularx}{\linewidth}{l X p{0.4\linewidth}}
        \toprule
        \rowcolor{gray!20} 
        \textbf{Criteria} & \textbf{Definition} & \textbf{Indicator} \\
        \midrule
        Factuality 
            & Measure whether the summary is factually accurate with respect to the source transcript. High factual groundedness indicates that no unsubstantiated or contradictory information appears in the summary.
            & Does every claim align with the original transcript? 
             Are there any claims that contradict known facts or the transcript itself? \\
        Completeness 
            & Evaluates how thoroughly the summary includes all critical information required by the target persona. A low score indicates that no essential data needed for decision-making or task execution is missing.
            & Are all key infos needed for the persona’s role present, including budget, deadlines, constraints?
             Is there evidence of a salient fact or figure that was in the source but left out? \\
        Relevance 
            & Measures how focused the summary is on content pertinent to the user’s (or persona’s) role and needs. A low score means minimal extraneous or off-topic information.
            & Is every piece of information purposeful for the user’s role?
             Does the summary emphasize tasks or decisions within the user’s domain? \\
        Goal Alignment 
            & Assesses whether the summary content directly addresses the persona’s primary objectives or responsibilities. Low-scoring summaries are tightly coupled to the persona’s overarching goals.
            & Does the summary thoroughly address the user’s stated or implicit goals?
             Does any part of the summary contradict or ignore the user’s known objectives? \\
        Priority Structuring 
            & Looks at the order and emphasis of information to see if the summary highlights the most urgent or important points first, reflecting the persona’s immediate needs or preferences.
            & Are the most pressing items placed first or highlighted?
             Does the layout or emphasis help the persona quickly find urgent/actionable items? \\
        Knowledge-Level Fit 
            & Check if the level of technical or conceptual detail matches the persona’s expertise. Low scores indicate an optimal level—neither too simplistic nor too advanced.
            & Is the chosen vocabulary suitable for the user’s domain expertise?
             Does the summary offer enough context without overwhelming or patronizing the user? \\
        Contextual Framing 
            & Assesses whether the summary includes the necessary context (historical decisions, cross-departmental references, relevant background) so the user can fully understand the situation.
            & Are past decisions or relevant external factors briefly explained?
             Does the summary clarify dependencies or references to others’ work? \\
        \bottomrule
    \end{tabularx}
    \caption{P-MESA evaluation criteria and indicators for easier identification.}
    \label{tab:p-mesa_criteria}
\end{table*}

\section{Balanced Accuracy Definition}
\label{sec:appendix_b-acc}
Accuracy (ACC) is a natural choice to measure the proportion of correctly predicted labels out of the total number of labels:

\begin{equation}
    ACC = \frac{(TP + TN)}{(TP + FN + FP + TN)}
\end{equation}

with TP - true positive, TN - true negative, FP - false positive, and FN - false negative.
As we cannot exclude a class imbalance within the individual dimensions, reporting accuracy is not suitable, and we report the balanced accuracy (B-ACC), i.e., the arithmetic mean of sensitivity (SEN) and specificity (SPE):

\begin{equation}
    SEN = \frac{TP}{(TP + FN}
\end{equation}

\begin{equation}
    SPE = \frac{TN}{(TN + FP)}
\end{equation}

\begin{equation}
    \text{B-ACC} = \frac{1}{2} (SEN + SPE)
\end{equation}

\section{Comparison with Self-Refinement}
\label{app:self-refinement-comparison}

A recent approach for refining meeting summaries is the assessment of feedback similar to the MESA categories and refining a summary according to this feedback \cite{KirsteinLG25a}.
We exclude this approach from our main experiments as it reiterates the same model and is therefore not a fair comparison to a single-pass approach, and because it is not easily adaptable for personalization due to the lack of feedback criteria, which we present with \pmetric{} (\Cref{sec:pmesa}).
In \Cref{tab:qmsum_ours_vs_feedback}, we compare \framework{} with a GPT backbone and the setup of \Cref{sec:experiments} against a feedback-based summarization refinement approach \cite{KirsteinLG25a} with one, two, and three iterations (abbreviated FB-1, -2, -3).
We use FEEDBACK with the reported best-performing setup, i.e., having multiple agents assess feedback, prompted to use chain-of-thought reasoning, for error assessment, and using the feedback directly along the chain-of-thought trace for refinement.
We observe that \framework{} beats FEEDBACK-1 and FEEDBACK-2 and slightly outperforms FEEDBACK-3.
\framework{} is consistently lower in hallucination (1 point vs. min. 3 points in FB-3).
Coreference, incoherence, and structure are similar consistently.
\framework{} slightly outperforms on omission and irrelevance.
When it comes to repetition and language, \framework{} beats FB-1 and FB-2, but is outperformed by FB-3.
So we conclude that \framework{} is comparable to three turns of feedback-based evaluation due to the fact-based approach that aids content understanding, it performs better with hallucination, omission, and repetition.

\begin{table}[ht]
\scriptsize
\centering
\begin{tabular}{lcccc}
\toprule

\textbf{Approach} & \framework{} & FB-1 & FB-2 & FB-3 \\
\midrule
\rowcolor{gray!20} 
\multicolumn{5}{c}{\textbf{MESA} (lower is better)} \\
\midrule
Coreference     & \cellcolor{highlightGreen}$0_{\textit{1.64}}$ & $1_{\textit{1.54}}$ & \cellcolor{highlightGreen}$0_{\textit{1.30}}$ & $0.5_{\textit{0.86}}$\\
Hallucination   & \cellcolor{highlightGreen}$1_{\textit{1.75}}$ & $4_{\textit{1.93}}$ & $4_{\textit{1.84}}$ & $3_{\textit{1.93}}$\\
Incoherence     & $3_{\textit{1.88}}$ & \cellcolor{highlightGreen}$2.5_{\textit{1.73}}$ & $4_{\textit{1.83}}$ & $3_{\textit{1.73}}$\\
Irrelevance     & \cellcolor{highlightGreen}$1_{\textit{1.45}}$ & $3_{\textit{1.60}}$ & $4_{\textit{1.58}}$ & $2_{\textit{1.60}}$\\
Language        & $1_{\textit{1.40}}$ & $3_{\textit{0.88}}$ & $2_{\textit{0.78}}$ & \cellcolor{highlightGreen}$0_{\textit{0.72}}$\\
Omission        & \cellcolor{highlightGreen}$1_{\textit{0.16}}$ & $4_{\textit{0.00}}$ & $3.5_{\textit{0.32}}$ & $2_{\textit{0.27}}$\\
Repetition      & $1_{\textit{1.23}}$ & $2_{\textit{0.97}}$ & $2_{\textit{1.21}}$ &\cellcolor{highlightGreen}v$0_{\textit{0.84}}$\\
Structure       & \cellcolor{highlightGreen}$3_{\textit{1.24}}$ & \cellcolor{highlightGreen}$3_{\textit{1.76}}$ & \cellcolor{highlightGreen}$3_{\textit{1.58}}$ & $4_{\textit{1.44}}$\\
\midrule
\rowcolor{gray!20} 
\multicolumn{5}{c}{\textbf{ROUGE (R-1, R-2, R-L) and BERTScore (BS)} (higher is better)} \\
\midrule
R-1             & $22.89_{\textit{5.80}}$ & \cellcolor{highlightGreen}$24.19_{\textit{4.74}}$ & $23.86_{\textit{5.34}}$ & $21.20_{\textit{4.44}}$\\
R-2             & $4.13_{\textit{2.44}}$ & $5.33_{\textit{2.86}}$ & \cellcolor{highlightGreen}$5.82_{\textit{2.88}}$ & $5.44_{\textit{2.83}}$\\
R-L             & $20.78_{\textit{5.23}}$ & \cellcolor{highlightGreen}$20.99_{\textit{4.83}}$ & $20.45_{\textit{4.48}}$ & $20.05_{\textit{4.51}}$\\
BS (F1)         & $85.67_{\textit{1.19}}$ & $84.85_{\textit{0.11}}$ & \cellcolor{highlightGreen}$86.50_{\textit{1.07}}$ & $85.90_{\textit{0.73}}$\\
\bottomrule
\end{tabular}
\caption{Comparison of \framework{} summary quality to refinement-based approaches \cite{KirsteinLG25a}. Values are Median$_{Std}$. MESA scores are 1--5 Likert ratings, 
    ROUGE (R-1/R-2/R-L) and BERTScore (BS) are 0--100. \colorbox{highlightGreen}{Green} is best in category.}
\label{tab:qmsum_ours_vs_feedback}
\end{table}

\section{Qualitative Examples}
\label{app:qualitative_example}

This section presents representative examples that illustrate the improvements achieved by our approaches.
These examples provide concrete demonstrations of how \framework{} and \reasoning{} address the challenges stated in \Cref{sec:introduction} and support the quantitative findings reported in \Cref{sec:experiments}. 
All examples are drawn from the QMSum dataset and are generated using the experimental setup described in \Cref{sec:experiments}.

\subsection{Impact of Summary Verification}
\label{subsec:summary_verification_example}

\Cref{tab:summary_refinement_changes} demonstrates the effect of \framework{}'s \texttt{Summary Verification} stage, which ensures conciseness and structural coherence while preserving core content.
This example illustrates how the validation step primarily serves as a quality control mechanism that:
\begin{itemize}
    \item Condenses verbose passages into more concise expressions
    \item Maintains all key decisions and action items
    \item Preserves attribution of statements to specific speakers
    \item Ensures adherence to length constraints (reduced from 313 to 192 words)
\end{itemize}
It supports our finding in \Cref{sec:ablation_studies} that the verification component primarily enforces structural constraints rather than correcting factual content when using a high-capability model like GPT-4o.

\subsection{FRAME vs. Single-LLM Summarization}
\label{subsec:frame_vs_single_llm_example}

\Cref{tab:summary_general_single_vs_ours} contrasts summaries from a single GPT-4o instance versus our \framework{} framework.
This comparison highlights several key advantages of the fact-based approach:
\begin{itemize}
    \item \textbf{Improved Structure}: \framework{}'s summary organizes information thematically rather than chronologically, grouping related concepts (data handling, anonymization, algorithms).
    
    \item \textbf{Higher Specificity}: \framework{} captures concrete decisions (e.g., "Professor D suggested digit recordings") rather than vague descriptions (e.g., "participants debated merits").
    
    \item \textbf{Better Speaker Attribution}: \framework{} consistently attributes statements to specific speakers (Professor D, PhD F, Grad B), preserving accountability and provenance.
    
    \item \textbf{Clearer Action Items}: \framework{} explicitly identifies next steps and responsibilities, whereas the single-LLM approach emphasizes discussion over outcomes.
\end{itemize}
These qualitative differences illustrate why \framework{} achieves lower hallucination and omission scores as reported in \Cref{tab:summary_performance}, demonstrating how our structured fact-based approach addresses the core challenges of meeting summarization identified in \Cref{sec:introduction}.

\subsection{Personalized Summarization}
\label{subsec:personalized_summary_example}

\Cref{tab:summary_personal_single_vs_ours} compares personalized summaries generated for a graduate student (Grad A) using single-LLM and \framework{} approaches.
This example illustrates how \framework{}'s fact-based approach enhances personalization even without explicit \reasoning{} reasoning:

\begin{itemize}
    \item \textbf{Reader-Relevant Focus}: The \framework{} summary prioritizes content most relevant to Grad A's role (model structure, decision nodes, implementation considerations).
    
    \item \textbf{Technical Detail Calibration}: The \framework{} summary provides appropriate technical depth for a graduate student working on the project.
    
    \item \textbf{Implied Consequences}: The \framework{} summary explains why particular design decisions matter (e.g., "This impacts the logical independence and timing of decisions").
    
    \item \textbf{Perspective Preservation}: While both summaries focus on Professor B's guidance, the \framework{} version better contextualizes this guidance from Grad A's perspective.
\end{itemize}
These improvements align with our findings in \Cref{sec:personalized_summarization} that fact-centric approaches provide a stronger foundation for personalization by enabling more deliberate content selection and organization.

\subsection{Impact of Fact Representation}
\label{subsec:fact_representation_example}

\Cref{tab:summary_ours_vs_molecular} compares summaries generated using the same \framework{} architecture but with different fact representation approaches.
This example demonstrates why our statement-context tuple approach outperforms molecular facts (\Cref{sec:app_fact_representation_comparison}):

\begin{itemize}
    \item \textbf{Enhanced Contextual Understanding}: Our approach captures relationships between meeting elements (e.g., connecting software choices to model structures).
    
    \item \textbf{Improved Responsibility Attribution}: Statement-context tuples enable clearer assignment of tasks to specific participants.
    
    \item \textbf{Coherent Organization}: Global context facilitates better organization around themes rather than isolated facts.
    
    \item \textbf{Strategic Prioritization}: Richer context enables more nuanced decisions about which content to emphasize.
\end{itemize}
This example supports our finding that global context enrichment significantly improves the model's ability to understand and reconstruct the meeting's underlying semantic structure.

\subsection{Threshold Sensitivity Analysis}
\label{subsec:threshold_analysis_example}

\Cref{tab:summary_varying_threshold} illustrates how varying the fact retention thresholds affects summary content and structure.
This example supports our threshold sensitivity analysis in \Cref{app:threshold_variation}:

\begin{itemize}
    \item \textbf{Default Threshold}: Creates a balanced summary with focused main points and appropriate supporting details.
    
    \item \textbf{Lower Threshold}: Includes more contextual information but introduces some repetition (e.g., multiple mentions of design simplicity).
    
    \item \textbf{Higher Threshold}: Over-prioritizes specific details (e.g., material properties) at the expense of broader context and coherence.
\end{itemize}

These examples demonstrate why our empirically determined thresholds ($r_i \geq 8$ for outline points, $r_i \geq 6$ for context) achieve optimal results in our main experiments.

\subsection{Personalization Approach Comparison}
\label{subsec:personalization_approach_example}

\Cref{tab:summary_different_personalization_approaches} compares summaries generated using three different personalization approaches: Reader-Tailoring ("Tailor To"), Role-Playing, and our SCOPE protocol.
All summaries are personalized for the same graduate student ("Grad A") and cover identical meeting content about belief networks and decision processes.
The comparison reveals distinctive patterns in how each approach handles personalization:

\begin{itemize}
    \item \textbf{Reader-Tailoring}: Mentions Grad A once but primarily presents comprehensive meeting content without consistently filtering for relevance to Grad A's interests. Includes many technical details without clear prioritization based on Grad A's specific needs.
    
    \item \textbf{Role-Playing}: Provides extensive meeting details with focus on Professor B's contributions, but never explicitly addresses Grad A's interests or needs. Includes tangential information about other participants that may not be directly relevant to Grad A.
    
    \item \textbf{SCOPE}: Consistently relates information back to Grad A's interests, explicitly mentioning how the content aligns with Grad A's interests.
    More concise, selecting only aspects most relevant to Grad A's focus on ``logical structures and reliable information.''
\end{itemize}

These examples illustrate \reasoning{}'s key advantage: explicit reasoning about relevance produces summaries that consistently prioritize content based on the reader's specific interests rather than just providing topic-relevant information. 
This targeted selection supports our quantitative findings in \Cref{sec:personalized_summarization}, demonstrating how \reasoning{}'s structured reasoning approach improves personalization quality.


\begin{table*}[]
\scriptsize
\begin{tabular}{lp{14cm}}
\toprule
\rowcolor{gray!20}
\textbf{Setup}       & \textbf{Summary} \\
\midrule
Before     & The meeting focused on procedural changes, data management, and anonymization strategies. Key outcomes included decisions on meeting formats, data handling, and anonymization methods. Professor D suggested digit recordings at the end of meetings to streamline the process and ensure accurate documentation. PhD F recommended having three or four different kinds of meetings to cater to various research needs and improve data collection. Professor D proposed engaging professors and senior doctoral students to conduct their meetings at the research site for better data collection. Participants agreed that CDs should not be given on the same day to ensure thorough review and accuracy. PhD C plans to use clustering or neural net algorithms for classification and exclude non-speech events from the study to focus on relevant data. Professor D suggested conducting sensitivity analyses to better understand neural network inputs and using a decision tree for classification. The importance of detecting overlaps and having multiple system models was highlighted to improve data accuracy. Various methods for anonymizing data, including sequential labeling and playing audio backward, were discussed to ensure participant privacy. 

Grad B will send an email to Haas Business School volunteers to engage participants. Postdoc E suggested burning the CD after it has been cleared with the transcript stage to ensure data accuracy. PhD A needs to delete the old dynamic data from Carmen's Spanish directory to optimize data storage. Grad B plans to complete the archive and verify it by the end of the day to ensure data integrity. PhD C plans to work on an automatic segmenter to improve data processing and label each frame with a class indicating the type of zone. PhD C also plans to extend the study to include all overlapping events in the future to broaden the research scope.\\
\midrule
After   & The meeting addressed procedural changes, data management, and anonymization strategies. Key decisions included meeting formats, data handling, and anonymization methods. Professor D suggested digit recordings for accurate documentation, while PhD F recommended varied meeting types to enhance data collection. Participants agreed CDs should not be given on the same day for thorough review. PhD C plans to use clustering or neural net algorithms for classification, excluding nonspeech events. Sensitivity analyses and decision trees were proposed for better understanding and classification. Methods for anonymizing data, such as sequential labeling and playing audio backwards, were discussed. Grad B will email Haas Business School volunteers, and Postdoc E suggested burning CDs post-transcript clearance. PhD A will delete old data to optimize storage, and Grad B will complete and verify the archive. PhD C plans to develop an automatic segmenter and extend the study to include overlapping events.\\
\bottomrule
\end{tabular}
\caption{\framework{} generated summary before and after applying the \texttt{Summary Verification} step.}
\label{tab:summary_refinement_changes}
\end{table*}


\begin{table*}[]
\scriptsize
\begin{tabular}{lp{14cm}}
\toprule
\rowcolor{gray!20}
\textbf{Setup}       & \textbf{Summary} \\
\midrule
Single LLM     & The meeting focused on discussing the format for storing and linking various types of data, such as word transcripts, annotations, and prosodic features, in a database. The participants debated the merits of different formats, including XML and P-files, and considered the use of ATLAS, a NIST-developed infrastructure. They emphasized the need for a flexible, easy-to-use format that can handle changes in word transcripts and time boundaries efficiently. The discussion also touched on the challenges of merging different annotations and maintaining consistency across versions. The meeting concluded with a plan to further investigate ATLAS and its tools to determine the best approach for their needs. \\
\midrule
\framework{}   & The meeting addressed procedural changes, data management, and anonymization strategies. Key decisions included meeting formats, data handling, and anonymization methods. Professor D suggested digit recordings for accurate documentation, while PhD F recommended varied meeting types to enhance data collection. Participants agreed CDs should not be given on the same day for thorough review. PhD C plans to use clustering or neural net algorithms for classification, excluding nonspeech events. Sensitivity analyses and decision trees were proposed for better understanding and classification. Methods for anonymizing data, such as sequential labeling and playing audio backwards, were discussed. Grad B will email Haas Business School volunteers, and Postdoc E suggested burning CDs post-transcript clearance. PhD A will delete old data to optimize storage, and Grad B will complete and verify the archive. PhD C plans to develop an automatic segmenter and extend the study to include overlapping events.\\
\bottomrule
\end{tabular}
\caption{Sample QMSum meeting transcript summarizes by a single GPT-4o instance and our \framework{} framework with GPT-4o backend.}
\label{tab:summary_general_single_vs_ours}
\end{table*}



\begin{table*}[]
\scriptsize
\begin{tabular}{lp{14cm}}
\toprule
\rowcolor{gray!20}
\textbf{Setup}       & \textbf{Summary} \\
\midrule
Single LLM  & Professor B suggested focusing on building influence links within the path domain, prioritizing those related to paths over database influences.  He proposed laying out the influence structure and determining necessary belief nodes, with the design completed before computation methods are decided.  Professor B confirmed the program being used is a GUI for a belief-net simulator. He advised focusing on strong influences when considering factors like "slurred" or "angry" speech.  He suggested merging selling, fixing, and servicing entities for endpoint decisions, potentially creating an intermediate "commercial" entity type alongside "landmark." Professor B recommended removing the direct link between the user and endpoint, instead creating an intermediate "endpoint U" node representing the endpoint decision from the user model perspective. He also suggested a "fit" node to measure compatibility between the entity and user.  Professor B advised against excessive complexity in the user model to avoid combinatorial explosion, suggesting a compact representation of user state or compatibility. He proposed shifting focus to other path-related decisions and their structure within the belief net.  Professor B emphasized the importance of avoiding loops in the belief net to prevent ill-defined computations. He highlighted the need for organizational clarity, explaining that discourse properties will eventually feed into discourse-based decision nodes.  Professor B raised the issue of combining influences mathematically at decision nodes, asking how to determine the rule of combination. He also requested a display solution for the complex diagrams. Finally, Professor B initiated a discussion about decision nodes and combination choices, prompting an explanation of expert weighting and mixture of experts models.\\
\midrule
\framework{}   & During the meeting, Professor B proposed merging selling, fixing, and servicing into an intermediate structure for endpoint decisions, emphasizing the creation of a new node based on user models. Professor B confirmed that the endpoint decision node should be linked to the user model, ensuring accurate decision-making, and agreed to use 'endpoint U' for this node. Additionally, Professor B suggested that the 'Go-there' decision should have a node for discourse and binary values.
Professor B planned to outline the influence structure, advising a focus on strong influences in the belief-net model and completing the basic design before computing it. Grad D suggested an intermediate node for the Endpoint and Go-there nodes, which Professor B agreed could feed into the Go-there node, incorporating situational factors. This impacts the logical independence and timing of decisions. Professor B emphasized avoiding loops in the belief net to prevent convergence issues, highlighting the need for careful knowledge engineering to manage system complexity. These insights are vital for Grad A's understanding of decision-making nuances and ensuring model reliability.
\\

\bottomrule
\end{tabular}
\caption{Example personalized summary of the same meeting for Grad Student A. We compare \framework{} with \reasoning{} against GPT-4o prompted to tailor to a specific reader.}
\label{tab:summary_personal_single_vs_ours}
\end{table*}


\begin{table*}[]
\scriptsize
\begin{tabular}{lp{14cm}}
\toprule
\rowcolor{gray!20}
\textbf{Setup}       & \textbf{Summary} \\
\midrule
Molecular  &  The meeting focused on the development and implementation of the JavaBayes program and the discourse model. Key decisions included switching to JavaBayes for its probabilistic modeling capabilities, adding a middle layer to enhance model complexity and accuracy, and standardizing presentation guidelines with 'Future Work' and 'What I Learned' slides for clarity. Grad A proposed reverting to a previous stable state to ensure model reliability, while Grad B set limits for searching past discussions to improve efficiency. A trial run for data collection was scheduled for Friday afternoon, with Fey volunteering as the wizard and Liz as the first subject. Grad A faces challenges in setting up probabilities for a flat structure due to exponential possibilities with N features. The discourse model aids in anaphora resolution and coordinating gesture screen issues, providing a clear interface for querying past discussions. \\
\midrule
Our facts  & The meeting focused on developing and implementing a discourse model and data collection strategies. Key outcomes included decisions on model structure, data collection plans, and software tools. Grad A was confirmed to streamline project management, providing clear leadership and accountability. The group agreed to add a middle layer to the model to enhance differentiation between user contexts, improving accuracy and functionality. Grad A suggested reverting to a previous stable version for better reliability. Grad B confirmed their presentation is ready. The discourse model will have a limit for searching past discussions to optimize performance. Features and variables such as prosody, discourse, verb choice, and context categories were discussed. The switch to JavaBayes was noted, and suggestions were made for generating probability tables and friendly data entry scripts. Data collection strategies included trial runs scheduled for Friday afternoon. Next steps involved Grad B switching to JavaBayes, identifying hidden nodes, and segmenting regions to create observable nodes. Responsibilities were assigned to Grad A, Grad B, Fey, and Liz.\\
\bottomrule
\end{tabular}
\caption{Example summaries of \framework{} with molecular facts and our fact definition.}
\label{tab:summary_ours_vs_molecular}
\end{table*}


\begin{table*}[]
\scriptsize
\begin{tabular}{lp{14cm}}
\toprule
\rowcolor{gray!20}
\textbf{Setup}       & \textbf{Summary} \\
\midrule
default  &  The meeting focused on the design and functionality of a new remote control, emphasizing simplicity and ease of use. Key decisions included essential functions like volume, channel selection, on-off, mute, and text TV buttons. Marketing stressed the importance of an attractive look to appeal to users, while technological innovation was secondary. The team opted against LCD and speech recognition to simplify design and reduce costs, setting a budget of twelve and a half Euros per unit. Rechargeable batteries and a scroll wheel for volume control were chosen for sustainability and functionality. The remote will have a rounded shape, transparent case, and modern look, targeting a young, trendy audience. Next steps include prototype development, engineering contacts, design review, and market testing. \\
\midrule
lower   & The meeting centered on designing a new remote control, emphasizing simplicity and user-friendliness. Key decisions included a basic design with minimal buttons to enhance usability, while retaining essential features like text TV functionality for compatibility with existing systems. The remote will be encased in durable hard plastic, balancing cost-effectiveness and reliability. To appeal to a young audience, the design will incorporate flashy colors and a modern aesthetic, featuring a Real Reaction logo for brand visibility. Rechargeable batteries will be used to promote sustainability, and a scroll wheel will provide intuitive volume control. The remote will slide into a docking station for convenient charging. The team opted against LCD and speech recognition due to complexity and cost, focusing instead on affordability with a budget of twelve and a half Euros per unit. The design aims to be trendy yet straightforward, ensuring ease of use and competitive appeal in the market.\\
\midrule
higher   & The meeting centered on designing a new remote control, focusing on its appearance, functionality, and target audience. Key decisions included opting for a basic design with minimal buttons to simplify the user interface and enhance usability. The remote will feature text TV functionality, ensuring compatibility with existing technology, and a spongy material for improved durability and comfort. The team decided against using LCD or speech recognition to reduce complexity and cost, setting the budget at twelve and a half Euros per unit. The remote will have essential buttons like on-off, mute, and possibly two others, with all buttons on one level for better usability. It will use rechargeable batteries and a regular chip for reliability, with a transparent case to allow backlight visibility. Normal infrared LEDs and backlight LEDs will be included, along with a scroll wheel for volume control and zapping functionality to enhance user satisfaction. The remote will feature the Real Reaction logo and will be modern yet basic, aiming to attract the target audience. \\
\bottomrule
\end{tabular}
\caption{Example summary with varying thresholds using \framework{}.}
\label{tab:summary_varying_threshold}
\end{table*}


\begin{table*}[]
\scriptsize
\begin{tabular}{lp{14cm}}
\toprule
\rowcolor{gray!20}
\textbf{Setup}       & \textbf{Summary} \\
\midrule
Tailor To  &  In the meeting, Professor B proposed focusing on building influence links to determine which sets of factors are relevant to specific decisions, starting with path-related ones. Grad A emphasized the importance of logical independence and timing in decision-making, suggesting that factors like "Go-there" should be binary and separate from other possibilities. Grad E discussed the complexity of defining discrete values for variables like running budgets and user interests, highlighting the challenge of capturing nuanced decision-making criteria. Grad D introduced the concept of weighted averages and mixtures of experts to combine multiple inputs into a single decision node, stressing the need for a principled approach to ensure valid probability distributions. The discussion also touched on the importance of integrating discourse features and ontology properties into the belief-net structure, with Grad A questioning how human-like decision processes could be modeled. The meeting concluded with a focus on organizing and combining influences mathematically, acknowledging the need for a display hack to manage the complexity of the belief-net diagrams. Overall, the meeting underscored the importance of logical structure, reliability, and detailed analysis in decision-making processes. \\
\midrule
Roleplaying   & During the meeting, Professor B emphasized the importance of building influence links to understand which sets of factors are relevant to specific decisions. He suggested starting with path-related decisions, considering database-related ones less interesting but necessary. Professor B and Bhaskara plan to identify belief nodes required for decision-making, ensuring the basic design of influence structures is completed before computing them. Grad E mentioned the user’s budget influencing decision outcomes, highlighting the need for a running total of finances. Professor B clarified that updates to the budget must be external to the belief-net and influence various decisions. Grad E expressed confusion about writing discrete values for a running budget, which Professor B acknowledged, suggesting external conditioning for updates.

The discussion shifted to decision nodes and their combination rules. Grad D explained the need for a principled way to combine expert decisions, proposing weighted averages and mixture models to handle varying reliability of experts in different situations. Professor B stressed the importance of avoiding loops in belief-nets to ensure well-defined computations. Grad E and Professor B discussed the necessity of integrating discourse properties into decision nodes, with Keith focusing on constructions indicating various properties and Johno committed to parsing wizardry. The meeting concluded with a focus on organizing influences and combining them mathematically, ensuring the system remains feasible and clean for decision-making processes. \\
\midrule
\reasoning{}   & Professor B emphasized the importance of laying out influence structures to understand what factors affect decisions, suggesting a focus on path influence links initially. This aligns with Grad A's interest in logical structures and decision-making processes. The discussion highlighted the need to establish belief nodes, which are crucial for computing decisions based on influence structures. Grad E and Professor B explored how user models and situation models contribute to decision-making, providing insights into logical independence and reliability of features. The conversation also touched on the technical aspects of belief nets, which are essential for Grad A's analytical approach. The urgency to complete the basic design before computation was noted, aligning with Grad A's priorities. The meeting provided valuable insights into how different factors influence decision-making, offering a detailed analysis that supports Grad A's focus on logical structures and reliable information. \\
\bottomrule
\end{tabular}
\caption{Different personalization approaches. We compare 'tailor-to' prompting against 'role-playing' prompting and our reason-out-loud protocol. All summaries cover the same meeting and are personalized to ``Grad A''.}
\label{tab:summary_different_personalization_approaches}
\end{table*}

\clearpage
\onecolumn

\clearpage
\onecolumn
\hypertarget{annotation}{}
\pagestyle{empty}
\lstset{
  basicstyle=\footnotesize\ttfamily,
  breaklines=true,
  breakatwhitespace=false,
  columns=flexible,
  numbers=none
}

\definecolor{Primary}{RGB}{59, 130, 246}    
\definecolor{PrimaryDark}{RGB}{30, 64, 175} 
\definecolor{LightBg}{RGB}{239, 246, 255}   
\definecolor{TextDark}{RGB}{31, 41, 55}     
\definecolor{TextMuted}{RGB}{107, 114, 128} 

\begin{tikzpicture}[remember picture, overlay]
  \fill[Primary] ([xshift=0cm,yshift=0cm]current page.north west) rectangle ([xshift=\paperwidth,yshift=-0.4cm]current page.north west);
\end{tikzpicture}

\vspace{0.8cm}
\begin{center}
  {\fontsize{22}{26}\selectfont\sffamily\bfseries \textcolor{PrimaryDark}{CiteAssist}}\\[0.2em]
  {\Large\sffamily\scshape \textcolor{TextMuted}{Citation Sheet}}\\[0.8em]
  {\small\sffamily Generated with \href{https://citeassist.uni-goettingen.de/}{\textcolor{Primary}{\texttt{citeassist.uni-goettingen.de}}}
  \CiteAssistCite{}
  }\end{center}

\begin{center}
\vspace{1em}
\begin{tikzpicture}
\draw[Primary, line width=0.6pt] (0,0) -- (\textwidth,0);
\end{tikzpicture}
\vspace{1.2em}
\end{center}

\begin{tcolorbox}[enhanced,
                 frame hidden,
                 boxrule=0pt,
                 borderline west={2pt}{0pt}{Primary},
                 colback=LightBg,
                 sharp corners,
                 fonttitle=\sffamily\bfseries\large,
                 coltitle=Primary,
                 title=BibTeX Entry,
                 attach title to upper={\vspace{0.2em}\par},
                 left=12pt]
\lstset{
    inputencoding = utf8,  
    extendedchars = true,  
    literate      =        
      {á}{{\'a}}1  {é}{{\'e}}1  {í}{{\'i}}1 {ó}{{\'o}}1  {ú}{{\'u}}1
      {Á}{{\'A}}1  {É}{{\'E}}1  {Í}{{\'I}}1 {Ó}{{\'O}}1  {Ú}{{\'U}}1
      {à}{{\`a}}1  {è}{{\`e}}1  {ì}{{\`i}}1 {ò}{{\`o}}1  {ù}{{\`u}}1
      {À}{{\`A}}1  {È}{{\`E}}1  {Ì}{{\`I}}1 {Ò}{{\`O}}1  {Ù}{{\`U}}1
      {ä}{{\"a}}1  {ë}{{\"e}}1  {ï}{{\"i}}1 {ö}{{\"o}}1  {ü}{{\"u}}1
      {Ä}{{\"A}}1  {Ë}{{\"E}}1  {Ï}{{\"I}}1 {Ö}{{\"O}}1  {Ü}{{\"U}}1
      {â}{{\^a}}1  {ê}{{\^e}}1  {î}{{\^i}}1 {ô}{{\^o}}1  {û}{{\^u}}1
      {Â}{{\^A}}1  {Ê}{{\^E}}1  {Î}{{\^I}}1 {Ô}{{\^O}}1  {Û}{{\^U}}1
      {œ}{{\oe}}1  {Œ}{{\OE}}1  {æ}{{\ae}}1 {Æ}{{\AE}}1  {ß}{{\ss}}1
      {ẞ}{{\SS}}1  {ç}{{\c{c}}}1 {Ç}{{\c{C}}}1 {ø}{{\o}}1  {Ø}{{\O}}1
      {å}{{\aa}}1  {Å}{{\AA}}1  {ã}{{\~a}}1  {õ}{{\~o}}1 {Ã}{{\~A}}1
      {Õ}{{\~O}}1  {ñ}{{\~n}}1  {Ñ}{{\~N}}1  {¿}{{?\`}}1  {¡}{{!\`}}1
      {„}{\quotedblbase}1 {“}{\textquotedblleft}1 {–}{$-$}1
      {°}{{\textdegree}}1 {º}{{\textordmasculine}}1 {ª}{{\textordfeminine}}1
      {£}{{\pounds}}1  {©}{{\copyright}}1  {®}{{\textregistered}}1
      {«}{{\guillemotleft}}1  {»}{{\guillemotright}}1  {Ð}{{\DH}}1  {ð}{{\dh}}1
      {Ý}{{\'Y}}1    {ý}{{\'y}}1    {Þ}{{\TH}}1    {þ}{{\th}}1    {Ă}{{\u{A}}}1
      {ă}{{\u{a}}}1  {Ą}{{\k{A}}}1  {ą}{{\k{a}}}1  {Ć}{{\'C}}1    {ć}{{\'c}}1
      {Č}{{\v{C}}}1  {č}{{\v{c}}}1  {Ď}{{\v{D}}}1  {ď}{{\v{d}}}1  {Đ}{{\DJ}}1
      {đ}{{\dj}}1    {Ė}{{\.{E}}}1  {ė}{{\.{e}}}1  {Ę}{{\k{E}}}1  {ę}{{\k{e}}}1
      {Ě}{{\v{E}}}1  {ě}{{\v{e}}}1  {Ğ}{{\u{G}}}1  {ğ}{{\u{g}}}1  {Ĩ}{{\~I}}1
      {ĩ}{{\~\i}}1   {Į}{{\k{I}}}1  {į}{{\k{i}}}1  {İ}{{\.{I}}}1  {ı}{{\i}}1
      {Ĺ}{{\'L}}1    {ĺ}{{\'l}}1    {Ľ}{{\v{L}}}1  {ľ}{{\v{l}}}1  {Ł}{{\L{}}}1
      {ł}{{\l{}}}1   {Ń}{{\'N}}1    {ń}{{\'n}}1    {Ň}{{\v{N}}}1  {ň}{{\v{n}}}1
      {Ő}{{\H{O}}}1  {ő}{{\H{o}}}1  {Ŕ}{{\'{R}}}1  {ŕ}{{\'{r}}}1  {Ř}{{\v{R}}}1
      {ř}{{\v{r}}}1  {Ś}{{\'S}}1    {ś}{{\'s}}1    {Ş}{{\c{S}}}1  {ş}{{\c{s}}}1
      {Š}{{\v{S}}}1  {š}{{\v{s}}}1  {Ť}{{\v{T}}}1  {ť}{{\v{t}}}1  {Ũ}{{\~U}}1
      {ũ}{{\~u}}1    {Ū}{{\={U}}}1  {ū}{{\={u}}}1  {Ů}{{\r{U}}}1  {ů}{{\r{u}}}1
      {Ű}{{\H{U}}}1  {ű}{{\H{u}}}1  {Ų}{{\k{U}}}1  {ų}{{\k{u}}}1  {Ź}{{\'Z}}1
      {ź}{{\'z}}1    {Ż}{{\.Z}}1    {ż}{{\.z}}1    {Ž}{{\v{Z}}}1  {ž}{{\v{z}}}1
  }
\begin{lstlisting}
@inproceedings{Kirstein2025d,
    title = "Re-{FRAME} the Meeting Summarization {SCOPE}: Fact-Based Summarization and Personalization via Questions",
    author = "Kirstein, Frederic  and
      Kumar, Sonu  and
      Ruas, Terry  and
      Gipp, Bela",
    editor = "Christodoulopoulos, Christos  and
      Chakraborty, Tanmoy  and
      Rose, Carolyn  and
      Peng, Violet",
    booktitle = "Findings of the Association for Computational Linguistics: EMNLP 2025",
    month = nov,
    year = "2025",
    address = "Suzhou, China",
    publisher = "Association for Computational Linguistics",
    url = "https://aclanthology.org/2025.findings-emnlp.1094/",
    doi = "10.18653/v1/2025.findings-emnlp.1094",
    pages = "20087--20137",
    ISBN = "979-8-89176-335-7",
}
\end{lstlisting}
\end{tcolorbox}


\end{document}